\documentclass{article} 
\usepackage{iclr2026_conference,times}

\usepackage{amsmath,amsfonts,bm}

\def\eqref#1{equation~\ref{#1}}

\def\1{\bm{1}}

\DeclareMathAlphabet{\mathsfit}{\encodingdefault}{\sfdefault}{m}{sl}
\SetMathAlphabet{\mathsfit}{bold}{\encodingdefault}{\sfdefault}{bx}{n}

\usepackage{hyperref}
\usepackage{url}

\usepackage[utf8]{inputenc}
\usepackage[T1]{fontenc}
\usepackage{graphicx}
\usepackage{comment}
\usepackage{xspace}
\usepackage{amsmath}
\usepackage{amssymb}
\usepackage{booktabs}
\usepackage{subcaption}
\usepackage{tcolorbox}
\tcbuselibrary{listingsutf8}
\usepackage{tabularx}
\usepackage{makecell}
\usepackage{siunitx}
\usepackage{fancyhdr}
\usepackage{xcolor}
\usepackage{nicefrac}
\usepackage{microtype}
\usepackage{enumitem}

\setlist{leftmargin=8.0mm}

\usepackage{wrapfig}

\usepackage{titlesec}

\titlespacing*{\section}
  {0pt}{*1.8}{*0.7}   

\titlespacing*{\subsection}
  {0pt}{*1.4}{*0.5}   

\def\mystrut(#1,#2){\vrule height #1pt depth #2pt width 0pt}   

\newtcblisting{promptbox}{
  colback   = white!5,
  colframe  = black!70,
  listing only,
  listing engine = listings,
  listing options = {
      basicstyle=\ttfamily\tiny,
  },
  left=1pt, right=1pt, top=1pt, bottom=1pt,
  boxrule=0.4pt, arc=2pt
}
\DeclareCaptionType{prompt}[Prompt][List of Prompts]

\definecolor{lightblue}{RGB}{135, 214, 255}
\definecolor{mediumblue}{RGB}{24, 123, 205}
\definecolor{darkblue}{RGB}{3, 37, 76}
\definecolor{gold}{RGB}{255, 215, 0}
\definecolor{grey}{RGB}{187, 187, 187}
\definecolor{lightpurple}{RGB}{232,209,255}
\definecolor{lightgray}{RGB}{239,239,239}

\newcommand{\lightbluetext}[1]{\colorbox{lightblue}{\mystrut(.5, .5) #1}}
\newcommand{\mediumbluetext}[1]{\colorbox{mediumblue}{\textcolor{lightgray}{\mystrut(.5, .5) #1}}}
\newcommand{\darkbluetext}[1]{\colorbox{darkblue}{\textcolor{white}{\mystrut(.5, .5) #1}}}
\newcommand{\goldtext}[1]{\colorbox{gold}{\mystrut(.5, .5) #1}}
\newcommand{\greytext}[1]{\colorbox{grey}{\mystrut(.5, .5) #1}}
\newcommand{\lightpurpletext}[1]{\colorbox{lightpurple}{\mystrut(.5, .5) #1}}

\title{Hidden in the Haystack: Smaller Needles \\ are More Difficult for LLMs to Find}

\author{\textbf{Owen Bianchi}\textsuperscript{1,2} \quad
  \textbf{Mathew J. Koretsky}\textsuperscript{1,2} \quad
  \textbf{Tanay Nayak}\textsuperscript{2,3} \quad
  \textbf{Chelsea X. Alvarado}\textsuperscript{1,2}  \\ 
  \textbf{Maya Willey}\textsuperscript{1,2} \quad
  \textbf{Adi Asija}\textsuperscript{2,3}  \quad
  \textbf{Nicole Kuznetsov}\textsuperscript{1,2} \quad
  \textbf{Mike A. Nalls}\textsuperscript{1,2,4}  \\
  \textbf{Faraz Faghri}\textsuperscript{1,2,4*} \quad
  \textbf{Daniel Khashabi}\textsuperscript{2,3*} 
  \\[1ex]
  \textsuperscript{1}Center for Alzheimer’s Disease and Related Dementias, NIA, NIH; \; 
  \textsuperscript{2}DataTecnica LLC;  \\ 
  \textsuperscript{3}Johns Hopkins University;  \;
  \textsuperscript{4}Laboratory of Neurogenetics, NIA, NIH; \; \small * Equal contribution
}

\newcommand{\changed}[1]{\textcolor{black}{\mystrut(.5, .5) #1}}

\iclrfinalcopy
\begin{document}

\maketitle

\begin{abstract}
Large language models (LLMs) face significant challenges with needle-in-a-haystack tasks, where relevant information (``the needle``) must be drawn from a large pool of irrelevant context (``the haystack``). Previous studies have highlighted positional bias and distractor quantity as critical factors affecting model performance, yet the influence of \textit{gold context size}, the length of the answer-containing document, has received little attention. We present the first systematic study of gold context size in long-context question answering, spanning three diverse benchmarks (general knowledge, biomedical reasoning, and mathematical reasoning), eleven state-of-the-art LLMs (including recent reasoning models), and more than 150K controlled runs. Our experiments reveal that LLM performance drops sharply when the gold context is shorter, i.e., \textbf{smaller gold contexts consistently degrade model performance and amplify positional sensitivity}, posing a major challenge for agentic systems that must integrate scattered, fine-grained information of \textit{varying lengths}. This effect persists under rigorous confounder analysis: even after controlling for gold context position, answer token repetition, gold-to-distractor ratio, distractor volume, and domain specificity, gold context size remains a decisive, independent predictor of success. Our work provides clear insights to guide the design of robust, context-aware LLM-driven systems.
\end{abstract}

\section{Introduction}

Large language models (LLMs) increasingly power applications that require reasoning over vast amounts of information, such as synthesizing evidence across scientific literature~\citep{sprueill2024chemreasoner,bazgir2025agentichypothesis,weiqiwang2025arxiv2table,gao2025sciencehierarchy} or  navigating complex codebases~\citep{liu2023repobench,zhang2023repocoder,bogomolov2024long}. 

A critical stage in such systems is \textit{aggregation}, the synthesis of retrieved evidence into an accurate, actionable response. This stage determines what content to include, cite, or ignore, and has direct implications for safety, reliability, and factual correctness. 
\changed{One specific, common variant of aggregation is }
 \textit{needle-in-a-haystack} \changed{(NIAH)} scenarios, where relevant evidence (the `\goldtext{gold context}') is embedded within a large volume of topically related or superficially plausible but ultimately irrelevant or misleading, `\greytext{distractor context}'~\citep{tay2021long,shaham2022scrolls}. Successful aggregation requires precise identification  of  essential evidence among a large number of distractor documents.

\changed{
In NIAH scenarios, prior work has explored phenomena such as \textit{positional bias}~\citep{wang2023primacy,liu2024lost}, showing that models disproportionately attended to early content and that distractors degrade performance. Yet one key dimension remains underexplored: \textit{how the size of the gold context influences model performance.}
This question is critical for the design of agentic systems, where autonomous agents must perform NIAH-style aggregation over heterogeneous information streams produced by specialized components, whose evidence can \textit{vary widely in length}.
}

\begin{figure}[t]
    \centering
    \vspace{-0.1cm}
    \includegraphics[width=0.99\textwidth,trim=1.2cm 5.8cm 4.2cm 0.8cm]{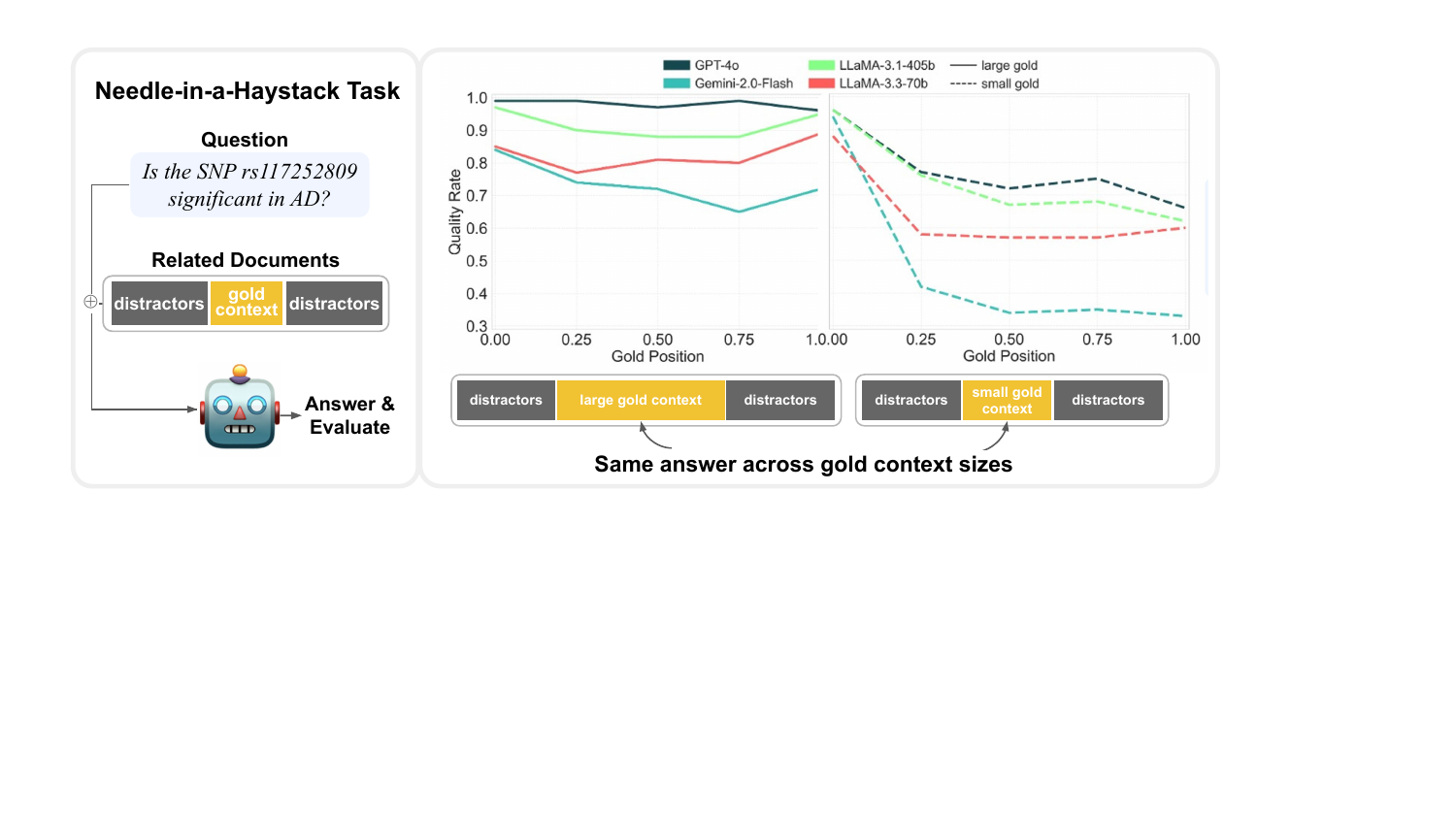}
        \caption{
        \changed{LLM performance on a needle-in-a-haystack task (CARDBiomedBench). The \textit{y}-axis is a measure of accuracy (higher is better).}
        Varying both the size and position of the \goldtext{gold context} within fixed \greytext{distractor} content, we find that LLMs perform \textit{worse} and exhibit \textit{greater positional sensitivity} when given short gold contexts (dashed line) compared to long ones (solid line).
        }
    \label{fig:teaser}
\end{figure}

\changed{
In this study, we present a systematic analysis of gold context size as an independent variable in LLM performance in long-context, single-needle NIAH settings. Our main finding (\S\ref{sec:findings}) is that smaller gold contexts consistently (1) \textbf{degrade performance} and (2)  \textbf{heighten sensitivity to positional bias}, as illustrated in Fig.\ref{fig:teaser}. This exposes a form of brittleness not captured in prior work. 
}

To support our analysis, we adapt three diverse datasets spanning mathematics, biomedical reasoning, general knowledge, systematically varying both the size and position of the gold context \textit{while keeping the distractor content fixed} (\S\ref{sec:experiments}). Our study comprises over 150K controlled runs across eleven state-of-the-art LLMs, both general-purpose and reasoning-focused, ensuring that the observed effects hold across architectures and tasks. 

\changed{
We conduct extensive confounder analyses to ensure that our findings are not driven by spurious factors (\S\ref{sec:confounding:factors}). Even after controlling for gold position, answer-token repetition, gold-to-distractor ratio, distractor volume, and domain specificity, gold context size remains a strong, independent predictor of performance. The overall pattern is consistent---larger gold contexts outperform smaller ones---though the magnitude of the effect varies across conditions. For instance, smaller gold contexts are particularly vulnerable to primacy bias, whereas larger gold ones exhibit greater robustness to both positional variation and distractor interference.
}


\changed{
These findings have practical implications. In real-world deployments, factors such as context size, position, distractor length, and noise are typically uncontrolled. Our results show that heterogeneity in document length can induce aggregation failures, particularly when small but critical evidence is overshadowed by much longer passages. This situation naturally arises in retrieval-augmented and agentic systems, where an aggregator must isolate a single relevant piece of evidence from a mixture of heterogeneous sources. Practitioners can mitigate this fragility by monitoring length disparities among retrieved documents and by avoiding pipelines that combine extremely short and very long contexts, thereby reducing size-induced attention asymmetries.
}




In summary, our contributions are as follows: 
\begin{itemize} [itemsep=0pt, parsep=0pt, topsep=0pt]
    \item \textbf{Novel determinant of long-context performance:} To the best of our knowledge, we are the first to demonstrate that the \textit{size of gold contexts} functions as a hidden variable in LLM performance on long-context NIAH. Smaller golds degrade accuracy and amplify positional bias, underscoring a potential fragility in real-world applications.
    \item \textbf{Robust to confounders:} We identify and analyze five potential confounding variables, (1) gold context position, (2) answer token repetition, (3) gold-to-distractor ratio, (4) distractor volume, and (5) domain specificity, demonstrating that gold context size remains a decisive predictor of success despite these factors.
    \item \textbf{Large-scale experimentation:} We repeat our experiments and aggregate findings across eleven state-of-the-art LLMs, three diverse benchmarks, three sizes of gold, and six positions in the context window totaling over 150k controlled runs.
\end{itemize}


\section{Experimental Setup}
\label{sec:experiments}

This section outlines our design objectives, benchmark adaptations, baseline validations. 

\subsection{Desiderata}
We begin by outlining the core desiderata guiding our experimental design.

\textbf{Realism.} In real-world agentic systems, aggregation involves synthesizing outputs from multiple specialized agents, each retrieving information from their domain of expertise. Usually, one agent returns the document containing the correct answer (the “gold context"), while others provide distractors, topically relevant but ultimately uninformative. We simulate this by inserting a gold context of varying size at different positions within a fixed-length sequence of distractors. Document order is randomized to reflect natural uncertainty in agent contributions and retrieval quality.

\textbf{Gold Size Variability.} We constructed three nested gold variants for each benchmark:
\begin{itemize} [itemsep=0pt, parsep=0pt, topsep=0pt]
    \item \textbf{\lightbluetext{Small Gold}}: Minimal span sufficient to answer the question.
    \item \textbf{\mediumbluetext{Medium Gold}}: Additional explanatory or supporting content.
    \item \textbf{\darkbluetext{Large Gold}}: Complete reasoning process and/or extended relevant context.
\end{itemize}
These were wrapped in pseudo-documents (titles, questions). Variants are hierarchically structured (\lightbluetext{small} $\subset$ \mediumbluetext{medium} $\subset$ \darkbluetext{large}) and validated for sufficiency. See Figure~\ref{fig:gold_sizes_examples} for examples. Performance is high and uniform when observing only the gold of any size (Appendix~\ref{apdx:baselines}).

\textbf{Distractors.} To simulate realistic scenarios, we curate distractors topically relevant and lexically similar to the question but lacking the answer. 
\changed{
The total distractor budget per benchmark is $\sim20k$ tokens---a deliberate design choice that provides a sufficiently long distractor context to be considered `long-context'  while remaining computationally manageable for extensive experimentation.
}


\textbf{Generality.} We select three diverse benchmarks spanning biomedical, general knowledge, and mathematical reasoning, and evaluate performance across eleven leading LLMs of varying architecture and scale. This ensures that our findings generalize across domains and model classes.

\subsection{Task Construction: Needles and Haystacks}
\label{subsec:task:construction}

\begin{figure}[ht]
    \centering
    \begin{subfigure}{0.329\textwidth}
        \includegraphics[width=\linewidth]{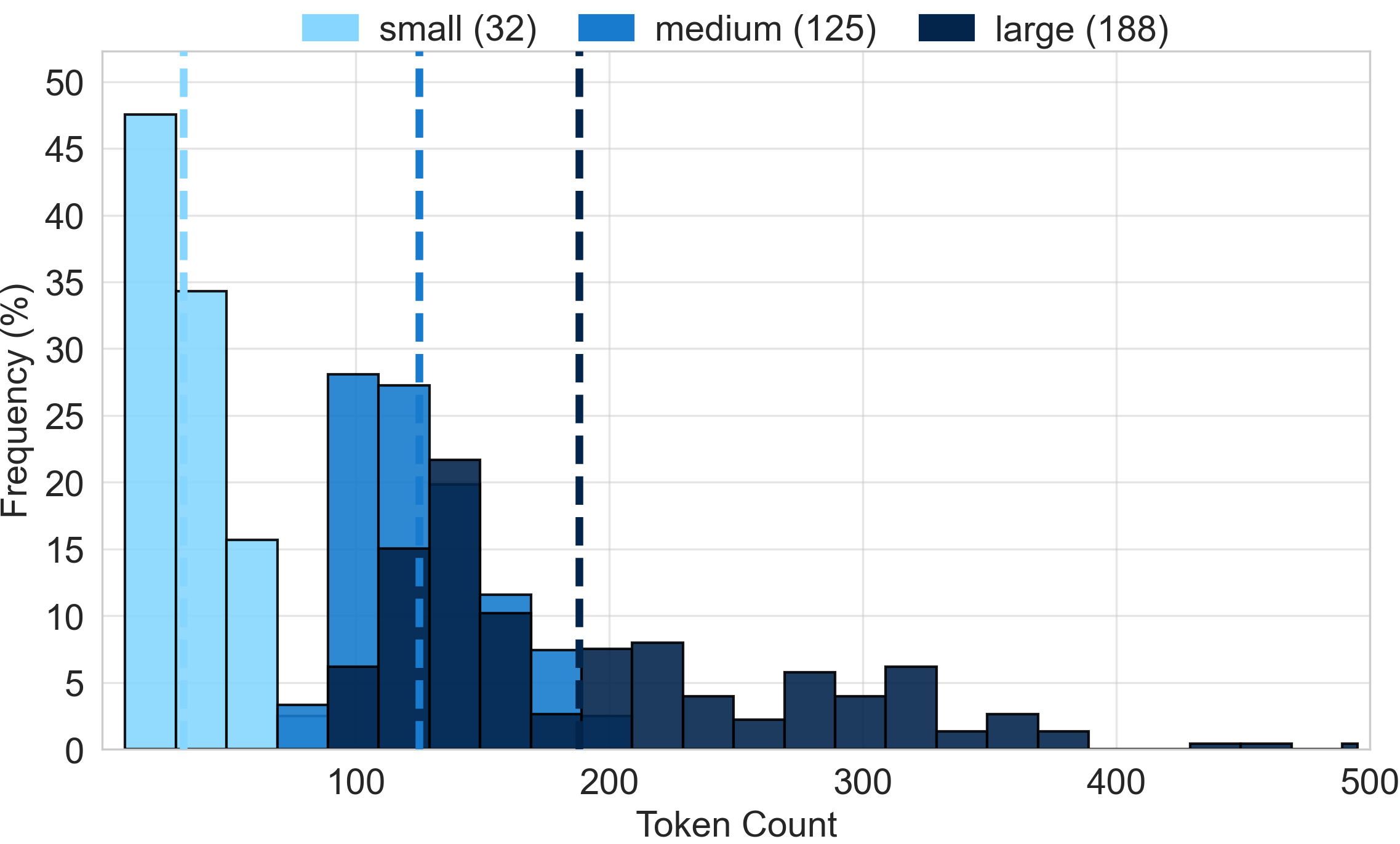}
        \caption{CARDBiomedBench}
    \end{subfigure}
    \hfill
    \begin{subfigure}{0.329\textwidth}
        \includegraphics[width=\linewidth]{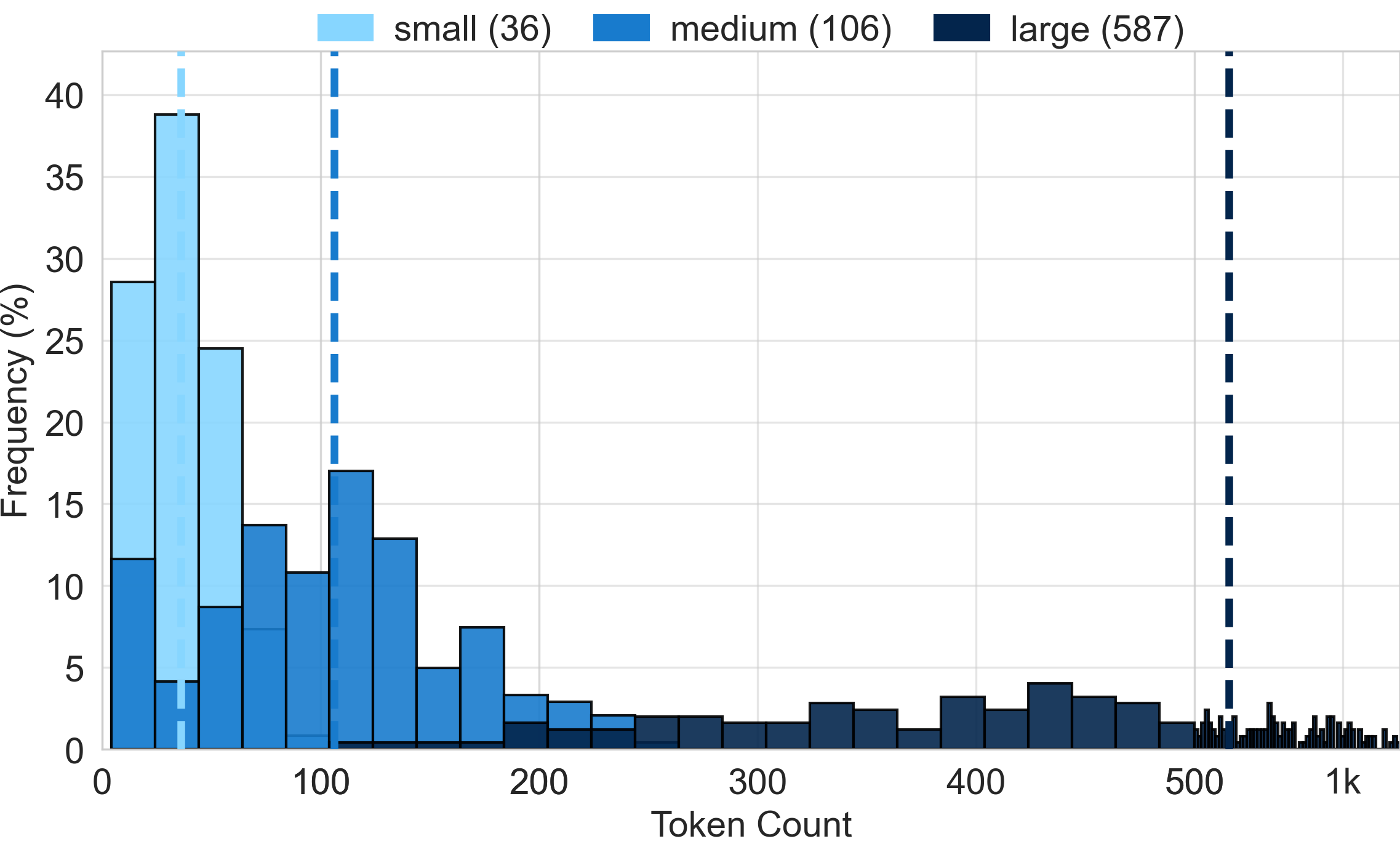}
        \caption{NaturalQuestions}
    \end{subfigure}
    \hfill
    \begin{subfigure}{0.329\textwidth}
        \includegraphics[width=\linewidth]{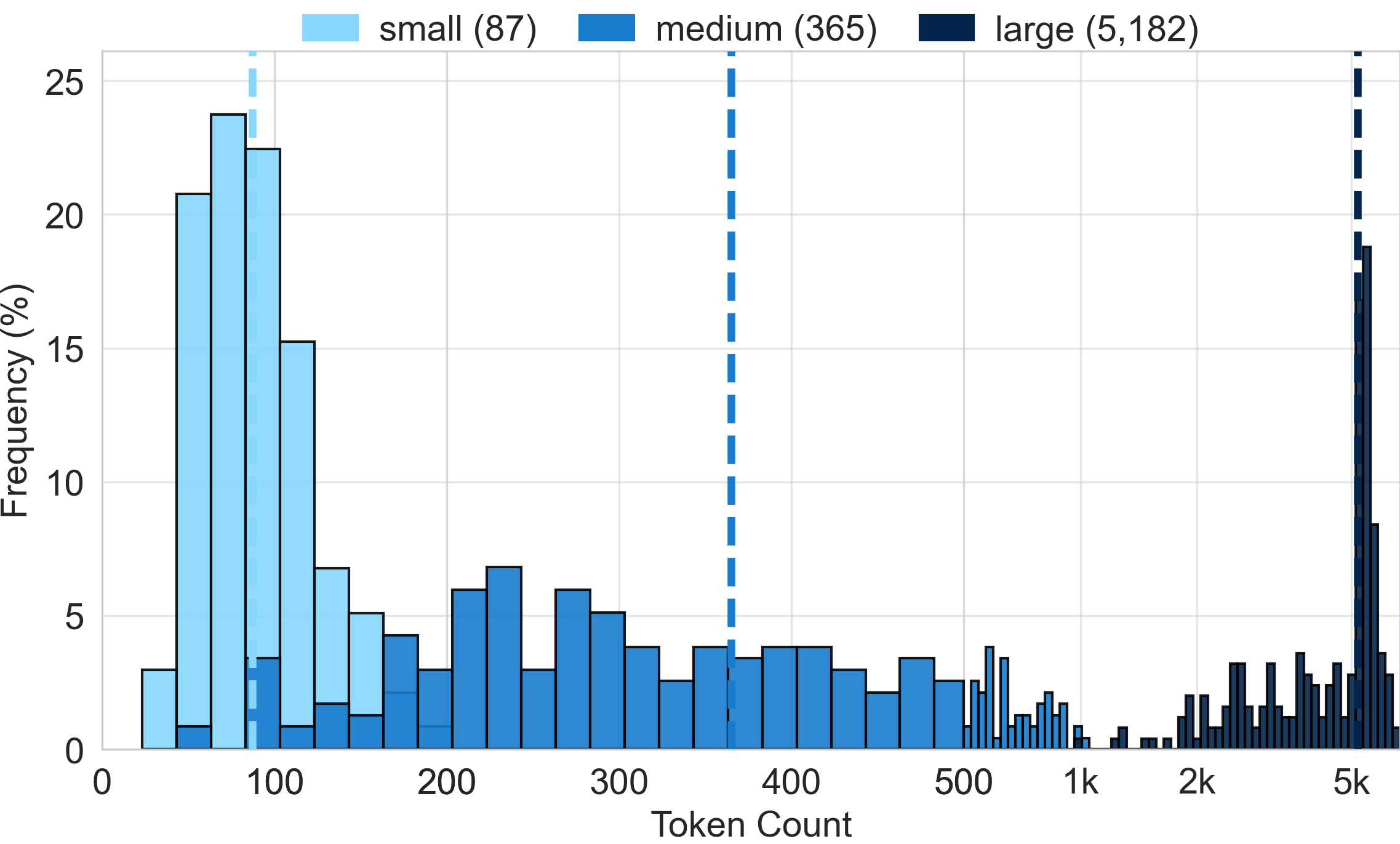}
        \caption{NuminaMath1.5}
    \end{subfigure}
    \caption{Token count distributions for varying sizes of gold context on each benchmark. Median token count is in parenthesis in the legend. X-axis is scaled as linear (0-500) and logarithmic (500+).}
    \label{fig:token-dist}
\end{figure}

We adapt three established question and answering benchmarks—CARDBiomedBench (biomedical reasoning), NaturalQuestions (general knowledge), and NuminaMath1.5 (mathematical reasoning)—to create controlled \changed{NIAH} settings. Gold context sizes were varied, accompanied by distractors explicitly designed to be topically relevant yet answer-free. Figure~\ref{fig:token-dist} displays token count distributions for the varying sizes of gold. \changed{We provide further details on datasets and their metrics in \S\ref{app:benchmarks}.} 

\textbf{CARDBiomedBench (CBB).}
CBB is a question-answering dataset focused on neurodegenerative diseases, designed to evaluate LLM performance on biomedical reasoning tasks involving genetic, molecular, and clinical information. The BiomedSQL~\citep{biomedsql_nih_card_2025} variant augments each example with SQL queries and database rows to support structured reasoning experiments:
\begin{center}
\footnotesize
\lightbluetext{Full answer} $\subset$ \mediumbluetext{SQL query + answer} $\subset$ \darkbluetext{SQL query + rows + answer}
\end{center}
Distractors were drawn from documents retrieved by four independent domain-specific agents with access to biomedical knowledge bases.
These documents are semantically relevant but verifiably do not contain the answer, presenting realistic aggregation challenges.

\textbf{NaturalQuestions (NQ).}
NQ is an open-domain QA benchmark of Google queries, with Wikipedia passages from the KILT corpus as evidence~\citep{petroni2021kiltbenchmarkknowledgeintensive}:
\begin{center}
\footnotesize
\lightbluetext{Sentence with the answer} $\subset$ \mediumbluetext{Paragraph with the sentence} $\subset$ \darkbluetext{Paragraph $\pm$ 4 paragraphs}
\end{center}
Distractors were drawn from the HELMET~\citep{yen2025helmetevaluatelongcontextlanguage} adaptation of NQ-KILT as 100-token segments. We exclude any documents labeled as evidence or containing the answer. This ensures distractors remain lexically and topically aligned with the question, but free of the answer.

\textbf{NuminaMath1.5 (NM).}
NM is the largest open-source dataset for math reasoning, with problems ranging from high school to International Mathematical Olympiad (IMO)-level difficulty, originating from diverse sources like Chinese K-12 exams, AMC/AIME contests, and global math forums. We used the OpenR1Math~\citep{openr1math220k} variant, which includes model-generated solution traces from DeepSeekR1~\citep{deepseekai2025deepseekr1incentivizingreasoningcapability} verified for correctness. We filter for examples with complete reasoning streams and define gold variants as:
\begin{center}
{
\footnotesize
\lightbluetext{Full answer}$\subset$\mediumbluetext{Textbook-style solution + answer}$\subset$\darkbluetext{Full LLM-generated chain-of-thought + solution + answer}
}
\end{center}
Distractors were reasoning traces to different questions. Due to length variability, we cap large gold contexts at the final $5k$ tokens, which include concluding reasoning and answers. 

\subsection{Baseline Experiments}
\label{subsec:baselines}

We run three baseline conditions to validate that observed performance differences in main experiments result from changes to gold size, rather than underlying flaws in datasets or distractor construction. \textbf{Baseline results across all benchmarks and models can be found in Appendix~\ref{apdx:baselines}}:
\begin{itemize} [itemsep=0pt, parsep=0pt, topsep=0pt]
    \item \textbf{Closed-book.} No context is provided, assessing whether models could answer from internal knowledge. This gauges possible benchmark saturation.
    
    \item \textbf{Gold-only.} Each gold context is presented alone, without distractors. This confirms variants were sufficient to solve the task and that downstream performance drops are due to aggregation effects (e.g., distractor interference or gold placement).
    
    \item \textbf{Distractor-only.} Models are given only distractor documents. For CBB, we also test distractors from each agent separately to confirm they were individually non-informative. These checks ensure that distractors lack sufficient signal to answer correctly.
\end{itemize}

\subsection{Main Experiments}

We simulate realistic aggregation scenarios by embedding each gold context size at varying positions within a fixed sequence of distractors. This tests both gold size and positional sensitivity simultaneously. We evaluate eleven leading LLMs:

\begin{itemize} [itemsep=0pt, parsep=0pt, topsep=0pt]
\item \textbf{Closed-weight}: o3-mini, GPT-4o~\citep{openai2024gpt4ocard}, GPT-4o-Mini~\citep{openai2024gpt4omini}, Gemini-2.5-Flash, Gemini-2.0-Flash, and Gemini-2.0-Flash-Lite~\citep{gemini2025flash}
\item \textbf{Open-weight}: DeepSeek-R1~\citep{deepseekai2025deepseekr1incentivizingreasoningcapability}, Phi-4-reasoning~\citep{abdin2025phi4reasoningtechnicalreport}, LLaMA-3.1-405B, LLaMA-3.3-70B, and LLaMA-3.1-8B~\citep{dubey2024llama3herdmodels}
\end{itemize}

We evaluate each model on every size-position combination in a deterministic setting. Prompts were standardized within each benchmark. This enables rigorous, cross-model evaluation of gold context sensitivity and simulates the type of unpredictable document ordering in LLM systems.

\begin{figure}[th]
    \centering

    \begin{subfigure}{0.329\textwidth}
        \centering
        \includegraphics[width=\linewidth,trim=0.0cm 0cm 0cm 0cm,clip=true]{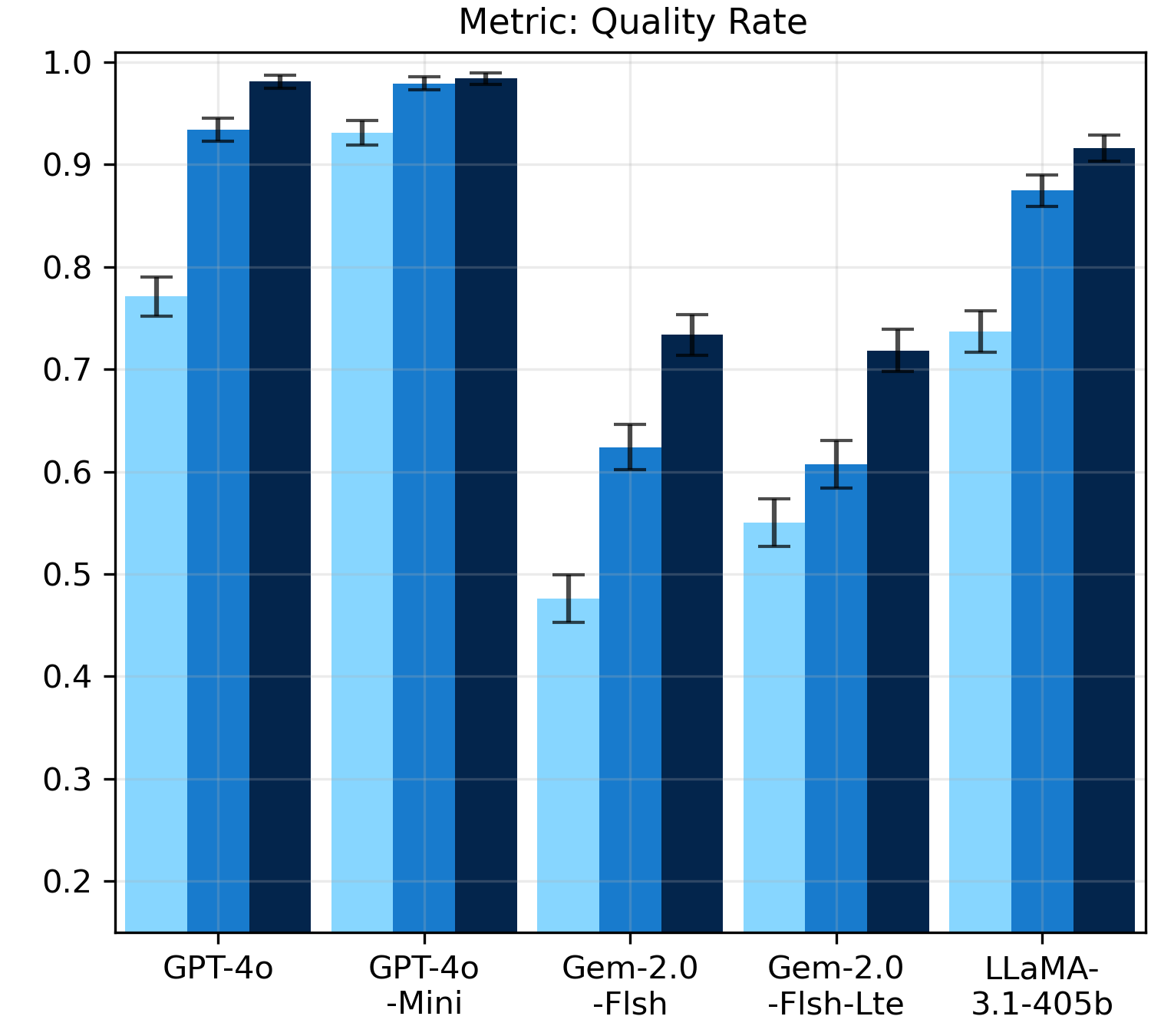}
    \end{subfigure}
    \hfill
    \begin{subfigure}{0.329\textwidth}
        \centering
        \includegraphics[width=\linewidth,trim=0.0cm 0cm 0cm 0cm,clip=true]{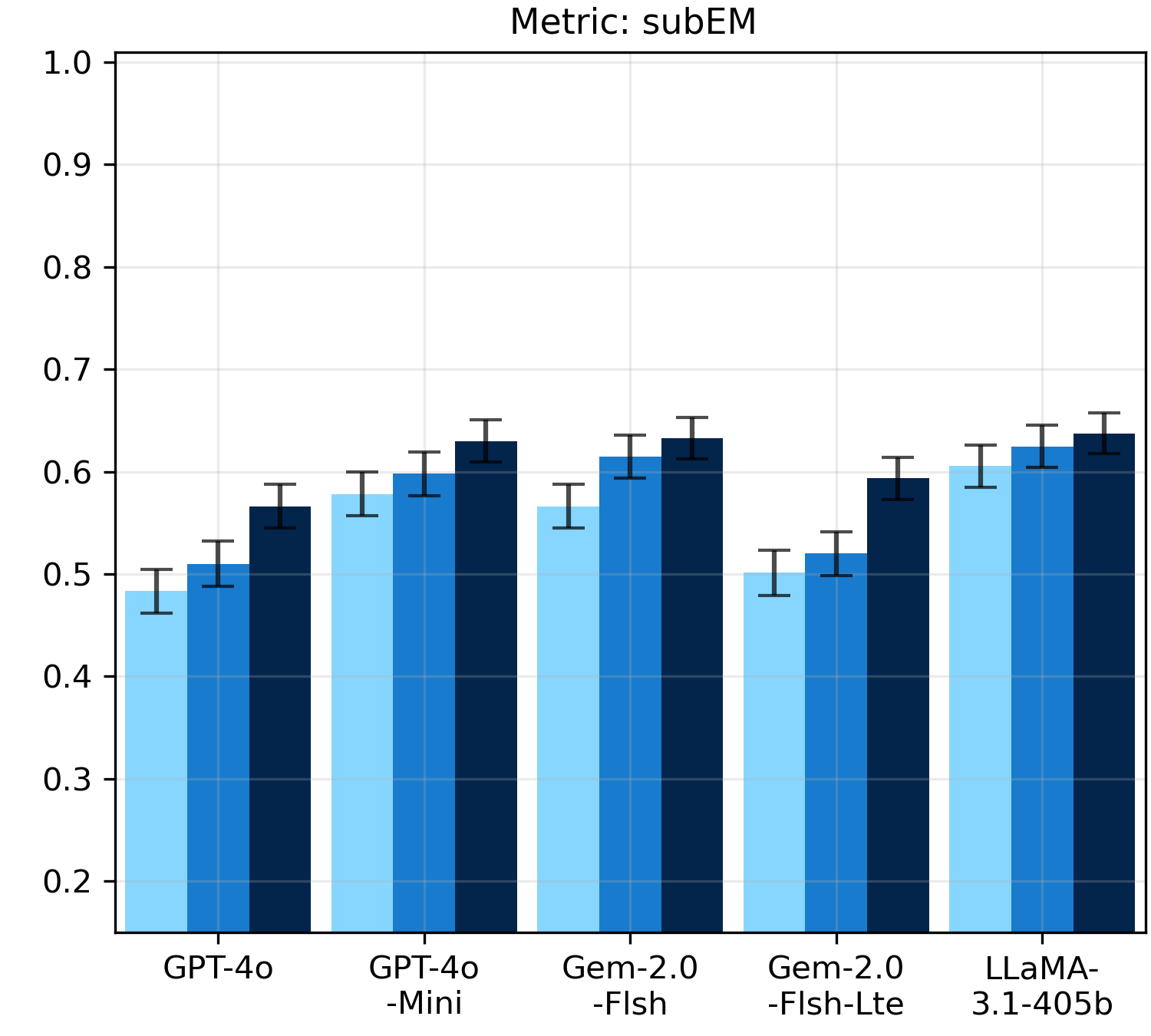}
    \end{subfigure}
    \hfill
    \begin{subfigure}{0.329\textwidth}
        \centering
        \includegraphics[width=\linewidth,trim=0.0cm 0cm 0cm 0cm,clip=true]{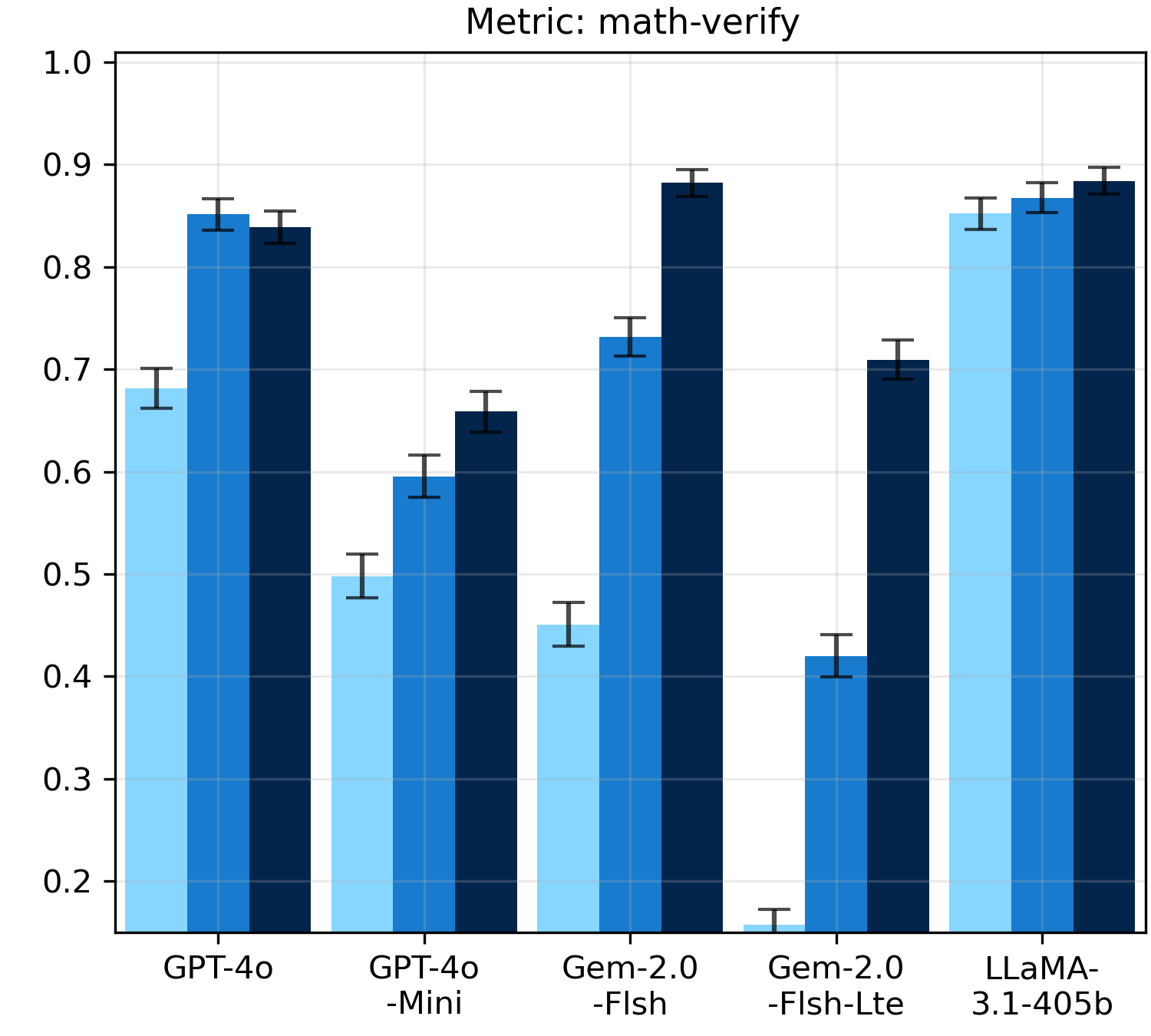}
    \end{subfigure}

    \vspace{0.5em}

    \begin{subfigure}{0.329\textwidth}
        \centering
        \includegraphics[width=\linewidth,trim=0.0cm 0cm 0cm 0cm,clip=true]{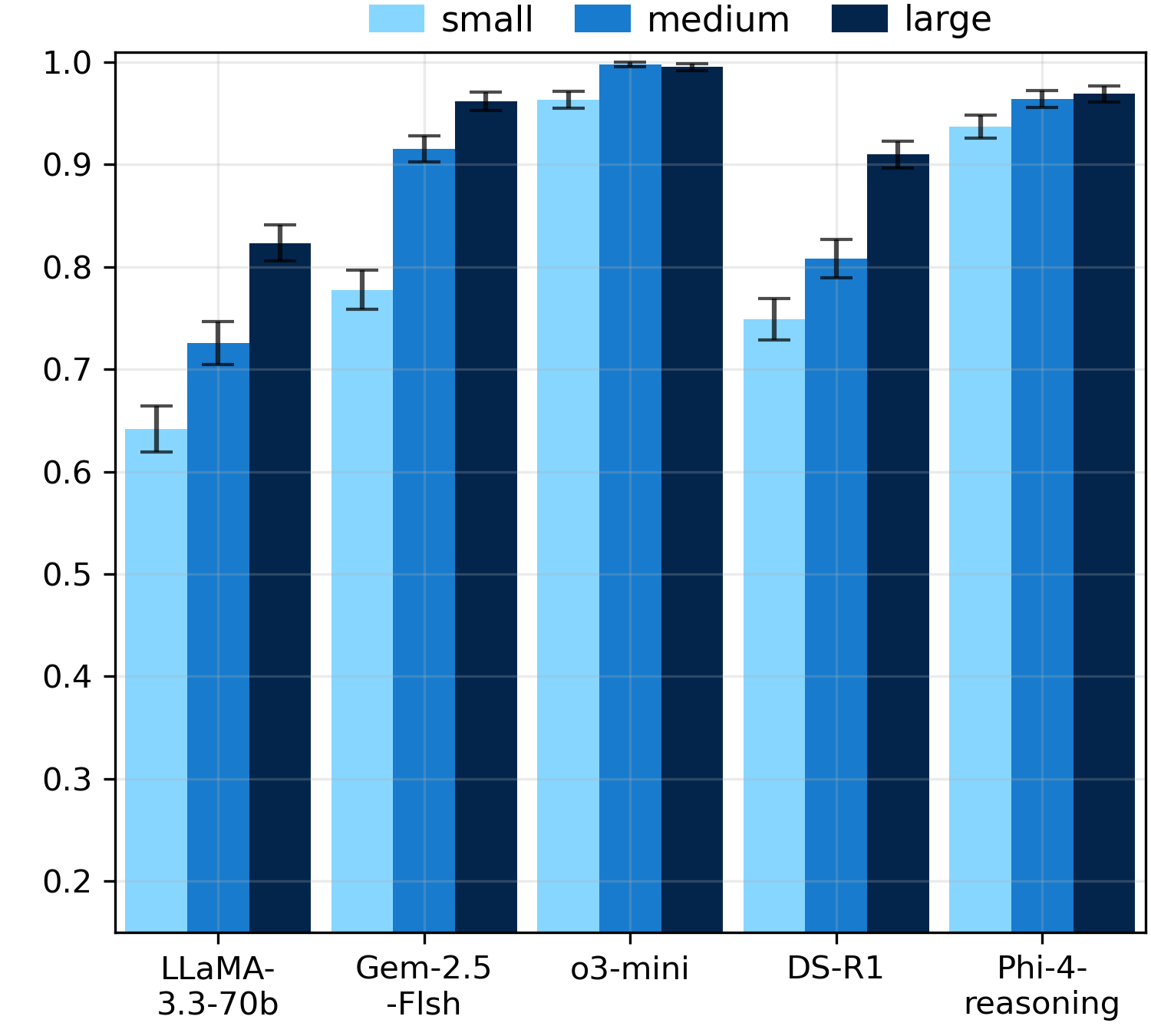}
        \caption{CARDBiomedBench}
    \end{subfigure}
    \hfill
    \begin{subfigure}{0.329\textwidth}
        \centering
        \includegraphics[width=\linewidth,trim=0.0cm 0cm 0cm 0cm,clip=true]{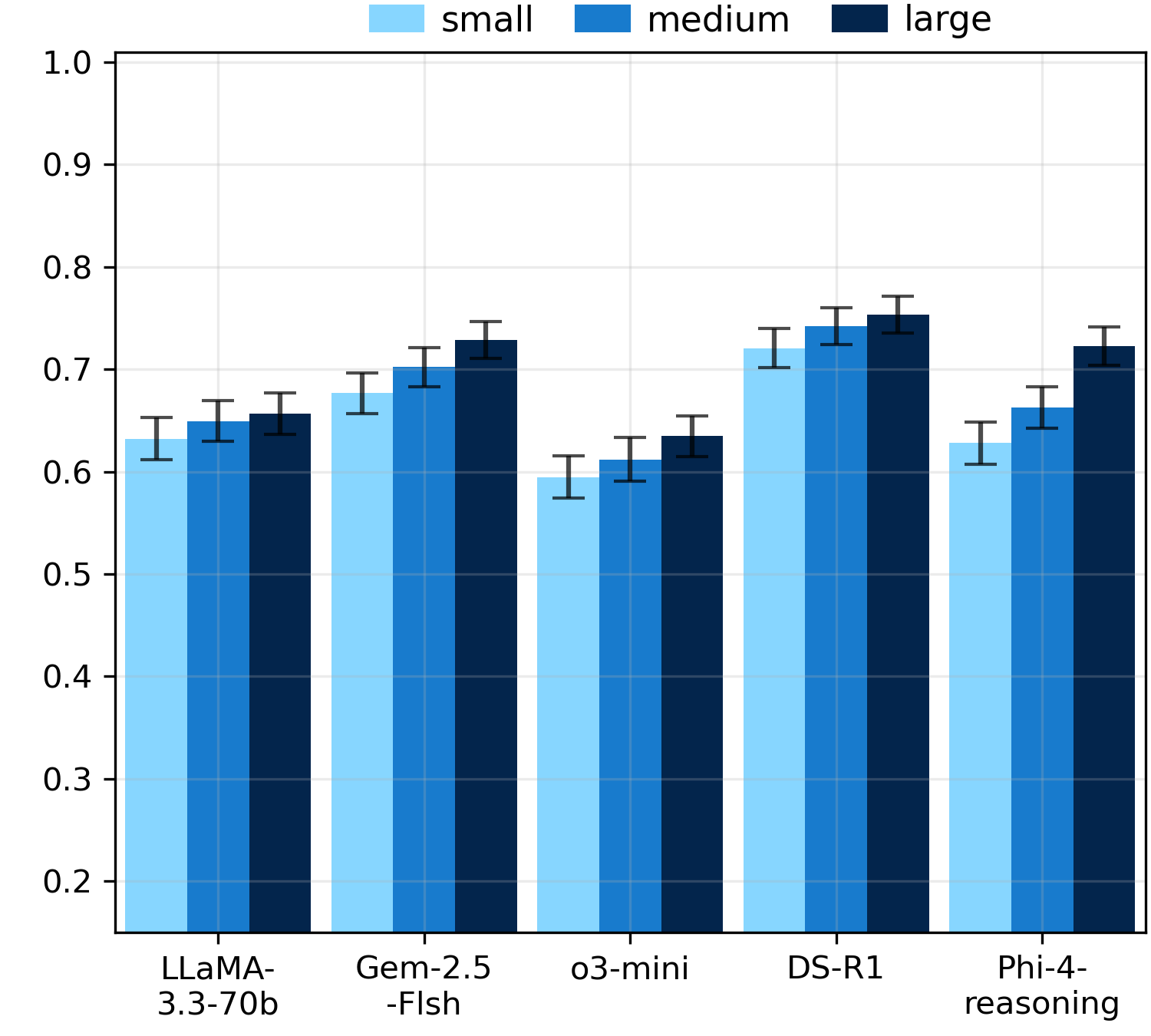}
        \caption{NaturalQuestions}
    \end{subfigure}
    \hfill
    \begin{subfigure}{0.329\textwidth}
        \centering
        \includegraphics[width=\linewidth,trim=0.0cm 0cm 0cm 0cm,clip=true]{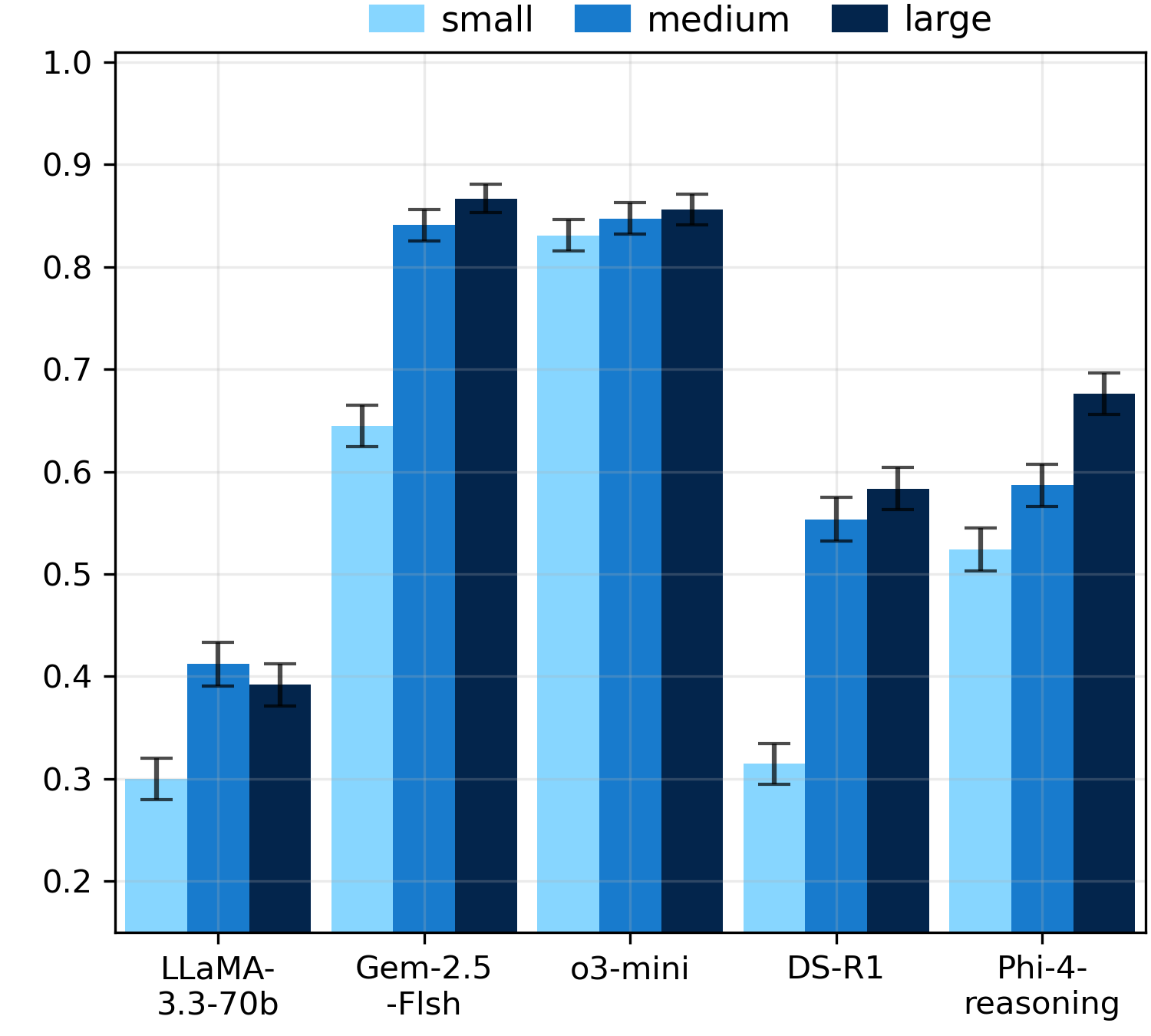}
        \caption{NuminaMath1.5}
    \end{subfigure}

    \caption{    Average performance across all gold positions for each benchmark and gold context size. Metrics are benchmark-specific (BioScore, subEM, math-verify). Higher is better.  Error bars indicate 90\% confidence intervals. Colors correspond to gold context sizes: \lightbluetext{small}, \mediumbluetext{medium}, \darkbluetext{large}.
    \textbf{Across all settings, performance improves monotonically with gold context size.}}
    \label{fig:two_by_three_grid}
\end{figure}

\section{Main Finding: Smaller Gold Contexts \changed{Lower the Performance}}
\label{sec:findings}

Our experiments reveal that gold context size has a substantial and consistent effect on long-context performance, irrespective of confounding variables, across different benchmarks and models.

Increasing the size of the gold context significantly improves accuracy (Figure~\ref{fig:two_by_three_grid}). On CBB, Gemini-2.0-Flash went from 48\% with small to 62\% with medium and 73\% with large. GPT-4o performs similarly, rising from 77\% (small) to 98\% (large), while LLaMA-3.1-405B went from 74\% to 92\%.

Notably, performance with large gold contexts approaches the \textit{gold-only baselines} (i.e., accuracy when the gold context is shown without any distractors) recorded at 96\% for Gemini-2.0-Flash, and 100\% for both GPT-4o and LLaMA-3.1-405B. This suggests that large gold contexts allow models to nearly recover ideal aggregation performance, while small golds fall significantly short.

\section{Analysis of Confounding Factors}
\label{sec:confounding:factors}

Our goal is to isolate the effect of \textit{gold context size} as an independent factor in LLM performance. 
\changed{However, there are various other factors that confounded or correlated with our target variable, making the attribution of the observed effect non-trivial. 
}
\changed{To better understand the underlying phenomenon here}, one must \changed{control for} 
any potential confounding factors that may impact the outcome of the findings. Specifically, in this section we study and \changed{control} 
the following confounds: gold context position in the context window (\S\ref{subsec:gold:doc:pos}), the total number of repetitions of the answer in the context (\S\ref{subsec:answer_repetition}),  relative length of gold context to the total distractor evidence length (\S\ref{subsec:gold-to-distractor}), total distractor length (\S\ref{subsec:distractor:volume}) and domain specificity (\S\ref{subsec:domain:specificity}). 


\begin{figure}[th]
    \centering
    \tiny
    \setlength{\tabcolsep}{2pt}
    \begin{tabular}{cccc}
         \lightpurpletext{CARDBiomedBench} & \lightpurpletext{NaturalQuestions} & \lightpurpletext{NuminaMath1.5} & 
 \\
    \begin{subfigure}{0.329\textwidth}
        \includegraphics[width=0.9\linewidth]{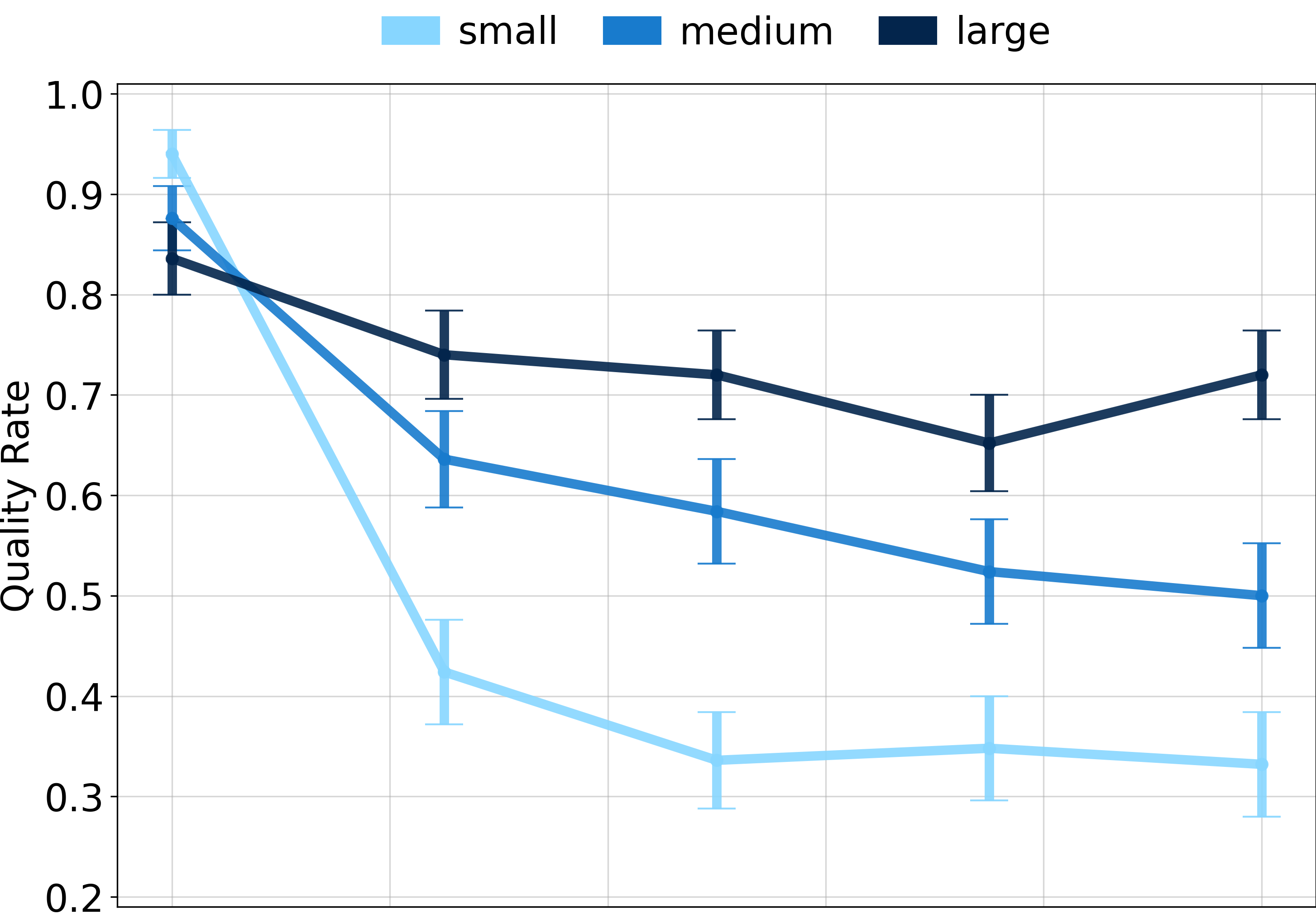}
    \end{subfigure} 
    & 
    \begin{subfigure}{0.329\textwidth}
        \includegraphics[width=0.9\linewidth]{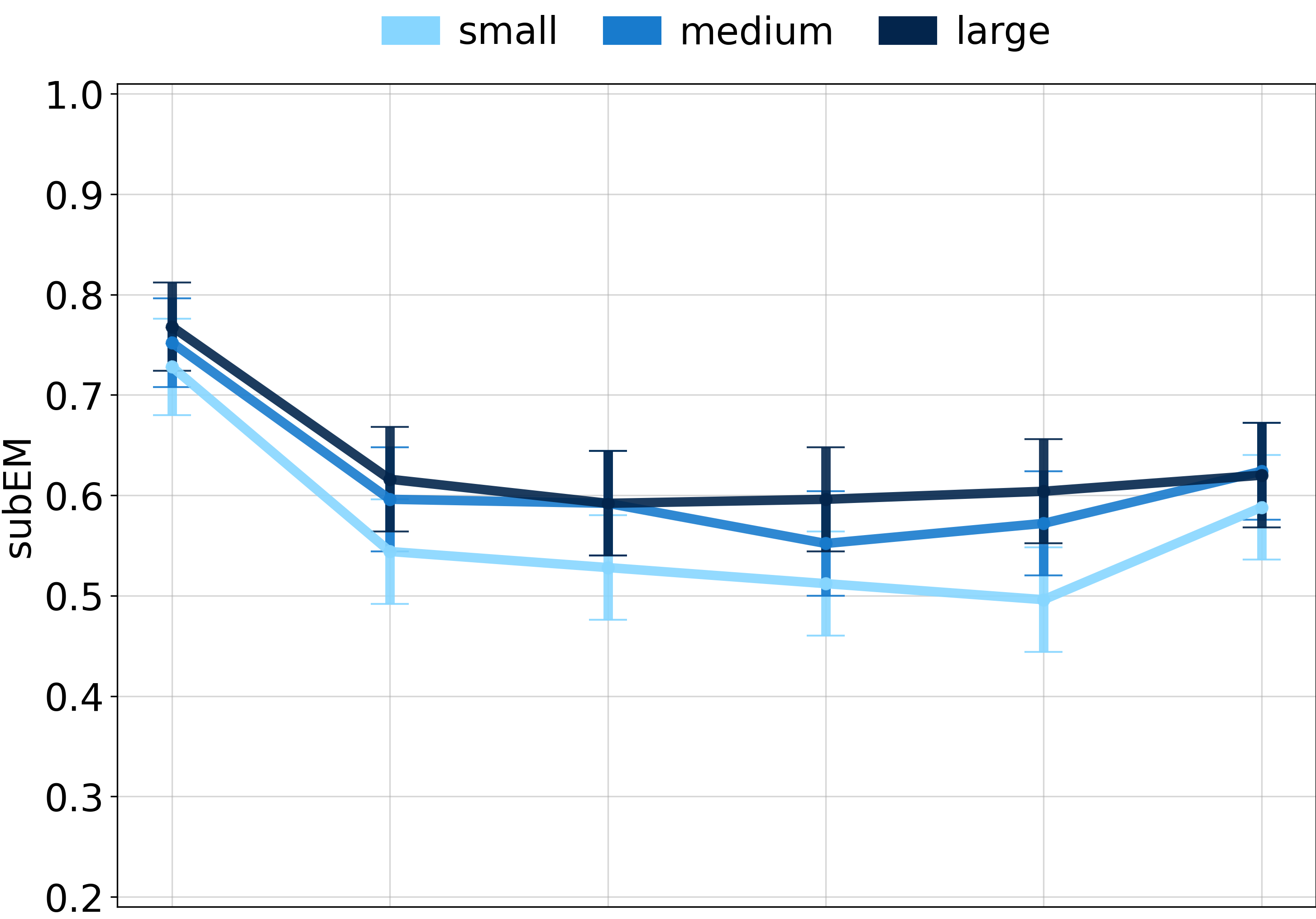}
    \end{subfigure}
    & 
    \begin{subfigure}{0.329\textwidth}
        \includegraphics[width=0.9\linewidth]{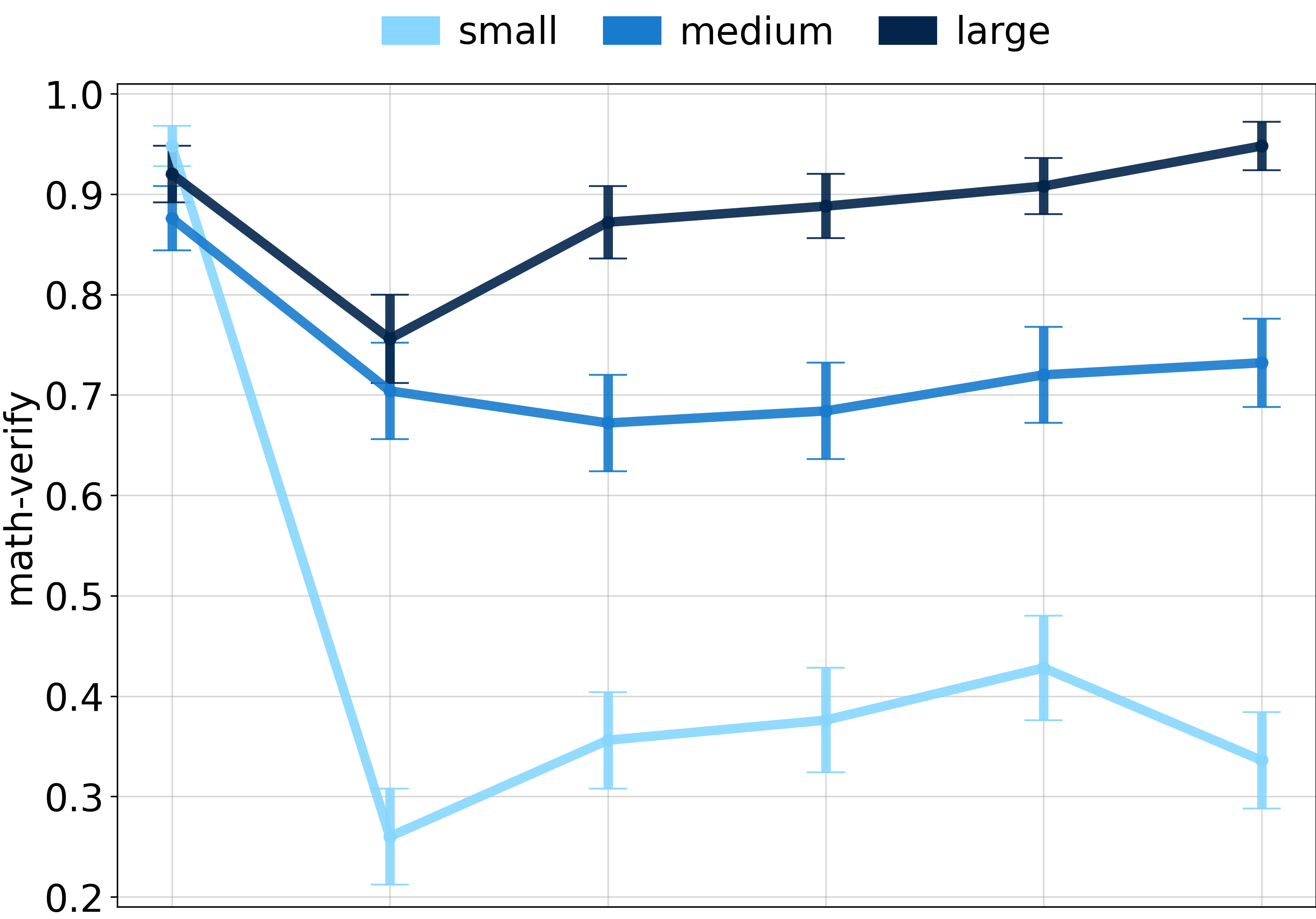}
    \end{subfigure}
    & 
    \hspace{-0.8cm} \raisebox{13ex}{\rotatebox[origin=c]{270}{\lightpurpletext{Gemini-2.0-Flash}}} 
    \\  

    \begin{subfigure}{0.329\textwidth}
        \includegraphics[width=0.9\linewidth]{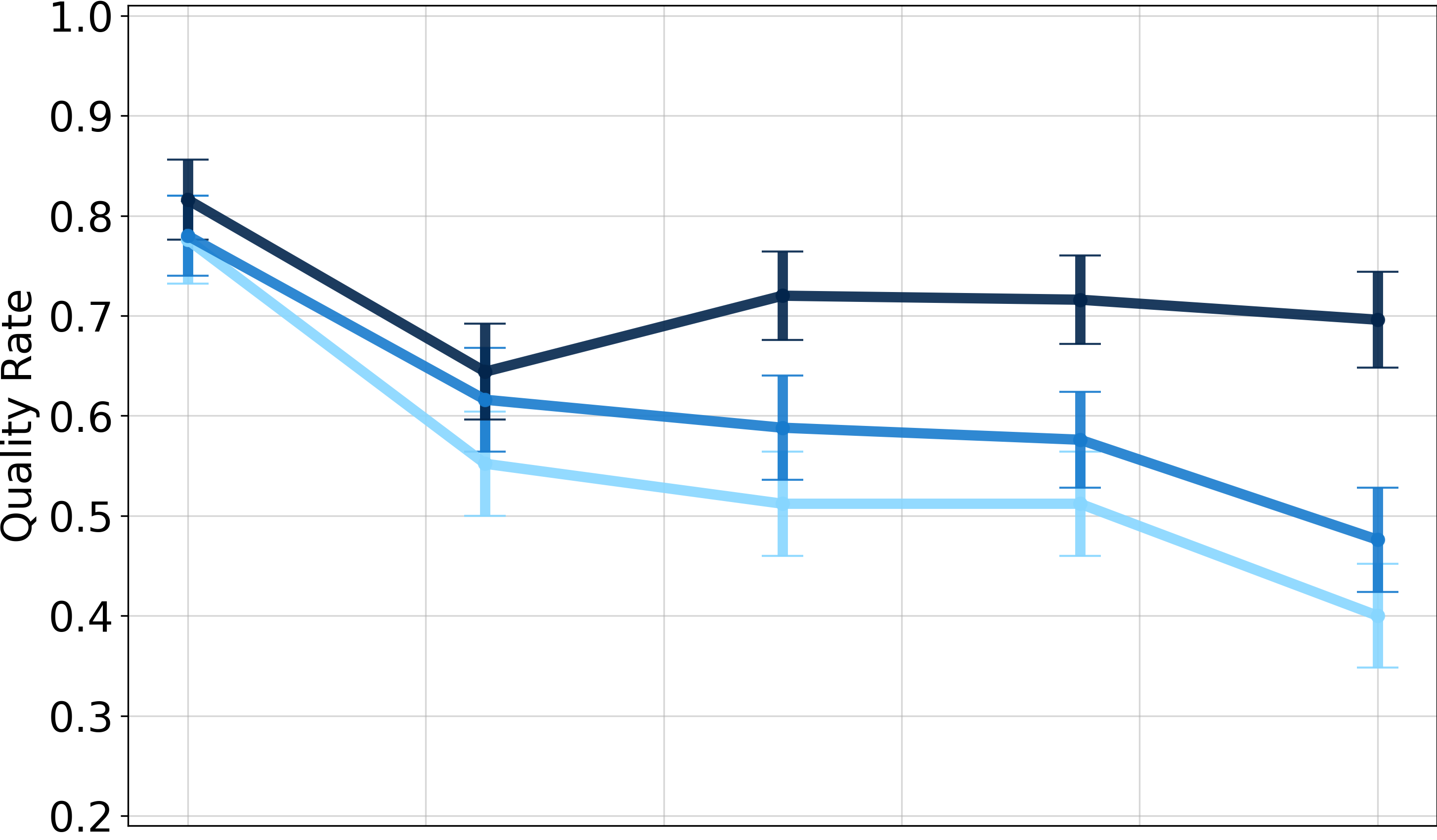}
    \end{subfigure} 
    & 
    \begin{subfigure}{0.329\textwidth}
        \includegraphics[width=0.9\linewidth]{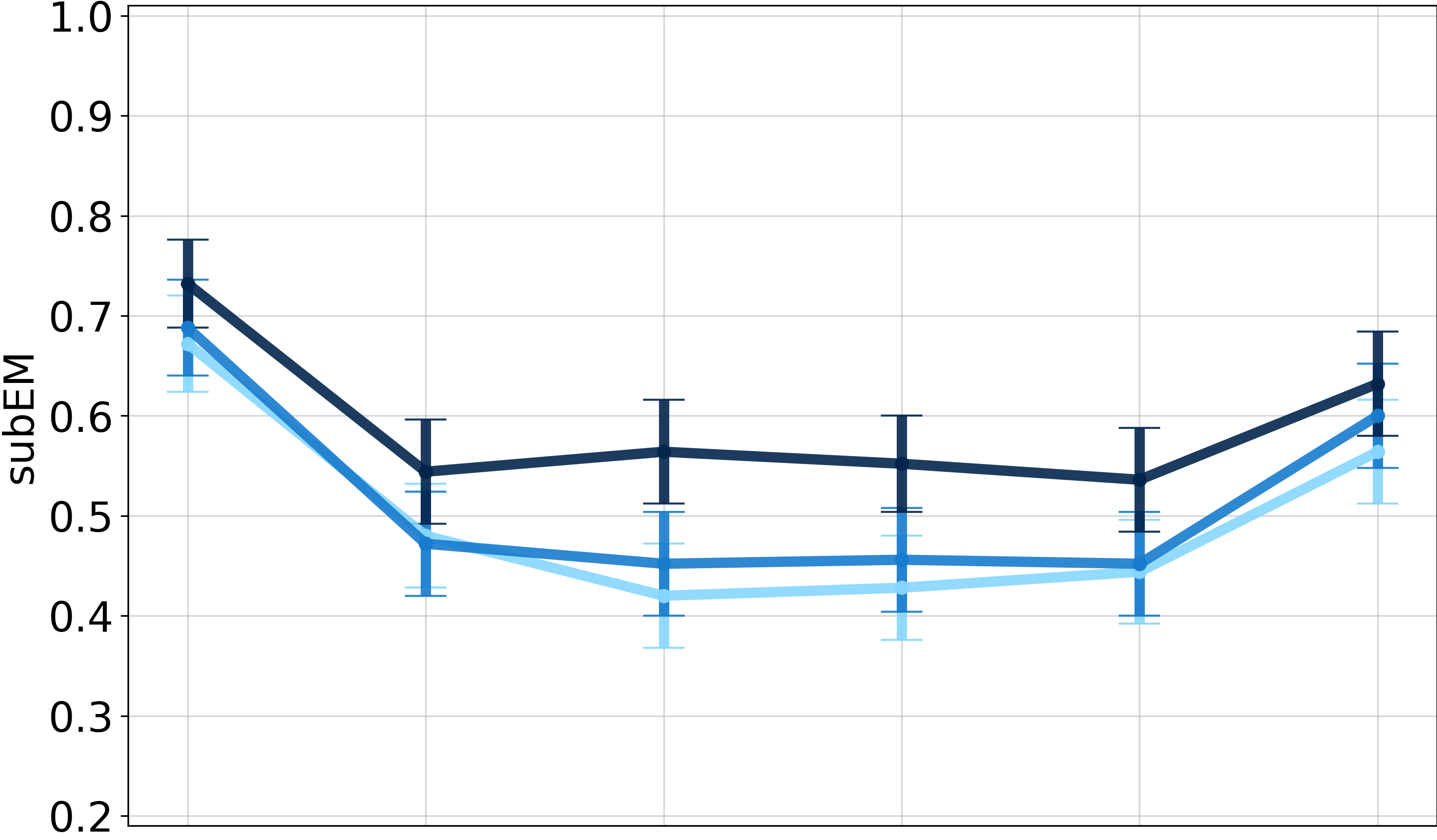}
    \end{subfigure}
    & 
    \begin{subfigure}{0.329\textwidth}
        \includegraphics[width=0.9\linewidth]{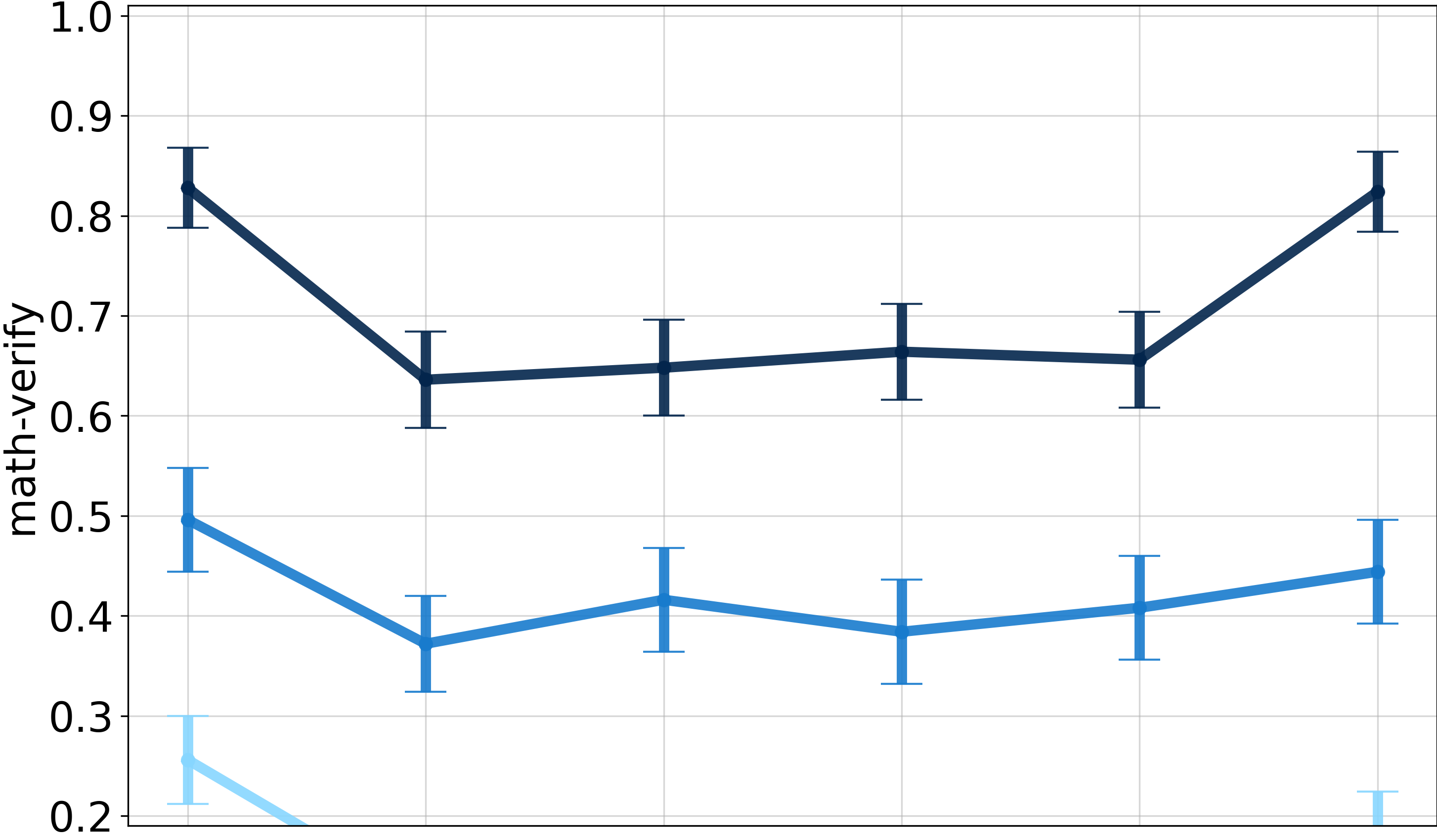}
    \end{subfigure}
    & 
    \hspace{-0.8cm} \raisebox{13ex}{\rotatebox[origin=c]{270}{\lightpurpletext{Gemini-2.0-Flash-Lite}}} 
    \\
    
    \begin{subfigure}{0.329\textwidth}
        \includegraphics[width=0.9\linewidth]{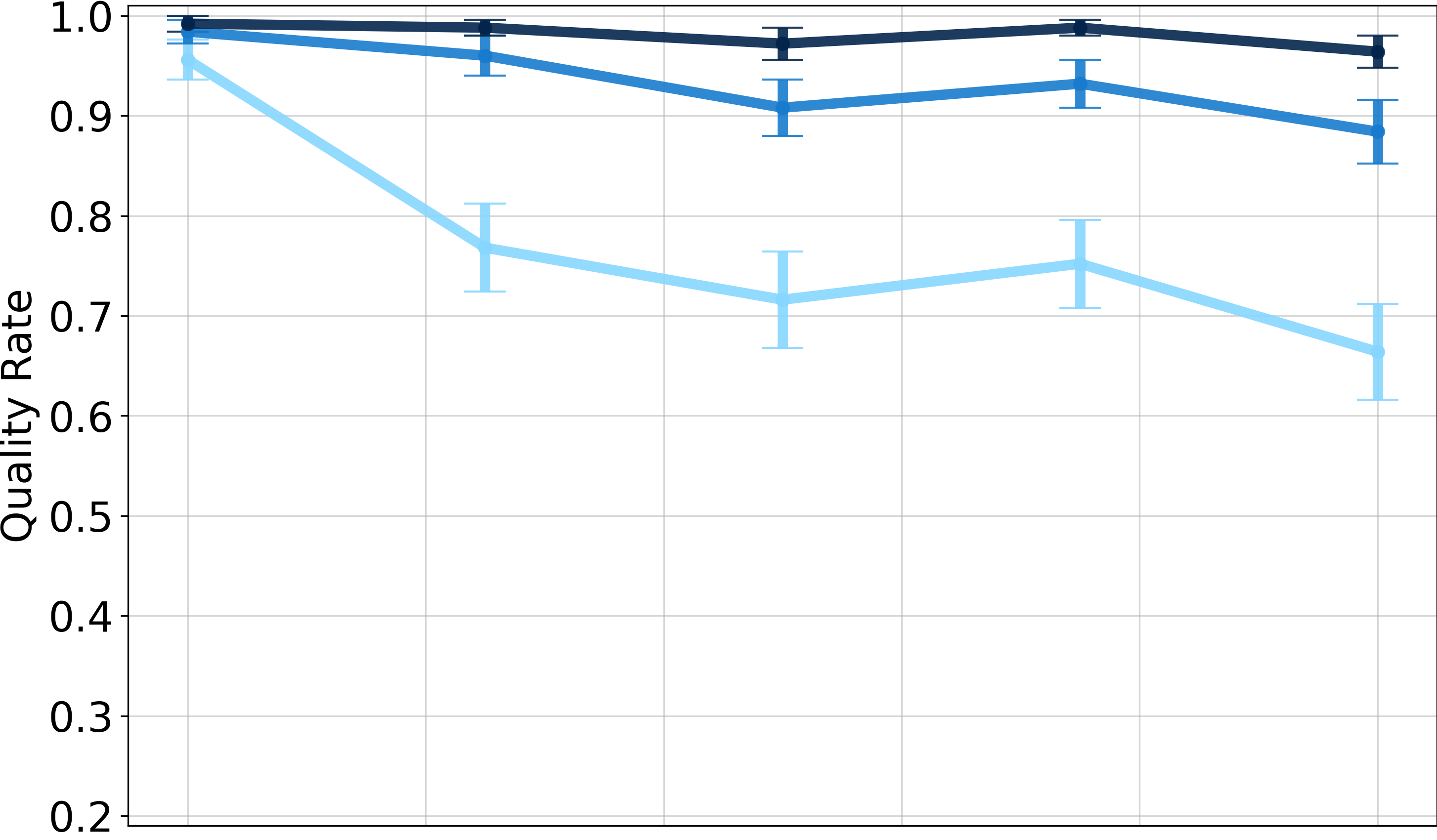}
    \end{subfigure}     
     & 
    \begin{subfigure}{0.329\textwidth}
        \includegraphics[width=0.9\linewidth]{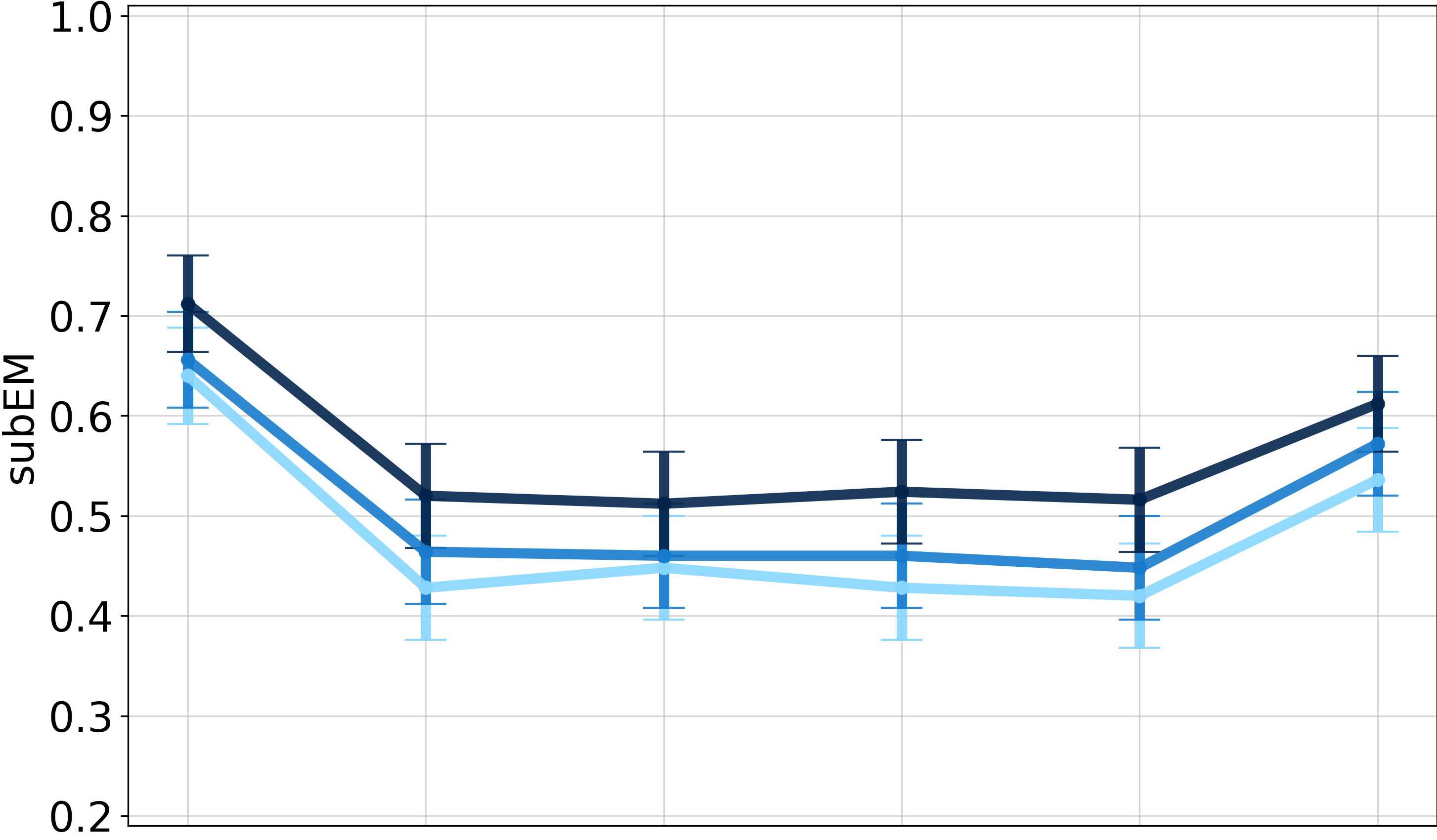}
    \end{subfigure}
     & 
    \begin{subfigure}{0.329\textwidth}
        \includegraphics[width=0.9\linewidth]{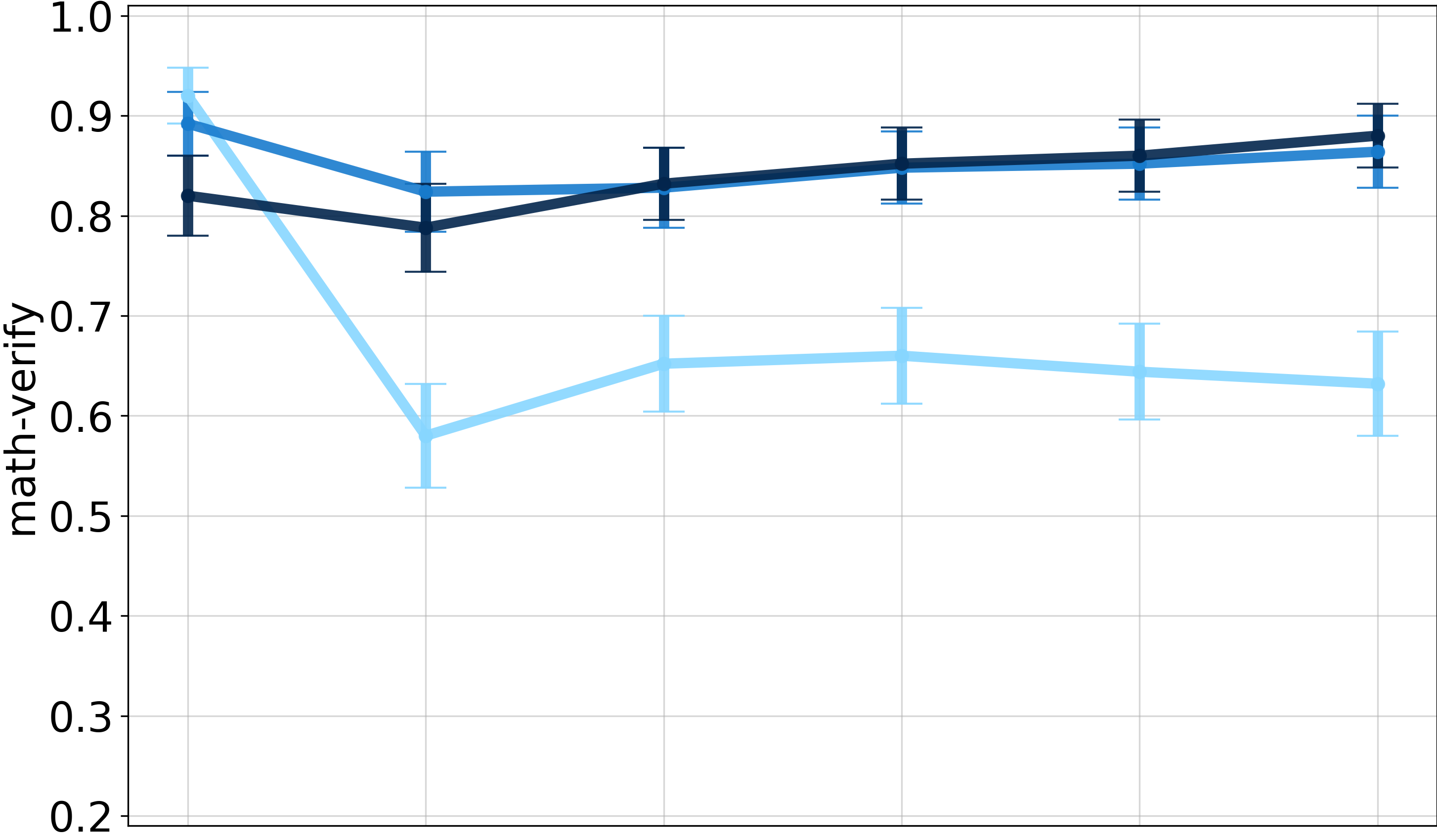}
    \end{subfigure}
    & \hspace{-0.8cm} \raisebox{9ex}{\rotatebox[origin=r]{270}{\lightpurpletext{GPT-4o}}} 
     \\ 
    \begin{subfigure}{0.329\textwidth}
        \includegraphics[width=0.9\linewidth]{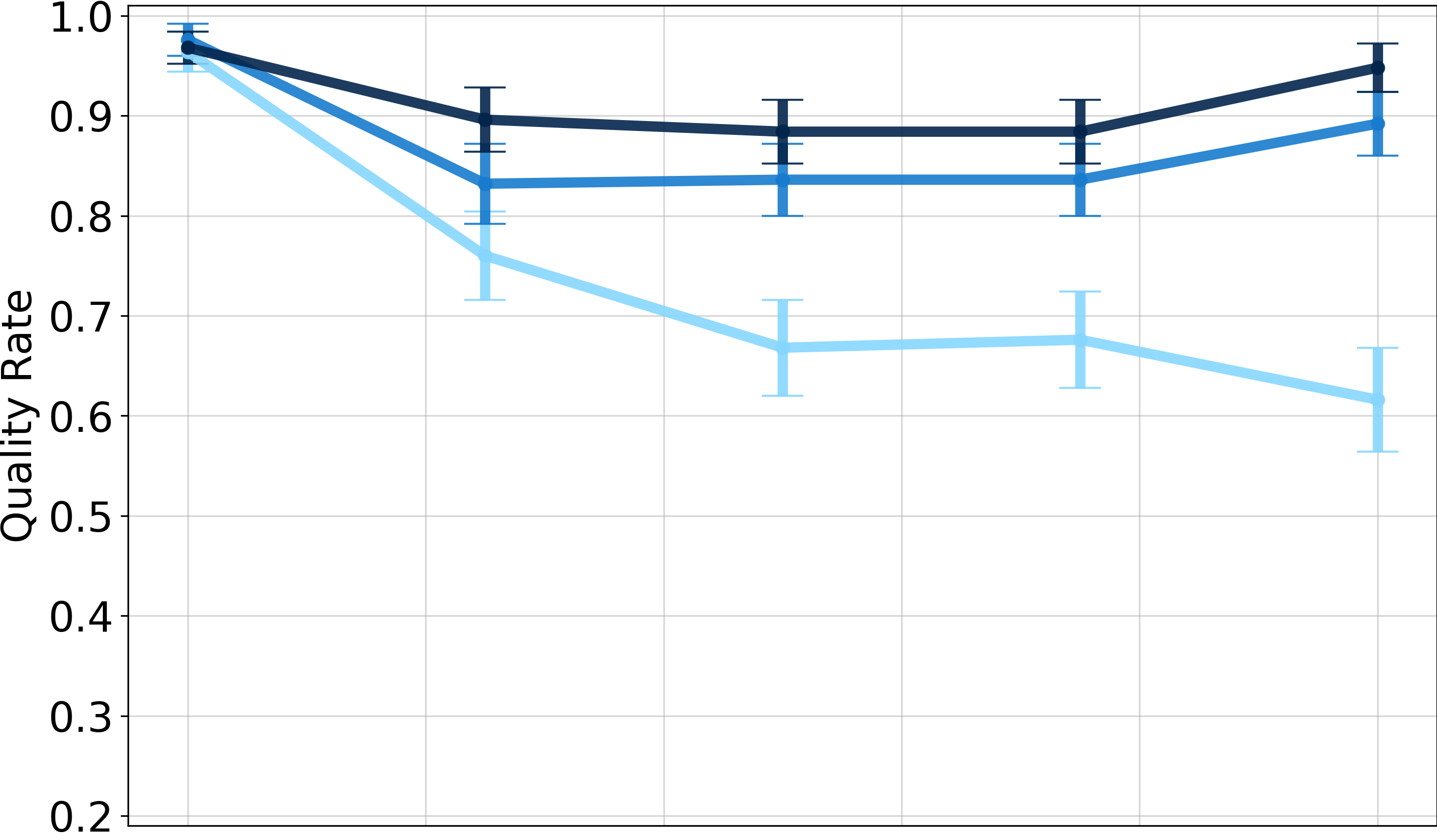}
    \end{subfigure}
    & 
    \begin{subfigure}{0.329\textwidth}
        \includegraphics[width=0.9\linewidth]{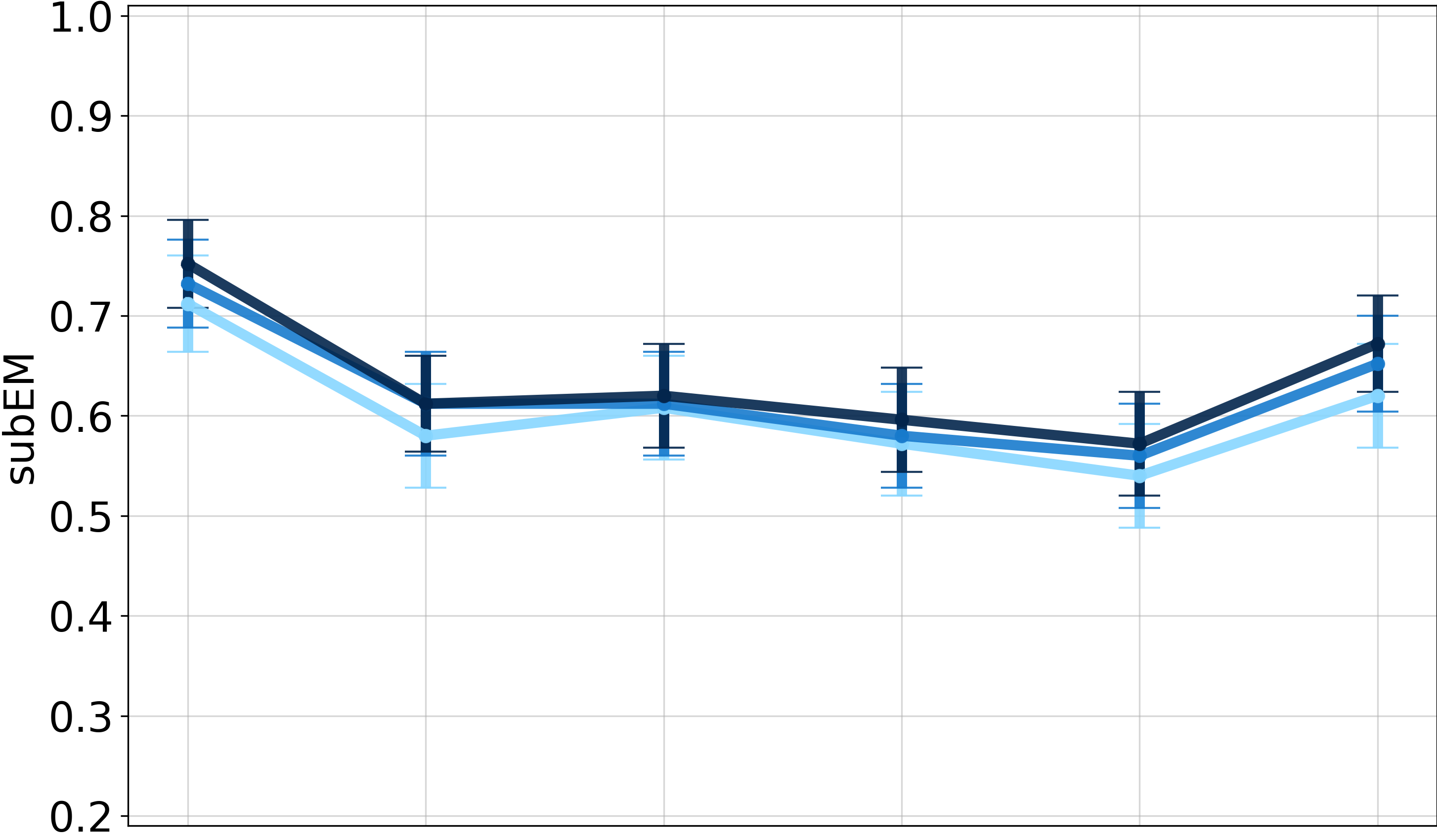}
    \end{subfigure}
    &  
    \begin{subfigure}{0.329\textwidth}
        \includegraphics[width=0.9\linewidth]{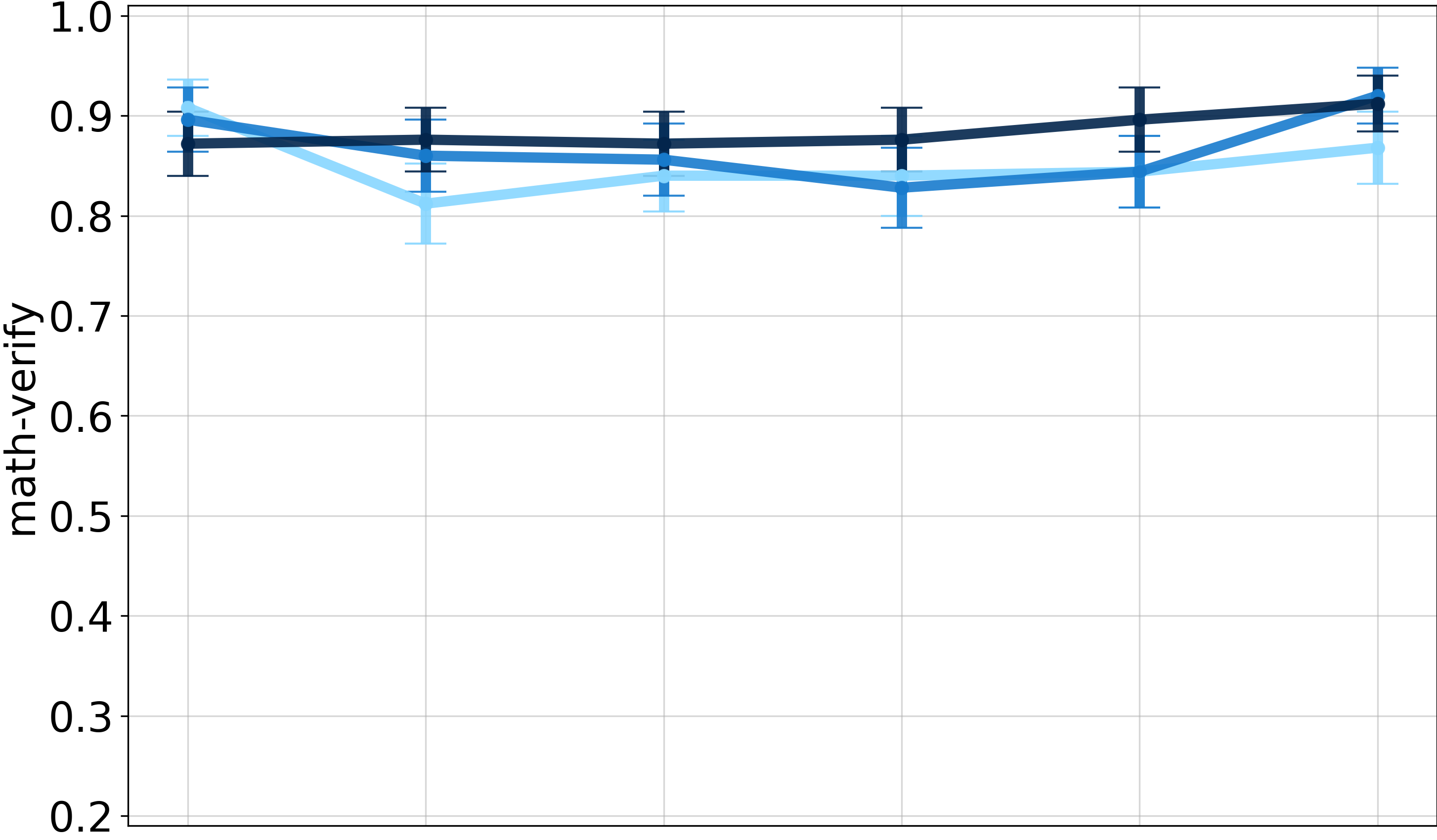}
    \end{subfigure}
    & \hspace{-0.8cm} \raisebox{3ex}{\rotatebox[origin=r]{270}{\lightpurpletext{LLaMA-3.1-405B}}}   
    \\  
    \begin{subfigure}{0.329\textwidth}
        \includegraphics[width=0.9\linewidth]{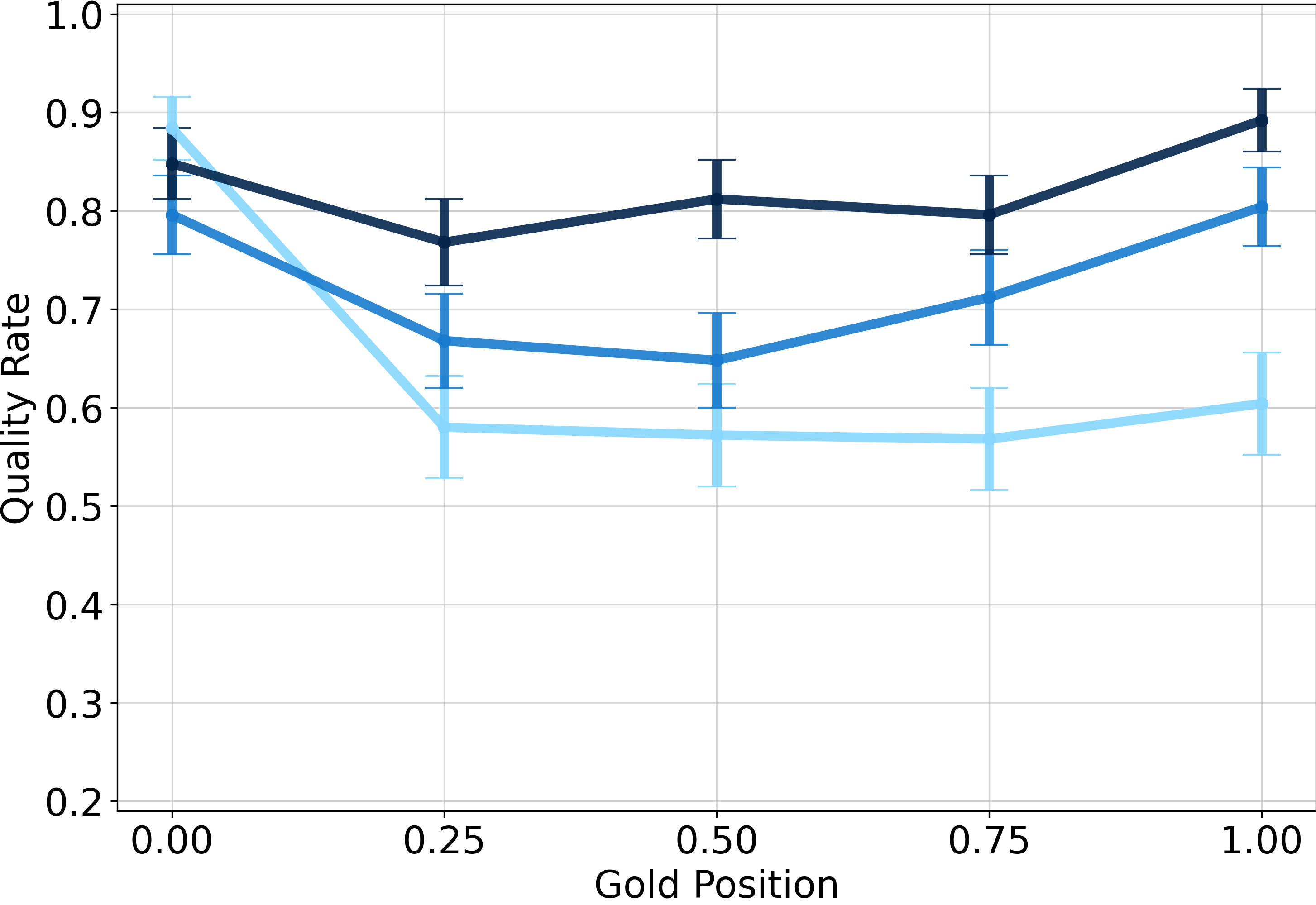}
    \end{subfigure}
    & 
    \begin{subfigure}{0.329\textwidth}
        \includegraphics[width=0.9\linewidth]{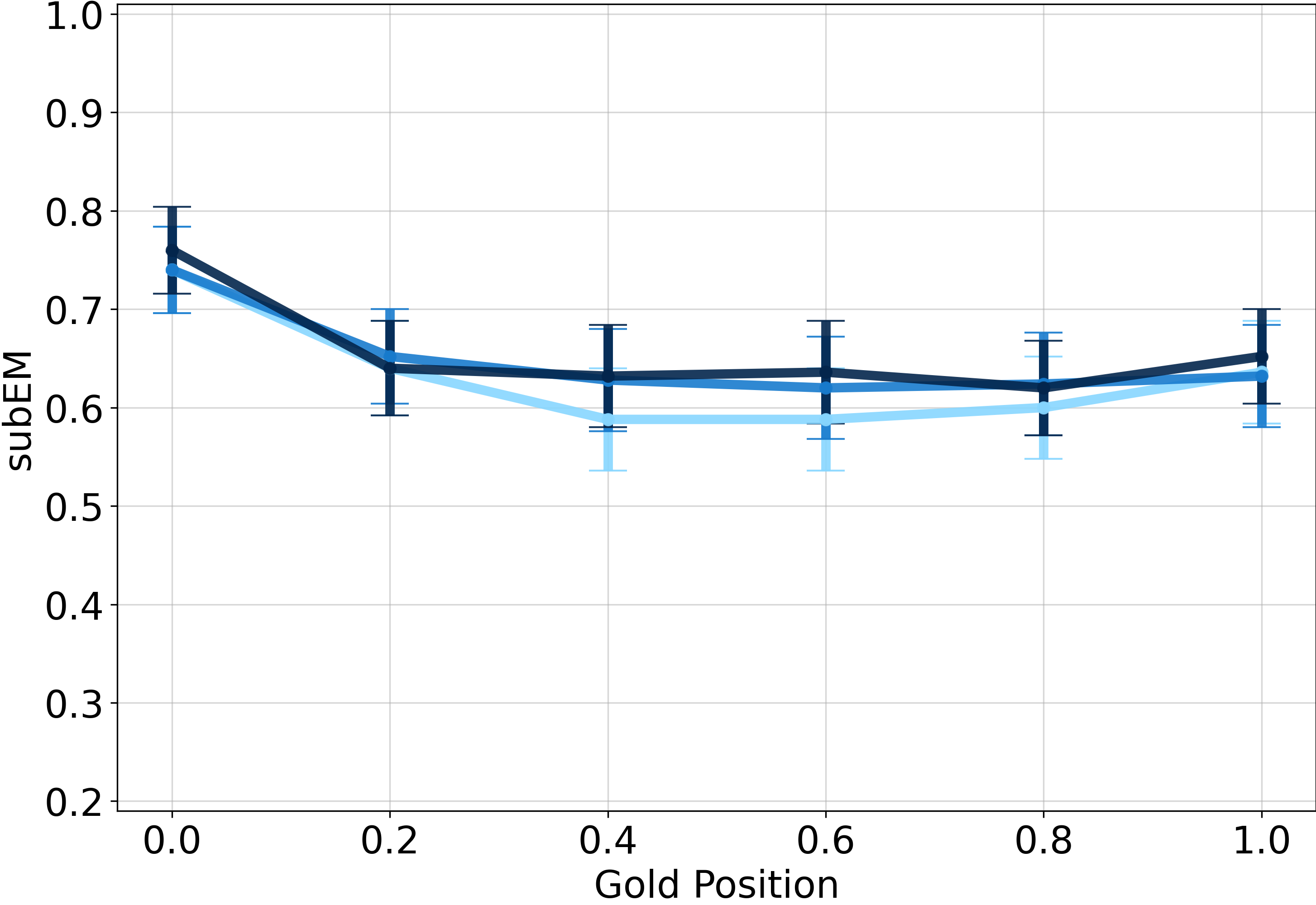}
    \end{subfigure}
    & 
    \begin{subfigure}{0.329\textwidth}
        \includegraphics[width=0.9\linewidth]{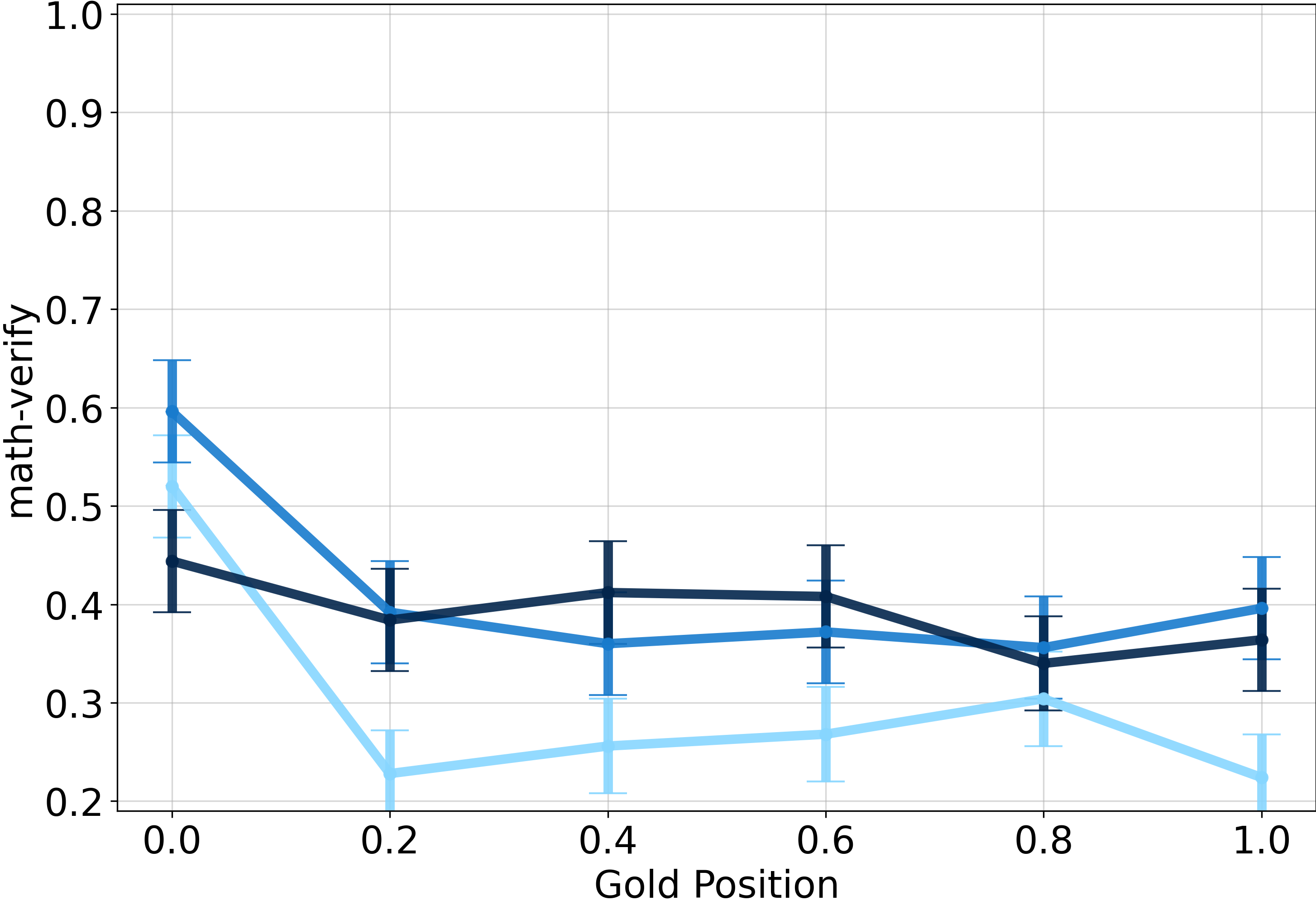}
    \end{subfigure}
    & 
    \hspace{-0.8cm} \raisebox{7ex}{\rotatebox[origin=r]{270}{\lightpurpletext{LLaMA-3.3-70B}}}
    \\

    \begin{subfigure}{0.329\textwidth}
        \includegraphics[width=0.9\linewidth]{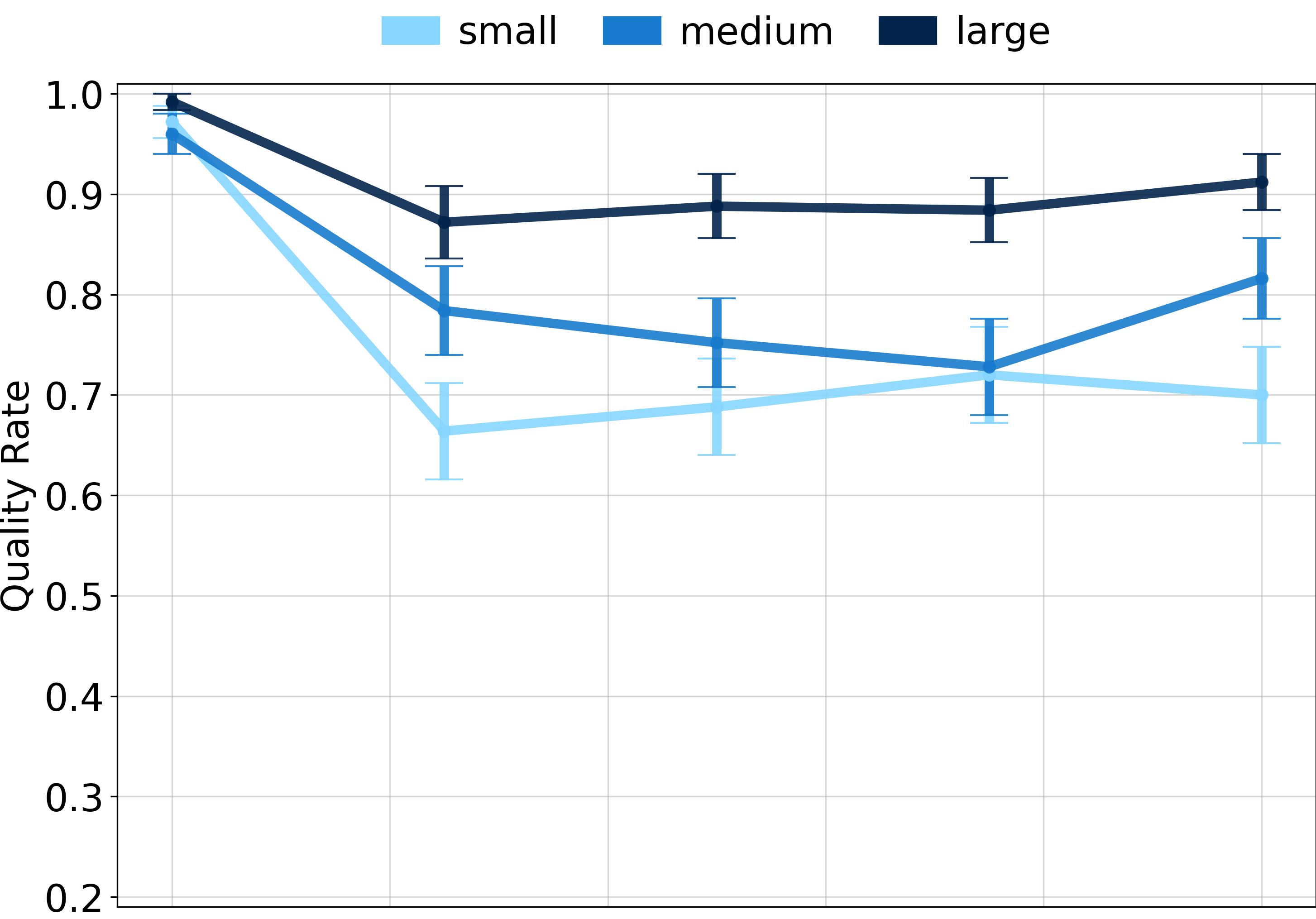}
    \end{subfigure} 
    & 
    \begin{subfigure}{0.329\textwidth}
        \includegraphics[width=0.9\linewidth]{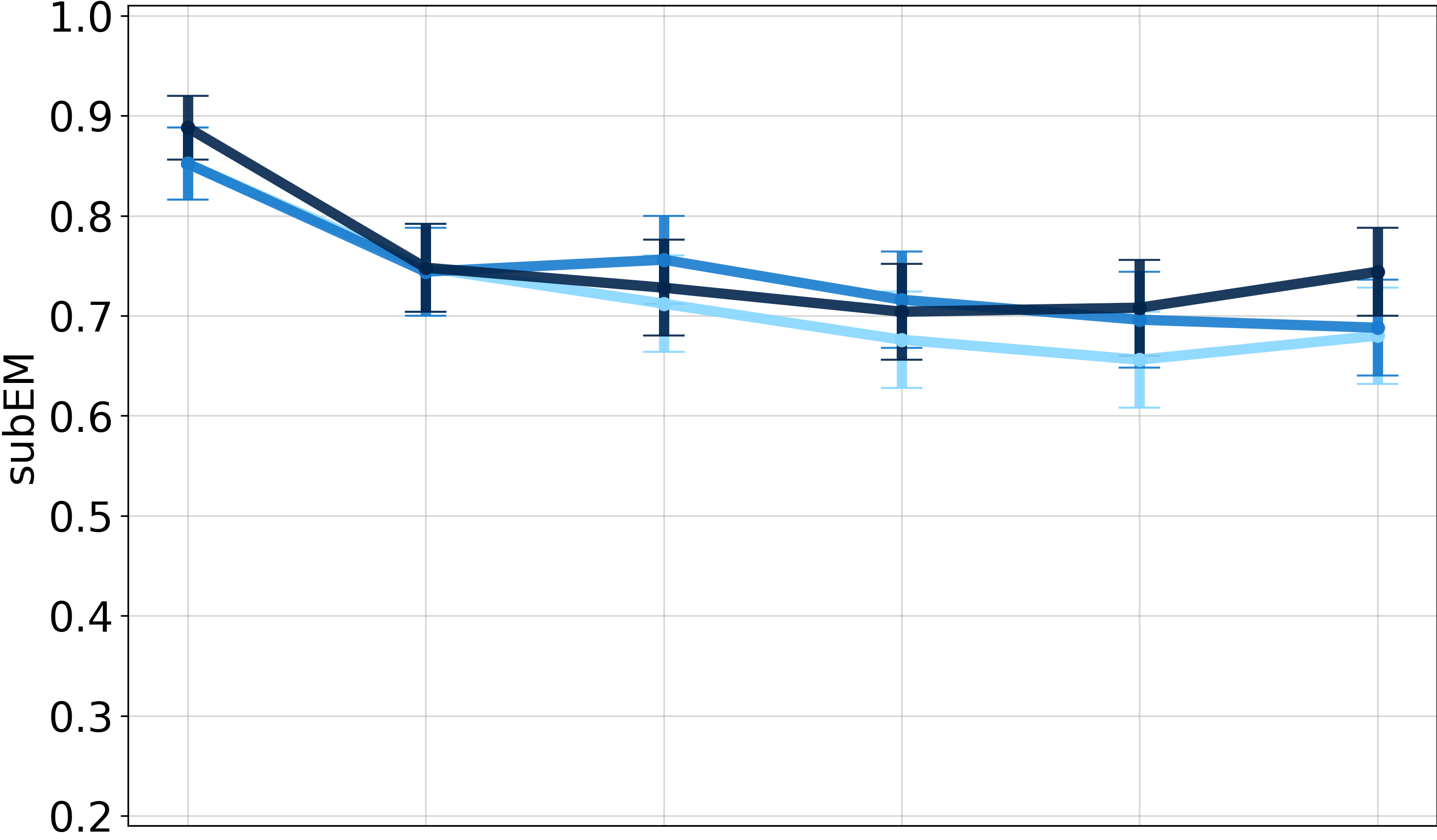}
    \end{subfigure}
    & 
    \begin{subfigure}{0.329\textwidth}
        \includegraphics[width=0.9\linewidth]{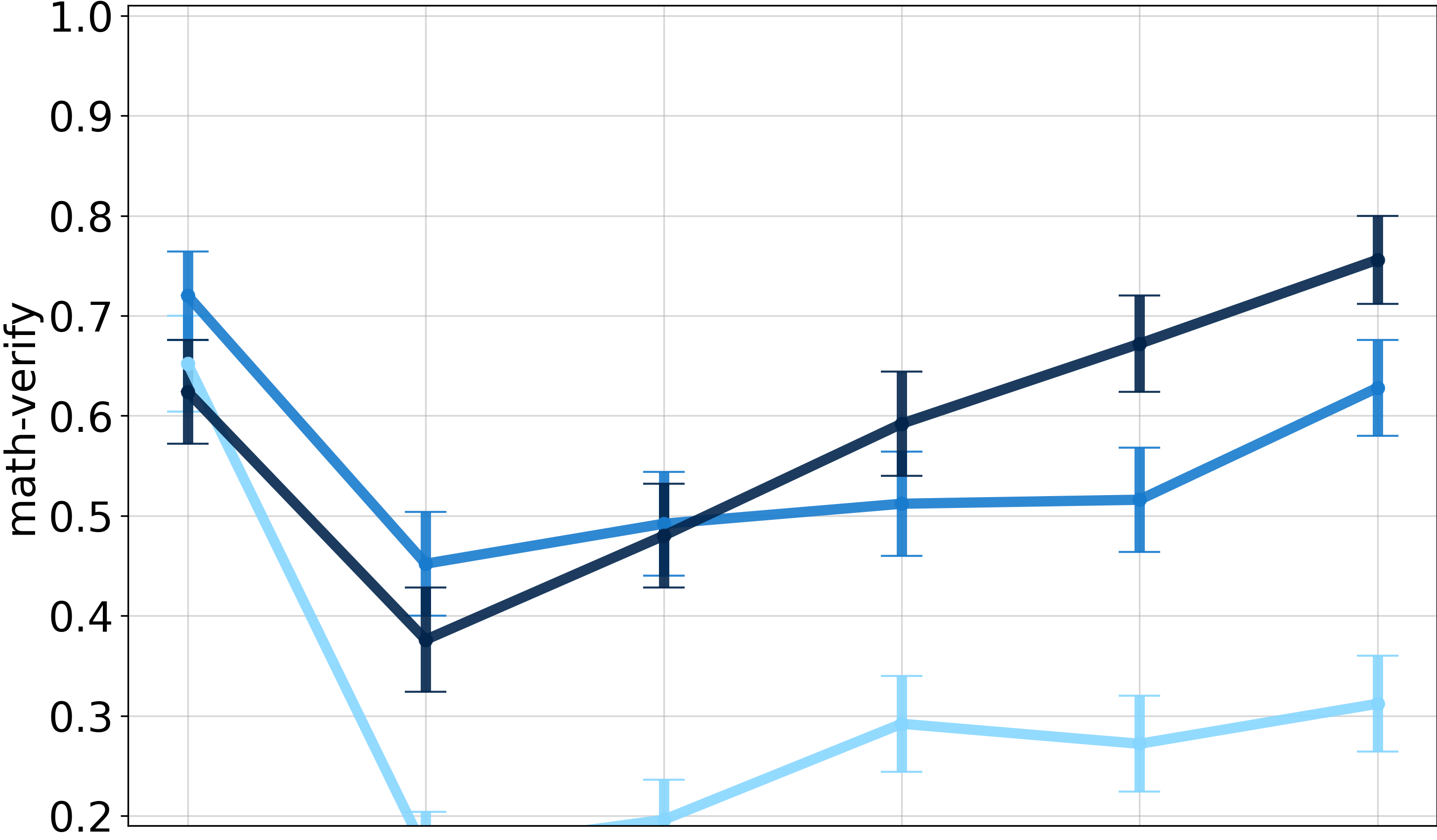}
    \end{subfigure}
    & 
    \hspace{-0.8cm} \raisebox{13ex}{\rotatebox[origin=c]{270}{\lightpurpletext{DeepSeek-R1}}}

    \end{tabular}    
    \caption{
    Model performance by gold context position (early to late in input), higher is better and error bars are 90\% CIs. Each row is a model, columns are benchmarks. 
    \textbf{Smaller gold contexts exhibit sharper performance degradation with later placement, especially in specialized domains (CBB, NM).}
    Larger contexts mitigate this sensitivity, highlighting the stabilizing effect of richer input.
    }
    \label{fig:byposition_all_models}
\end{figure}

\subsection{Gold Context Position}
\label{subsec:gold:doc:pos}

From the results in Figure~\ref{fig:byposition_all_models}, we observe that \textbf{smaller gold contexts are hard to find \textit{regardless of their position}}. Nevertheless, \textbf{certain positions amplify the bias against smaller gold contexts}. 
Performance systematically declines when small gold contexts appear later in the input, while large gold contexts are more robust to position (Full results in Appendix~\ref{appdx:by_position}).

For instance, in CBB, Gemini-2.0-Flash achieves 94\% accuracy when the small gold context is placed at the start of the context window, but only 33\% when placed near the end, a 61-point drop. In contrast, the large gold context declines more gradually, from 84\% to 65\%, demonstrating greater positional resilience. This pattern held across all evaluated models and benchmarks \changed{with some effects more amplified than others}.

Importantly, the positional effect is more pronounced in domain-specific tasks (CBB and NM) than in general knowledge (NQ), suggesting that task type and gold size compound aggregation difficulty.

We also observe that models \textbf{exhibit stronger primacy bias with smaller gold contexts}: performance is consistently higher when the gold context appears early in the input window. This effect is especially pronounced for small gold contexts. In some cases, small gold contexts placed at the beginning of the input even outperformed medium or large contexts placed later.
This occurs often in the left and right columns of Figure~\ref{fig:byposition_all_models}, where the \lightbluetext{small gold} line starts at the top at gold position 0.0 before crossing to the bottom.

This inversion highlights the sensitivity of model attention to positional cues when dealing with minimal evidence. While some bias exists for larger contexts, they are substantially more robust to position and do not exhibit the same sharp drop in middle and tail placements.

\subsection{Answer Token Repetition}
\label{subsec:answer_repetition}

\begin{wrapfigure}[19]{r}{0.49\textwidth}
    \vspace{-1cm}
    \centering
    \includegraphics[width=\linewidth,clip=false,trim=0.0cm 1.4cm 1.5cm 0.0cm]{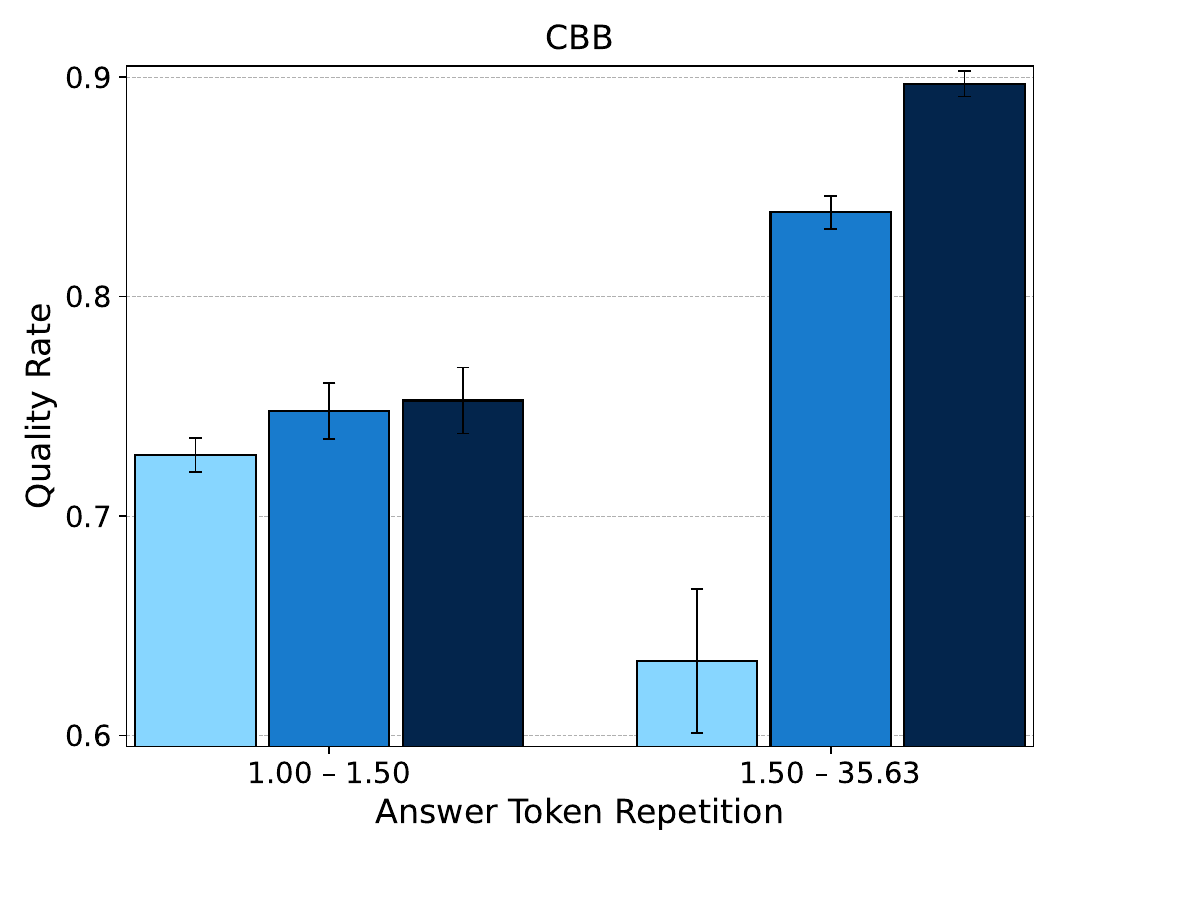}
    \caption{\changed{Performance across all models on CARDBiomedBench when bucketing tasks into two categories: answer token repetition less than or equal to, and greater than, 1.5. \lightbluetext{Smaller golds} yield lower accuracy compared to \darkbluetext{larger golds} across both bins. Error bars are 95\% confidence intervals, with some wider due to sample size.}
    }
    \label{fig:confounder_perf:answer:tok}
\end{wrapfigure}

\changed{If larger gold contexts contained the exact answer tokens more frequently, this could partly explain the phenomenon we observed---repeated encoding of the answer tokens would make it easier for the model to detect and attend to them. To test this, we compute \textit{answer token repetition}, defined for an answer $a$ and a context $c$ as: $\text{AnsTokRepetition}(a, c) = \frac{1}{|T(a)|^*} \sum_{t \in T(c)} \mathbf{1}[t \in T(a)],$ where $T(x)$ are the tokens of $x$, $|T(x)|^*$ is the number of \textit{unique} tokens in $x$, and $\mathbf{1}[\cdot]$ is the indicator function. Distributions of this metric across gold sizes and benchmarks are provided in Appendix~\ref{apdx:conf_dist}. We then bucket tasks using the median repetition value (1.5) and compare performance within each bin for CBB (Figure~\ref{fig:confounder_perf:answer:tok}).}

\changed{When answer repetition is low, larger gold contexts still outperform smaller ones, showing that repetition alone cannot explain the size effect. When repetition is high, performance improves for medium and large golds but not for small ones, indicating that repetition further amplifies the size effect. \textbf{While repetition is beneficial for LLM performance, it is insufficient to explain the observed size effect. Small gold contexts remain consistently harder to detect and use.}
}




\subsection{Gold-to-distractor Ratio}
\label{subsec:gold-to-distractor}

\begin{wrapfigure}[18]{r}{0.49\textwidth}
    \vspace{-0.9cm}
    \centering
    {\includegraphics[width=\linewidth,clip=false,trim=0.0cm 0.9cm 2.1cm 0.0cm]{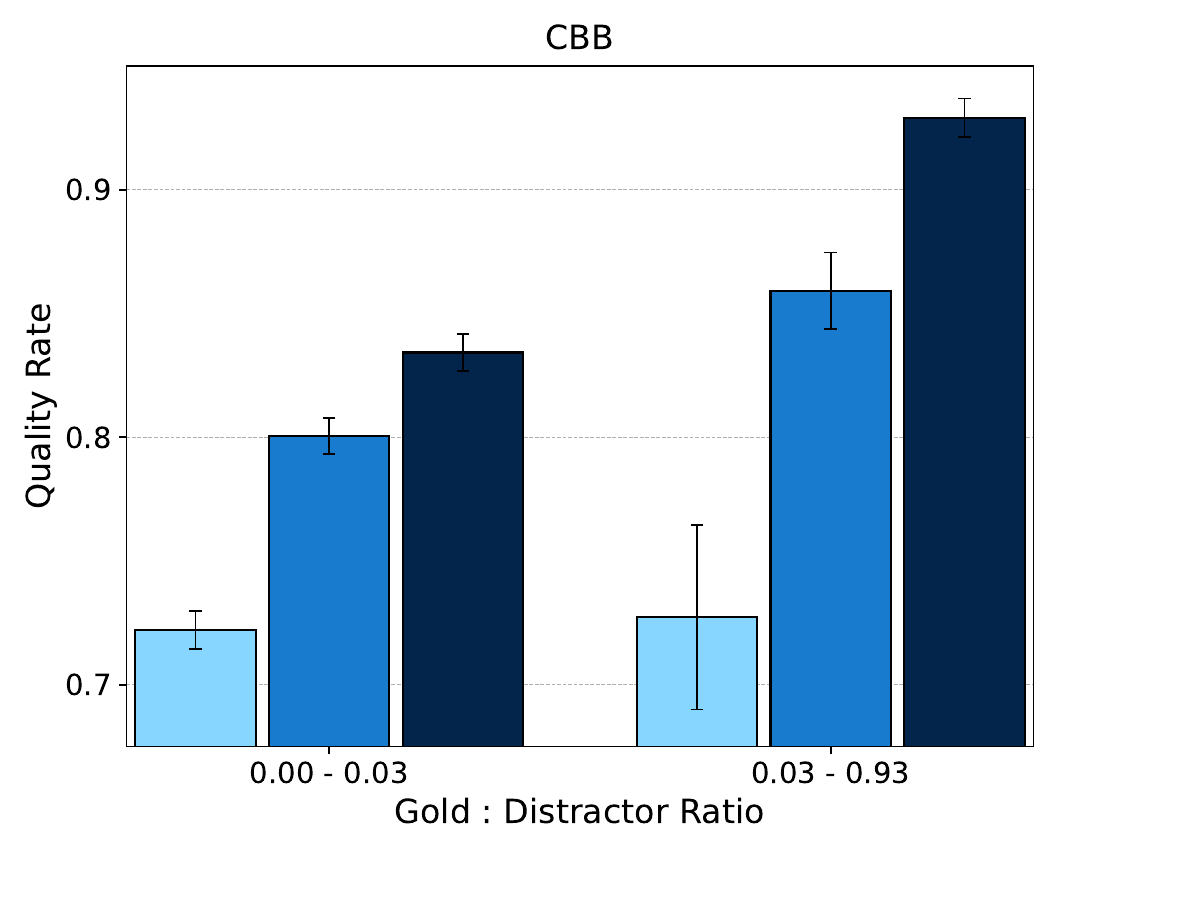}}
    \caption{Performance \changed{across all models on CARDBiomedBench when bucketing tasks by the mean (0.029)} for gold-to-distractor ratio. \lightbluetext{Smaller golds} yields lower accuracy compared to \darkbluetext{larger golds}. Error bars are 95\% confidence intervals, some are larger due to small sample size in that bin.}
    \label{fig:confounder_perf:gold_to_distractor}
\end{wrapfigure}

Given a fixed distractor budget, changing the size of the gold context also changes the proportion of \goldtext{gold tokens} to \greytext{distractor tokens} within a context window. \changed{For a gold context $g$ and a distractor set $D$, we can compute: $ \text{Gold-to-Distractor Ratio}(g, D) = \frac{|T(g)|}{\sum_{d \in D} |T(d)|},$ where $T(x)$ are the tokens of $x$ and $|T(x)|$ is the total number of tokens.} This raises the question of, if the positive effect of larger gold contexts is due to the size itself, or the increased ratio? \changed{Bucketing the tasks using the mean ratio (0.029)} into similar ranges of gold-to-distractor ratio, we can see if size still has an effect when the ratio is held constant. Figure~\ref{fig:confounder_perf:gold_to_distractor} shows just this, and larger golds consistently outperform smaller ones within bins. \textbf{Even after controlling for gold-to-distractor ratio, gold context size remains a strong indicator of performance.}




\subsection{Distractor Volume}
\label{subsec:distractor:volume}

\begin{figure}[ht]
    \centering
    \begin{subfigure}{0.329\textwidth}
        \includegraphics[width=\linewidth]{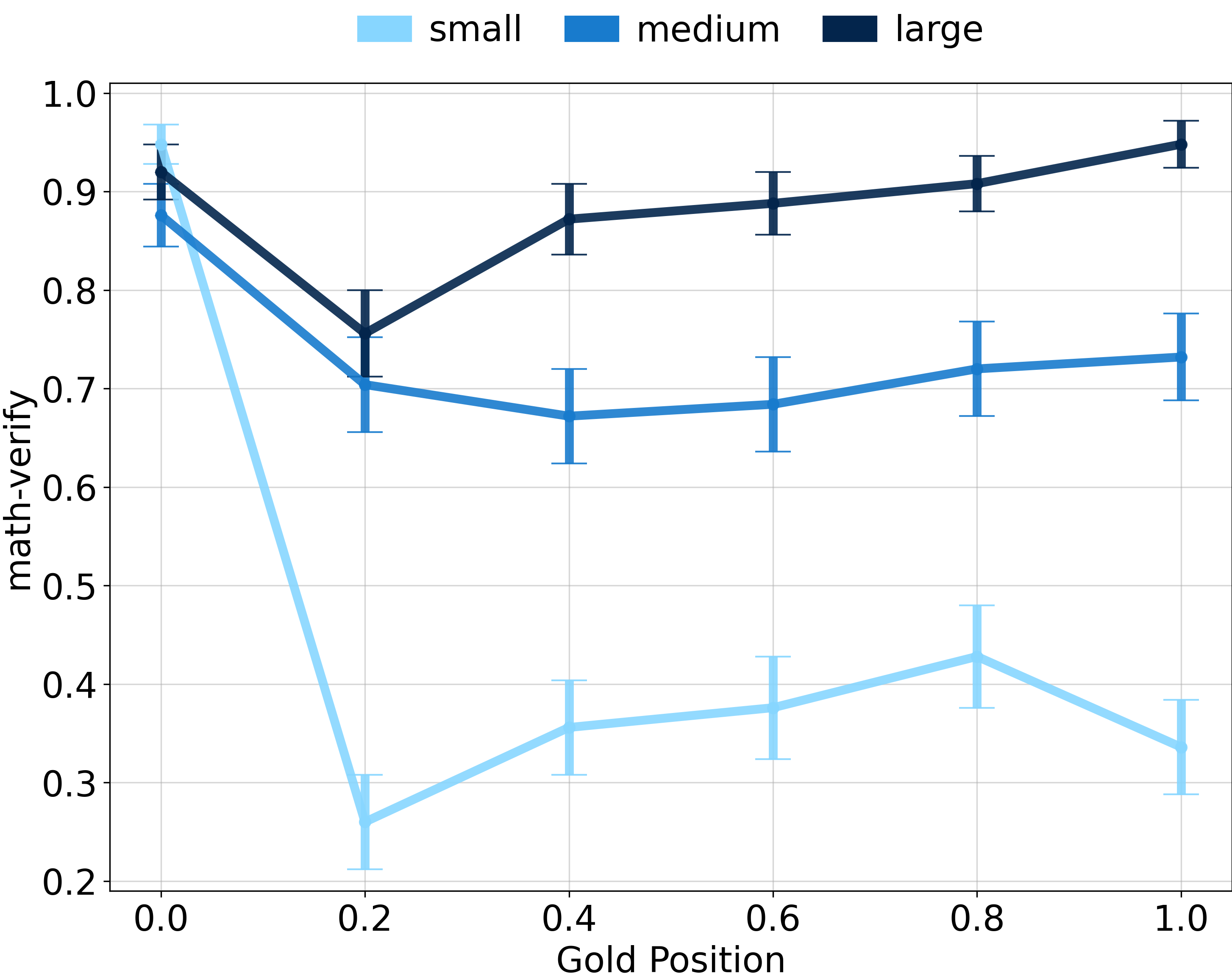}
        \caption{5 Distractors}
    \end{subfigure}
    \hfill
    \begin{subfigure}{0.329\textwidth}
        \includegraphics[width=\linewidth]{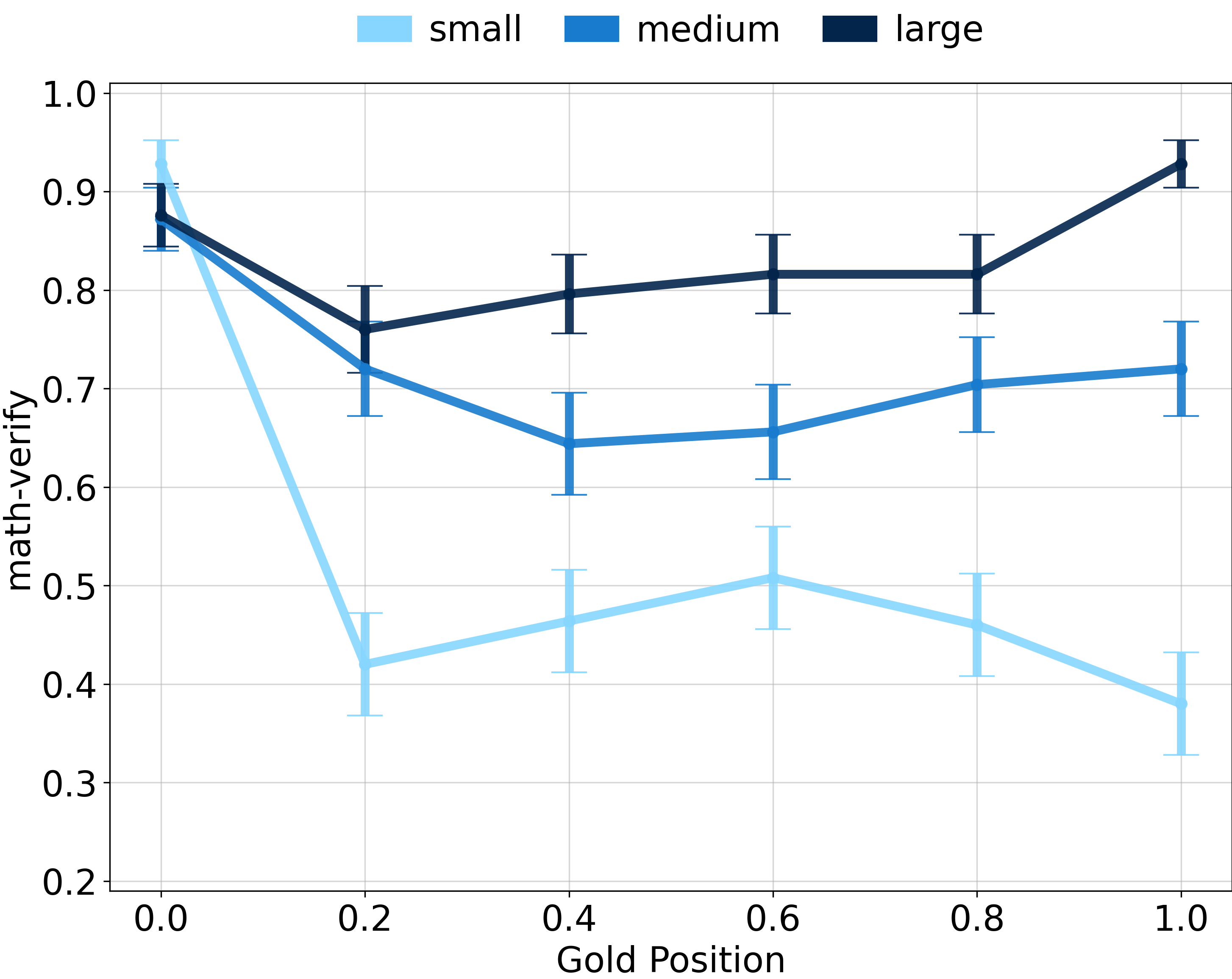}
        \caption{10 Distractors}
    \end{subfigure}
    \hfill
    \begin{subfigure}{0.329\textwidth}
        \includegraphics[width=\linewidth]{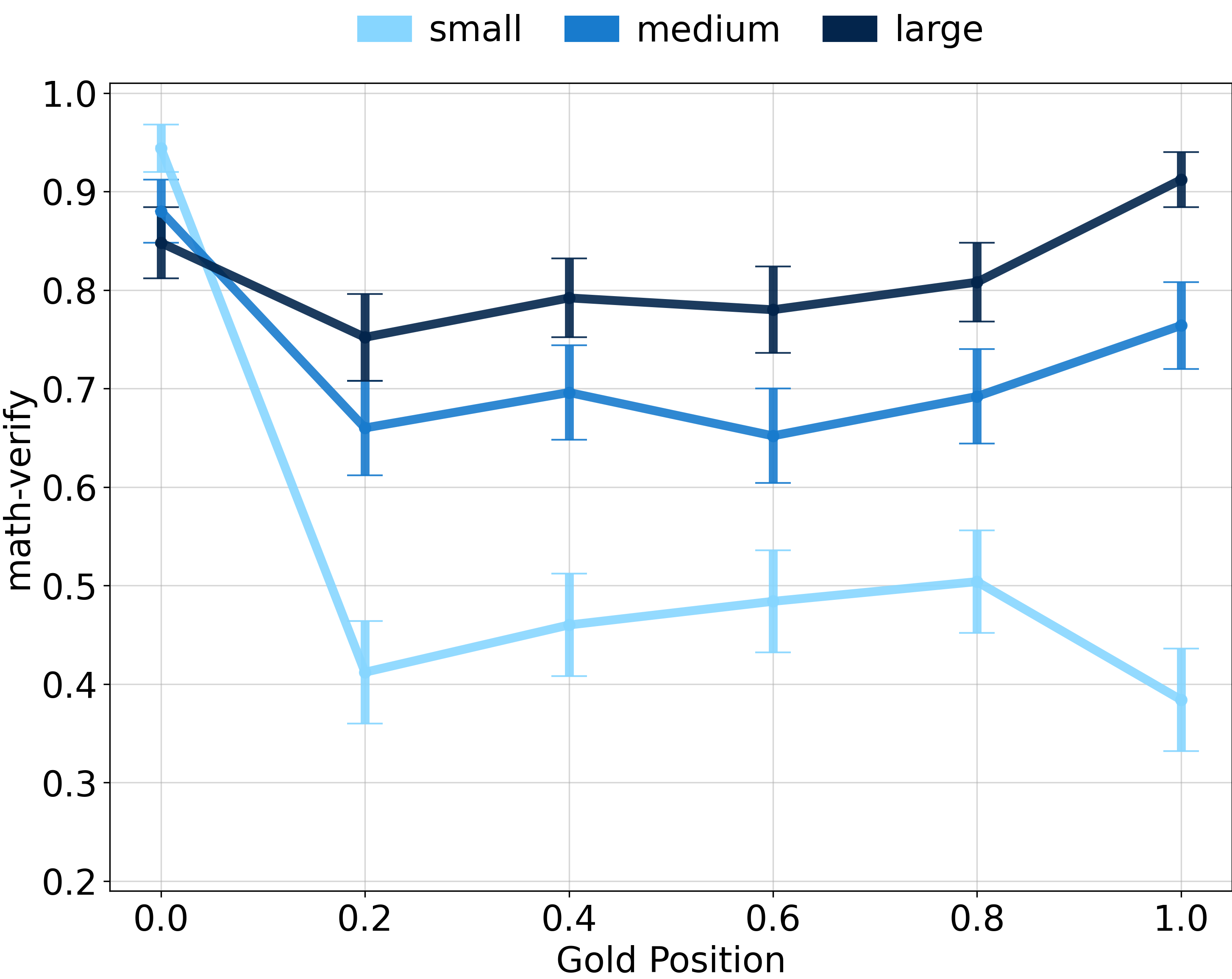}
        \caption{15 Distractors}
    \end{subfigure}
    \caption{
    Gemini-2.0-Flash performance on NuminaMath1.5 as the number of distractor documents increases (error bars are 90\% CIs). Despite growing distractor noise (up to $\sim$75k tokens), the performance gap between small and large gold contexts persists. 
    \textbf{This confirms that gold context size remains a key factor in long-context reasoning under high-noise conditions.}
    }
    \label{fig:increase_distractors}
\end{figure}

To evaluate the robustness of the gold context size effect under varying degrees of context noise, we systematically increased the number of distractor documents. We leveraged our adaptation of NuminaMath1.5 to run experiments with 5, 10, and 15 distractors, approximately 25k, 50k, and 75k distractor tokens, respectively. Figure~\ref{fig:increase_distractors} shows that performance is strongly influenced by gold context size, regardless of distractor volume. This reinforces that size remains a dominant variable, even when noise levels change.

\subsection{Domain Specificity of Tasks}
\label{subsec:domain:specificity}
\begin{wrapfigure}[19]{r}{0.51\textwidth}
    \vspace{-0.8cm}
    \centering
    \includegraphics[width=\linewidth,trim=0cm 0.1cm 0cm 0.2cm]{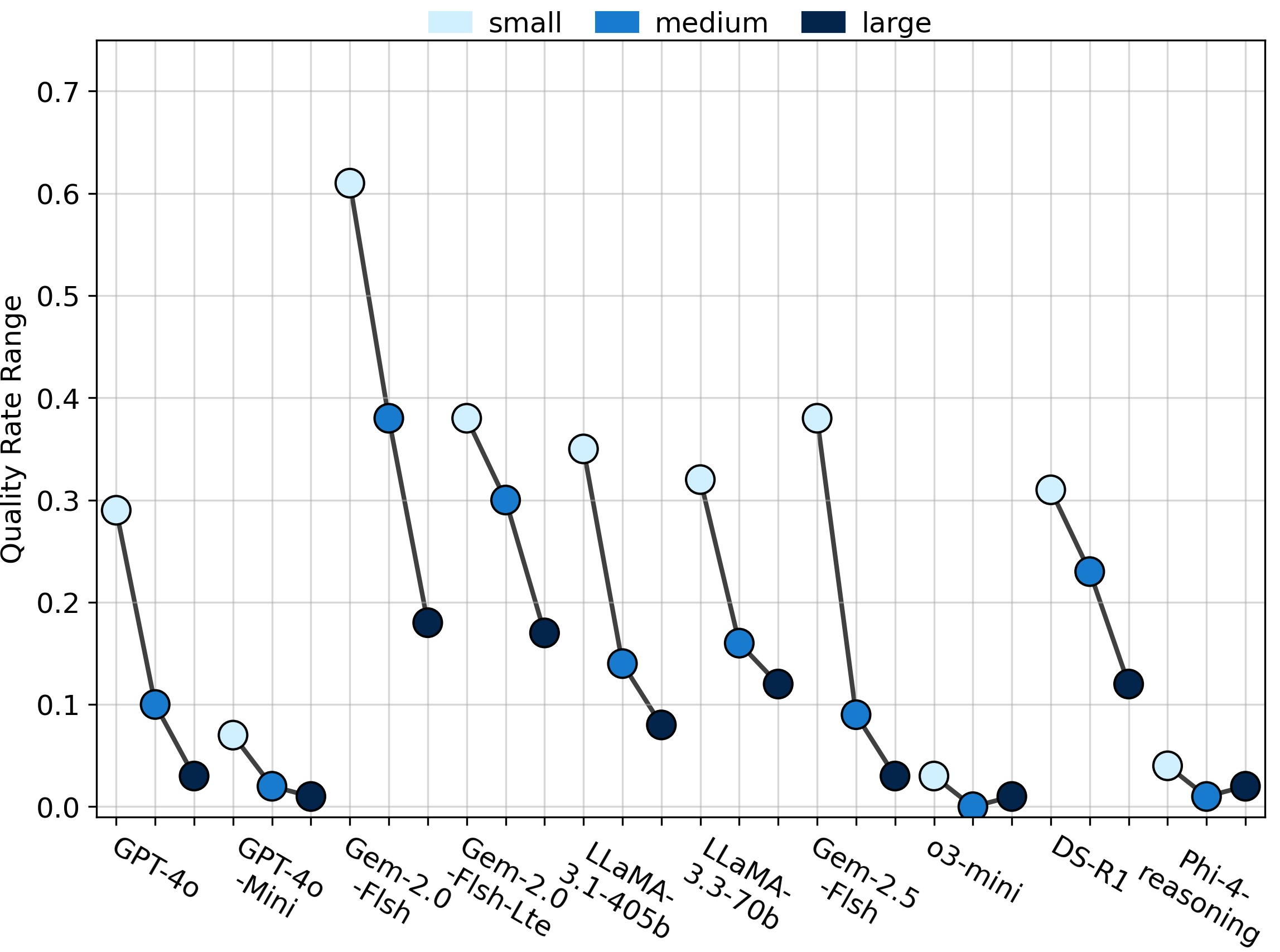}
    \caption{
        Positional sensitivity for CARDBiomedBench. For each model and gold context size, we compute the performance range across positions. \textbf{Smaller gold contexts exhibit much higher sensitivity (larger ranges), especially in domain-specific tasks. Larger gold contexts yield more stable performance across positions.}
    }
    \label{fig:range_cbb}
\end{wrapfigure}
The effects of gold context size are notably amplified in domain-specific tasks compared to general knowledge. Figure~\ref{fig:range_cbb} quantifies this \changed{for CBB} by measuring the range in model performance across different gold context positions.
\changed{(The results for other benchmarks are in Appendix~\ref{subsec:pos_sens_appdx}).}
For each model and gold size, we compute the performance range as the difference between maximum and minimum scores across all positions:
\begin{equation}
\begin{aligned}
\text{Range}
&= \mathop{\max}_{i \in \{1,\dots,n\}} \text{perf}(\text{position}_i) \\
&\;\;-\;\mathop{\min}_{i \in \{1,\dots,n\}} \text{perf}(\text{position}_i)
\end{aligned}
\label{eq:range}
\end{equation}
For example, on NuminaMath1.5, Gemini-2.0-Flash showed a performance range of 72\% for small gold contexts, compared to only 20\% for large gold. A similar pattern held in CARDBiomedBench. In contrast, NaturalQuestions exhibited smaller variation across all sizes, likely due to easier questions and higher closed-book baseline scores. This suggests that general knowledge tasks may be inherently more resilient to gold context variability.


\changed{
\subsection{Joint Modeling of Gold-Size \& Confounders with Multivariate Regression}
\label{subsec:logreg}
To assess the independent contribution of gold size relative to other factors, we fit a multivariate logistic regression predicting correctness from gold size, answer-token repetition, gold-to-distractor ratio, and position across all tasks and models. Earlier sections analyzed these variables \textit{in isolation}, but in realistic long-context settings they co-vary, making it unclear which effects genuinely stem from gold size. The joint regression provides a unified \textit{estimate of each factor’s partial effect}.
}

\changed{
\textbf{Model specification.} For each benchmark, we regress the binary correctness outcome on four categorical predictors:
\begin{equation}
\Pr\!\big(\mathrm{correct}\big)
=
\sigma\!\left(
\beta_0
+ \beta_{\text{size}}\, X_{\text{size}}
+ \beta_{\text{rep}}\, X_{\text{rep}}
+ \beta_{\text{ratio}}\, X_{\text{ratio}}
+ \beta_{\text{pos}}\, X_{\text{pos}}
\right)
\end{equation}
All variables are encoded categorically for direct interpretability. Repetition and ratio are binarized into \textit{high} vs.\ \textit{low} relative to within-benchmark means, and all predictors enter the model through their corresponding indicator variables $X_{\text{size}}, X_{\text{rep}}, X_{\text{ratio}}, X_{\text{pos}}$.
}

\changed{
Here are how these variables defined: 
(1) \textbf{Gold Size ($X_{\text{size}}$).}
        A categorical factor encoding gold size, with \textit{small} as the reference and indicators for \textit{medium} and \textit{large}. $\beta_{\text{size}}$ gives the change in log-odds when moving from \textit{small} to larger contexts.
(2)  \textbf{Answer Token Repetition ($X_{\text{rep}}$).}
    Token overlap with the answer, binarized at the benchmark mean into \textit{low} (reference) vs.\ \textit{high}. $\beta_{\text{rep}}$ gives the log-odds difference between the two levels.
(3) \textbf{Gold-to-Distractor Ratio ($X_{\text{ratio}}$).}
    The proportion of gold to distractor tokens in the context window, binarized at the benchmark mean into \textit{low} (reference) vs.\ \textit{high}. The coefficient $\beta_{\text{ratio}}$ quantifies the effect of moving from a lower share of gold tokens to a higher one.
(4) \textbf{Position ($X_{\text{pos}}$).}
    The gold context’s normalized start position, with $0.0$ as the reference. Indicators represent other positions, and $\beta_{\text{pos}}$ captures how placement in the window affects correctness.
}

\begin{wraptable}[17]{r}{0.49\textwidth}
\vspace{-0.5cm}
\centering
\small
\setlength{\tabcolsep}{10pt}
\begin{tabular}{lcc}
    \toprule
    \textbf{Predictor} & \textbf{Coef} & \textbf{95\% CI} \\
    \midrule
    Intercept & 1.6459 & [1.571, 1.720] \\
    \midrule
    \multicolumn{3}{l}{\textit{Gold Size}} \\
    \quad md vs.\ sm & 0.3933  &  [0.336, 0.451] \\
    \quad lg vs.\ sm  & 0.5354  &  [0.466, 0.605] \\
    \midrule
    \multicolumn{3}{l}{\textit{Answer Token Repetition}} \\
    \quad high vs.\ low & 0.5924  &  [0.511, 0.674] \\
    \midrule
    \multicolumn{3}{l}{\textit{Gold-to-Distractor Ratio}} \\
    \quad high vs.\ low & 0.3699  &  [0.279, 0.461] \\
    \midrule
    \multicolumn{3}{l}{\textit{Position}} \\
    \quad 0.25 vs.\ 0.0 & -0.8087  &  [-0.896, -0.721] \\
    \quad 0.50 vs.\ 0.0 & -0.9001  &  [-0.987, -0.813] \\
    \quad 0.75 vs.\ 0.0 & -0.8930  &  [-0.980, -0.806] \\
    \quad 1.00 vs.\ 0.0 & -0.8701  &  [-0.957, -0.783] \\
    \bottomrule
\end{tabular}
\caption{Logistic Regression Results (CBB)}
\label{tab:logreg:cbb:main}
\end{wraptable}
\changed{
\textbf{Results.}
We show CBB results in Table~\ref{tab:logreg:cbb:main} where we see that gold size has a strong independent effect: medium and large contexts significantly boost correctness after adjusting for all confounders. Repetition and ratio also help, while position has the largest negative impact---performance drops sharply when the gold appears anywhere other than at the start of the window. \textbf{CBB performance improves reliably with larger gold contexts, in ways not explained by repetition, ratio, or position alone.}
}

\changed{
Appendix~\ref{apdx:logreg} also reports all three benchmarks.
\textbf{Across all benchmarks, gold context size remains a significant, independent predictor of correctness even after controlling for repetition, ratio, and position.} Medium and large golds consistently increase the log-odds of producing the correct answer relative to small ones.
}




\section{Related Work}
We review related work in the context of long-context reasoning, focusing on three themes: positional biases in LLMs, long-context evaluation frameworks, and mitigation strategies.

\textbf{\changed{Long-context} biases in LLMs.}
Position bias, the tendency of LLMs to over- or under-attend to different parts of the input, has emerged as a fundamental challenge. Prior work has identified several variants: \textit{primacy bias}, where early content is favored~\citep{wang2023primacy};  \textit{recency bias}, where later content dominates~\citep{zheng2023large}; and \textit{U-shaped bias}, where mid-context is under-attended~\citep{liu2024lost}. These effects persist across model architectures, alignment strategies~\citep{liu2024lost}, extended context lengths~\citep{lee2024can,veselipositional}, and, to some extent, in internal representations~\citep{lu2024insights}. Our work contributes to this literature by introducing a new dimension: we show that \textit{the size of the gold context modulates the strength of positional bias}. 

A few recent works have examined different trade-offs in NIAH settings.
\citet{levy2024task} study variable input lengths and show that, \changed{for a \textit{fixed} gold context, \textit{adding more distractors} degrades performance. 
This is essentially the inverse of our setup, where we fix the distractor content and systematically vary the length of the gold context.}
\changed{
The parallel work of \citet{levy2025more} investigates the case where \textit{all documents have equal length} while \textit{the total context length is held fixed}. 
They find that \textit{longer documents} (with fewer total documents) make the gold easier to discover.
This is complementary to our work: instead of varying absolute length under equal-length constraints, we keep distractors fixed and vary \textit{the relative size of the same gold evidence}.
}
Finally, \citet{dai2024deniahlincontextfeaturesinfluence} study NIAH performance in synthetic key–value retrieval tasks, focusing on varying answer-span lengths and task lengths. 
In contrast, we \textit{hold the question and answer fixed} and vary the size of the surrounding gold context in open-ended QA, enabling a controlled analysis of how gold context size alone affects aggregation robustness.


\textbf{Frameworks for long-context evaluation.}
Long-context evaluation has progressed from synthetic toy tasks to increasingly realistic settings. Early work such as Long-Range Arena~\citep{tay2021long} introduced standardized tasks for comparing transformer variants. Subsequent benchmarks expanded this space~\citep{bai2024longbench,li2024loogle,gao2025uniah,li2025larabench,modarressi2025nolima,ling2025longreason,zhang2024toolbehonest,ye2024analobench,yen2024helmet}, exploring document synthesis~\citep{shaham2022scrolls,shaham2023zeroscrolls}, document-level retrieval~\citep{yen2025helmetevaluatelongcontextlanguage}, citation verification~\citep{zhang2024longcite}, and biomedical reasoning~\citep{adams2024longhealth,cui2025curie}. Most adopt a needle-in-a-haystack (NIAH) formulation~\citep{kamradt2023needle,hsieh2024ruler}, where a small relevant span must be recovered from distractors. Others move beyond strict NIAH by incorporating aggregation, multi-hop inference~\citep{zhuang2025docpuzz,katsis2025mtrag}, or mixed-modality inputs~\citep{wu2025longmemeval}.
\changed{Our work builds on this trajectory by adapting natural, domain-specific datasets to simulate realistic NIAH aggregation.}

\textbf{Mitigation strategies for \changed{long-context} biases.}
Prior work proposes various methods for reducing positional sensitivity, including context compression~\citep{jiang2024longllmlingua}, distilling long-context information into model weights~\citep{cao2025infiniteicl}, attention calibration~\citep{hsieh2024found}, modified positional encodings~\citep{zhang2024found,zheng2024dape}, and debiased fine-tuning~\citep{xiong2024artificial}. While these methods can alleviate positional bias, many introduce side effects~\citep{zhao2024understanding}, and robust long-context generalization remains challenging.
Our contribution is diagnostic rather than corrective: we uncover a novel bias connected to the gold context size. Whether existing mitigation strategies can address this effect remains an open question.


\section{Discussion, Limitations, and Conclusion}
\label{sec:disc_lim_conc}


\textbf{Why does gold context size strongly affect the accuracy?}
\changed{
At a high level, the finding here is pretty intuitive: Larger gold contexts spread semantically relevant information across more tokens, making the signal more resilient to positional noise and less likely to be overlooked amid distractors. 
However, we do not claim this is the only mechanism involved. 
Our findings suggest that interacting factors may be at play here, as discussed in  \S\ref{sec:confounding:factors}---including positional dynamics, answer-context richness, evidence competition, and domain effects. While the overall trend (large > medium > small) is consistent across all analyses, the precise interaction among these factors remains open and is an important direction for future work.}



\textbf{Practical implications of our findings.}
While prior work has studied factors like positional bias and distractor count, our results highlight an overlooked factor: the effect of heterogeneity in the size of evidence documents. 
\changed{
Aggregation quality can degrade sharply when small-but-critical evidence is mixed with much longer retrieved passages.
This naturally arises in retrieval-augmented systems or agentic pipelines where an aggregator must select a single piece of evidence from a concatenation of heterogeneous sources. 
Practitioners may mitigate this by ensuring that the documents are balanced in lengths before being aggregated or by avoiding pipelines that combine extremely short and very long contexts, thereby reducing size-induced attention asymmetries.
}

\textbf{Limitations of our study.}  
We fixed distractor lengths to better reflect real-world conditions, resulting in varying gold-to-distractor ratios. This may confound whether performance differences stem from gold context size alone or its relative share. Parallel work has  investigated this, offering complementary results to ours~\citep{levy2025more}.
Second, the tasks we use could be further grounded in real deployment scenarios.
\changed{Third, our study focuses on aggregating a single needle; extending this to multi-needle or multi-hop settings is an important next step.} 
Future work should address these. 

\textbf{Conclusion.}
Our study reveals a fundamental yet previously overlooked limitation in LLM aggregation capabilities: \textbf{the size of relevant information critically influences aggregation effectiveness in long-context tasks}. Through systematic evaluation, we demonstrated that smaller gold contexts degrade model performance substantially and exacerbate positional sensitivity, especially in domain-specific tasks. This discovery underscores a crucial vulnerability in real-world agentic deployments, where relevant evidence often appears unpredictably scattered amidst extensive distractors. As language models become central to applications requiring precise and trustworthy reasoning-from scientific discovery to personalized assistants-our findings highlight the urgent need to rethink aggregation strategies. Future LLM-driven systems must explicitly address context-size variability to ensure reliability, safety, and user trust in the face of complex, noisy real-world information streams.


\bibliography{ref}

@string{emnlp = "Conference on Empirical Methods in Natural Language Processing \CNFX{EMNLP}"}

@string{acl = "Annual Meeting of the Association for Computational Linguistics \CNFX{ACL}"}

@string{chi = "Conference on Human Factors in Computing Systems \CNFX{CHI}"}

@string{sigir = "Conference of the Association for Computing Machinery Special Interest Group in Information Retrieval \CNFX{SIGIR}"}

@string{science = "Science"}

@preamble{"\providecommand{\CNFX}[1]{{\em{\textrm{(#1)}}}}" }

@article{levy2025more,
  title={More Documents, Same Length: Isolating the Challenge of Multiple Documents in RAG},
  author={Levy, Shahar and Mazor, Nir and Shalmon, Lihi and Hassid, Michael and Stanovsky, Gabriel},
  journal={arXiv preprint arXiv:2503.04388},
  year={2025}
}

@inproceedings{tay2021long,
	title        = {Long Range Arena : A Benchmark for Efficient Transformers},
	author       = {Yi Tay and Mostafa Dehghani and Samira Abnar and Yikang Shen and Dara Bahri and Philip Pham and Jinfeng Rao and Liu Yang and Sebastian Ruder and Donald Metzler},
	year         = 2021,
	booktitle    = {International Conference on Learning Representations},
	url          = {https://openreview.net/forum?id=qVyeW-grC2k},
}

@misc{dubey2024llama3herdmodels,
	title        = {The Llama 3 Herd of Models},
	author       = {Abhimanyu Dubey and Abhinav Jauhri and Abhinav Pandey and Abhishek Kadian and Ahmad Al-Dahle and Aiesha Letman and Akhil Mathur and Alan Schelten and Amy Yang and Angela Fan and Anirudh Goyal and Anthony Hartshorn and Aobo Yang and Archi Mitra and Archie Sravankumar and Artem Korenev and Arthur Hinsvark and Arun Rao and Aston Zhang and Aurelien Rodriguez and Austen Gregerson and Ava Spataru and Baptiste Roziere and Bethany Biron and Binh Tang and Bobbie Chern and Charlotte Caucheteux and Chaya Nayak and Chloe Bi and Chris Marra and Chris McConnell and Christian Keller and Christophe Touret and Chunyang Wu and Corinne Wong and Cristian Canton Ferrer and Cyrus Nikolaidis and Damien Allonsius and Daniel Song and Danielle Pintz and Danny Livshits and David Esiobu and Dhruv Choudhary and Dhruv Mahajan and Diego Garcia-Olano and Diego Perino and Dieuwke Hupkes and Egor Lakomkin and Ehab AlBadawy and Elina Lobanova and Emily Dinan and Eric Michael Smith and Filip Radenovic and Frank Zhang and Gabriel Synnaeve and Gabrielle Lee and Georgia Lewis Anderson and Graeme Nail and Gregoire Mialon and Guan Pang and Guillem Cucurell and Hailey Nguyen and Hannah Korevaar and Hu Xu and Hugo Touvron and Iliyan Zarov and Imanol Arrieta Ibarra and Isabel Kloumann and Ishan Misra and Ivan Evtimov and Jade Copet and Jaewon Lee and Jan Geffert and Jana Vranes and Jason Park and Jay Mahadeokar and Jeet Shah and Jelmer van der Linde and Jennifer Billock and Jenny Hong and Jenya Lee and Jeremy Fu and Jianfeng Chi and Jianyu Huang and Jiawen Liu and Jie Wang and Jiecao Yu and Joanna Bitton and Joe Spisak and Jongsoo Park and Joseph Rocca and Joshua Johnstun and Joshua Saxe and Junteng Jia and Kalyan Vasuden Alwala and Kartikeya Upasani and Kate Plawiak and Ke Li and Kenneth Heafield and Kevin Stone and Khalid El-Arini and Krithika Iyer and Kshitiz Malik and Kuenley Chiu and Kunal Bhalla and Lauren Rantala-Yeary and Laurens van der Maaten and Lawrence Chen and Liang Tan and Liz Jenkins and Louis Martin and Lovish Madaan and Lubo Malo and Lukas Blecher and Lukas Landzaat and Luke de Oliveira and Madeline Muzzi and Mahesh Pasupuleti and Mannat Singh and Manohar Paluri and Marcin Kardas and Mathew Oldham and Mathieu Rita and Maya Pavlova and Melanie Kambadur and Mike Lewis and Min Si and Mitesh Kumar Singh and Mona Hassan and Naman Goyal and Narjes Torabi and Nikolay Bashlykov and Nikolay Bogoychev and Niladri Chatterji and Olivier Duchenne and Onur \c{C}elebi and Patrick Alrassy and Pengchuan Zhang and Pengwei Li and Petar Vasic and Peter Weng and Prajjwal Bhargava and Pratik Dubal and Praveen Krishnan and Punit Singh Koura and Puxin Xu and Qing He and Qingxiao Dong and Ragavan Srinivasan and Raj Ganapathy and Ramon Calderer and Ricardo Silveira Cabral and Robert Stojnic and Roberta Raileanu and Rohit Girdhar and Rohit Patel and Romain Sauvestre and Ronnie Polidoro and Roshan Sumbaly and Ross Taylor and Ruan Silva and Rui Hou and Rui Wang and Saghar Hosseini and Sahana Chennabasappa and Sanjay Singh and Sean Bell and Seohyun Sonia Kim and Sergey Edunov and Shaoliang Nie and Sharan Narang and Sharath Raparthy and Sheng Shen and Shengye Wan and Shruti Bhosale and Shun Zhang and Simon Vandenhende and Soumya Batra and Spencer Whitman and Sten Sootla and Stephane Collot and Suchin Gururangan and Sydney Borodinsky and Tamar Herman and Tara Fowler and Tarek Sheasha and Thomas Georgiou and Thomas Scialom and Tobias Speckbacher and Todor Mihaylov and Tong Xiao and Ujjwal Karn and Vedanuj Goswami and Vibhor Gupta and Vignesh Ramanathan and Viktor Kerkez and Vincent Gonguet and Virginie Do and Vish Vogeti and Vladan Petrovic and Weiwei Chu and Wenhan Xiong and Wenyin Fu and Whitney Meers and Xavier Martinet and Xiaodong Wang and Xiaoqing Ellen Tan and Xinfeng Xie and Xuchao Jia and Xuewei Wang and Yaelle Goldschlag and Yashesh Gaur and Yasmine Babaei and Yi Wen and Yiwen Song and Yuchen Zhang and Yue Li and Yuning Mao and Zacharie Delpierre Coudert and Zheng Yan and Zhengxing Chen and Zoe Papakipos and Aaditya Singh and Aaron Grattafiori and Abha Jain and Adam Kelsey and Adam Shajnfeld and Adithya Gangidi and Adolfo Victoria and Ahuva Goldstand and Ajay Menon and Ajay Sharma and Alex Boesenberg and Alex Vaughan and Alexei Baevski and Allie Feinstein and Amanda Kallet and Amit Sangani and Anam Yunus and Andrei Lupu and Andres Alvarado and Andrew Caples and Andrew Gu and Andrew Ho and Andrew Poulton and Andrew Ryan and Ankit Ramchandani and Annie Franco and Aparajita Saraf and Arkabandhu Chowdhury and Ashley Gabriel and Ashwin Bharambe and Assaf Eisenman and Azadeh Yazdan and Beau James and Ben Maurer and Benjamin Leonhardi and Bernie Huang and Beth Loyd and Beto De Paola and Bhargavi Paranjape and Bing Liu and Bo Wu and Boyu Ni and Braden Hancock and Bram Wasti and Brandon Spence and Brani Stojkovic and Brian Gamido and Britt Montalvo and Carl Parker and Carly Burton and Catalina Mejia and Changhan Wang and Changkyu Kim and Chao Zhou and Chester Hu and Ching-Hsiang Chu and Chris Cai and Chris Tindal and Christoph Feichtenhofer and Damon Civin and Dana Beaty and Daniel Kreymer and Daniel Li and Danny Wyatt and David Adkins and David Xu and Davide Testuggine and Delia David and Devi Parikh and Diana Liskovich and Didem Foss and Dingkang Wang and Duc Le and Dustin Holland and Edward Dowling and Eissa Jamil and Elaine Montgomery and Eleonora Presani and Emily Hahn and Emily Wood and Erik Brinkman and Esteban Arcaute and Evan Dunbar and Evan Smothers and Fei Sun and Felix Kreuk and Feng Tian and Firat Ozgenel and Francesco Caggioni and Francisco Guzm\'{a}n and Frank Kanayet and Frank Seide and Gabriela Medina Florez and Gabriella Schwarz and Gada Badeer and Georgia Swee and Gil Halpern and Govind Thattai and Grant Herman and Grigory Sizov and Guangyi and Zhang and Guna Lakshminarayanan and Hamid Shojanazeri and Han Zou and Hannah Wang and Hanwen Zha and Haroun Habeeb and Harrison Rudolph and Helen Suk and Henry Aspegren and Hunter Goldman and Ibrahim Damlaj and Igor Molybog and Igor Tufanov and Irina-Elena Veliche and Itai Gat and Jake Weissman and James Geboski and James Kohli and Japhet Asher and Jean-Baptiste Gaya and Jeff Marcus and Jeff Tang and Jennifer Chan and Jenny Zhen and Jeremy Reizenstein and Jeremy Teboul and Jessica Zhong and Jian Jin and Jingyi Yang and Joe Cummings and Jon Carvill and Jon Shepard and Jonathan McPhie and Jonathan Torres and Josh Ginsburg and Junjie Wang and Kai Wu and Kam Hou U and Karan Saxena and Karthik Prasad and Kartikay Khandelwal and Katayoun Zand and Kathy Matosich and Kaushik Veeraraghavan and Kelly Michelena and Keqian Li and Kun Huang and Kunal Chawla and Kushal Lakhotia and Kyle Huang and Lailin Chen and Lakshya Garg and Lavender A and Leandro Silva and Lee Bell and Lei Zhang and Liangpeng Guo and Licheng Yu and Liron Moshkovich and Luca Wehrstedt and Madian Khabsa and Manav Avalani and Manish Bhatt and Maria Tsimpoukelli and Martynas Mankus and Matan Hasson and Matthew Lennie and Matthias Reso and Maxim Groshev and Maxim Naumov and Maya Lathi and Meghan Keneally and Michael L. Seltzer and Michal Valko and Michelle Restrepo and Mihir Patel and Mik Vyatskov and Mikayel Samvelyan and Mike Clark and Mike Macey and Mike Wang and Miquel Jubert Hermoso and Mo Metanat and Mohammad Rastegari and Munish Bansal and Nandhini Santhanam and Natascha Parks and Natasha White and Navyata Bawa and Nayan Singhal and Nick Egebo and Nicolas Usunier and Nikolay Pavlovich Laptev and Ning Dong and Ning Zhang and Norman Cheng and Oleg Chernoguz and Olivia Hart and Omkar Salpekar and Ozlem Kalinli and Parkin Kent and Parth Parekh and Paul Saab and Pavan Balaji and Pedro Rittner and Philip Bontrager and Pierre Roux and Piotr Dollar and Polina Zvyagina and Prashant Ratanchandani and Pritish Yuvraj and Qian Liang and Rachad Alao and Rachel Rodriguez and Rafi Ayub and Raghotham Murthy and Raghu Nayani and Rahul Mitra and Raymond Li and Rebekkah Hogan and Robin Battey and Rocky Wang and Rohan Maheswari and Russ Howes and Ruty Rinott and Sai Jayesh Bondu and Samyak Datta and Sara Chugh and Sara Hunt and Sargun Dhillon and Sasha Sidorov and Satadru Pan and Saurabh Verma and Seiji Yamamoto and Sharadh Ramaswamy and Shaun Lindsay and Shaun Lindsay and Sheng Feng and Shenghao Lin and Shengxin Cindy Zha and Shiva Shankar and Shuqiang Zhang and Shuqiang Zhang and Sinong Wang and Sneha Agarwal and Soji Sajuyigbe and Soumith Chintala and Stephanie Max and Stephen Chen and Steve Kehoe and Steve Satterfield and Sudarshan Govindaprasad and Sumit Gupta and Sungmin Cho and Sunny Virk and Suraj Subramanian and Sy Choudhury and Sydney Goldman and Tal Remez and Tamar Glaser and Tamara Best and Thilo Kohler and Thomas Robinson and Tianhe Li and Tianjun Zhang and Tim Matthews and Timothy Chou and Tzook Shaked and Varun Vontimitta and Victoria Ajayi and Victoria Montanez and Vijai Mohan and Vinay Satish Kumar and Vishal Mangla and V\'{\i}tor Albiero and Vlad Ionescu and Vlad Poenaru and Vlad Tiberiu Mihailescu and Vladimir Ivanov and Wei Li and Wenchen Wang and Wenwen Jiang and Wes Bouaziz and Will Constable and Xiaocheng Tang and Xiaofang Wang and Xiaojian Wu and Xiaolan Wang and Xide Xia and Xilun Wu and Xinbo Gao and Yanjun Chen and Ye Hu and Ye Jia and Ye Qi and Yenda Li and Yilin Zhang and Ying Zhang and Yossi Adi and Youngjin Nam and Yu and Wang and Yuchen Hao and Yundi Qian and Yuzi He and Zach Rait and Zachary DeVito and Zef Rosnbrick and Zhaoduo Wen and Zhenyu Yang and Zhiwei Zhao},
	year         = 2024,
	url          = {https://arxiv.org/abs/2407.21783},
	eprint       = {2407.21783},
	archiveprefix = {arXiv},
	primaryclass = {cs.AI},
}

@inproceedings{ye2024analobench,
	title        = {{AnaloBench:} Benchmarking the Identification of Abstract and Long-context Analogies},
	author       = {Xiao Ye and Andrew Wang and Jacob Choi and Yining Lu and Shreya Sharma and Lingfeng Shen and Vijay Tiyyala and Nicholas Andrews and Daniel Khashabi},
	year         = 2024,
	booktitle    = emnlp,
	url          = {https://arxiv.org/abs/2402.12370},
	code         = {https://github.com/JHU-CLSP/AnaloBench},
	data         = {https://huggingface.co/datasets/jhu-clsp/AnaloBench},
	tweet        = {https://x.com/andrewwnlp/status/1767234121355735294},
	poster       = {https://danielkhashabi.com/files/2024\_analobench/EMNLP\_2024\_Analogies\_Poster.pdf},
	slides       = {https://danielkhashabi.com/files/2024\_analobench/Analogies-presentation-novideo.pdf},
	talk         = {https://youtu.be/X\_U\_jdXXfv0},
}

@article{weiqiwang2025arxiv2table,
	title        = {Can LLMs Generate Tabular Summaries of Science Papers? Rethinking the Evaluation Protocol},
	author       = {Weiqi Wang and Jiefu Ou and Yangqiu Song and Benjamin Van Durme and Daniel Khashabi},
	year         = 2025,
	journal      = {arXiv preprint arXiv:2504.10284},
	url          = {https://arxiv.org/abs/2504.10284},
	eprint       = {2504.10284},
	archiveprefix = {arXiv},
	primaryclass = {cs.CL},
	data         = {https://github.com/JHU-CLSP/arXiv2Table},
}

@article{gao2025sciencehierarchy,
	title        = {Science Hierarchography: Hierarchical Abstractions of Scientific Literature},
	author       = {Gao, Muhan and Shah, Jash and Wang, Weiqi and Khashabi, Daniel},
	year         = 2025,
	journal      = {arXiv preprint arXiv:2504.13834},
	url          = {https://arxiv.org/abs/2504.13834},
	eprint       = {2504.13834},
	archiveprefix = {arXiv},
	primaryclass = {cs.CL},
	data         = {https://github.com/JHU-CLSP/science-hierarchography/},
}

@inproceedings{zheng2023large,
  title={Large language models are not robust multiple choice selectors},
  author={Zheng, Chujie and Zhou, Hao and Meng, Fandong and Zhou, Jie and Huang, Minlie},
  booktitle={The Twelfth International Conference on Learning Representations},
  year={2023},
}

@inproceedings{wang2023primacy,
  title={Primacy Effect of ChatGPT},
  author={Wang, Yiwei and Cai, Yujun and Chen, Muhao and Liang, Yuxuan and Hooi, Bryan},
  booktitle={Proceedings of the 2023 Conference on Empirical Methods in Natural Language Processing},
  pages={108--115},
  year={2023}
}

@article{lee2024can,
  title={Can Long-Context Language Models Subsume Retrieval, RAG, SQL, and More?},
  author={Lee, Jinhyuk and Chen, Anthony and Dai, Zhuyun and Dua, Dheeru and Sachan, Devendra Singh and Boratko, Michael and Luan, Yi and Arnold, S{\'e}bastien MR and Perot, Vincent and Dalmia, Siddharth and others},
  journal={arXiv preprint arXiv:2406.13121},
  year={2024}
}

@inproceedings{jiang2024longllmlingua,
  title={Longllmlingua: Accelerating and enhancing llms in long context scenarios via prompt compression},
  author={Jiang, Huiqiang and Wu, Qianhui and Luo, Xufang and Li, Dongsheng and Lin, Chin-Yew and Yang, Yuqing and Qiu, Lili},
  booktitle={Findings of the Association for Computational Linguistics ACL 2024},
  pages={1658–1677},
  year={2024}
}

@inproceedings{hsieh2024found,
  title={Found in the middle: Calibrating Positional Attention Bias Improves Long Context Utilization},
  author={Hsieh, Cheng-Yu and Chuang, Yung-Sung and Li, Chun-Liang and Wang, Zifeng and Le, Long and Kumar, Abhishek and Glass, James and Ratner, Alexander and Lee, Chen-Yu and Krishna, Ranjay and others},
  booktitle={Findings of the Association for Computational Linguistics ACL 2024},
  pages={14982--14995},
  year={2024}
}

@article{lu2024insights,
  title={Insights into LLM Long-Context Failures: When Transformers Know but Don't Tell},
  author={Lu, Taiming and Gao, Muhan and Yu, Kuai and Byerly, Adam and Khashabi, Daniel},
  journal={arXiv preprint arXiv:2406.14673},
  year={2024}
}

@inproceedings{
cui2025curie,
title={{CURIE}: Evaluating {LLM}s on Multitask Scientific Long-Context Understanding and Reasoning},
author={Hao Cui and Zahra Shamsi and Gowoon Cheon and Xuejian Ma and Shutong Li and Maria Tikhanovskaya and Peter Christian Norgaard and Nayantara Mudur and Martyna Beata Plomecka and Paul Raccuglia and Yasaman Bahri and Victor V. Albert and Pranesh Srinivasan and Haining Pan and Philippe Faist and Brian A Rohr and Michael J. Statt and Dan Morris and Drew Purves and Elise Kleeman and Ruth Alcantara and Matthew Abraham and Muqthar Mohammad and Ean Phing VanLee and Chenfei Jiang and Elizabeth Dorfman and Eun-Ah Kim and Michael Brenner and Sameera S Ponda and Subhashini Venugopalan},
booktitle={The Thirteenth International Conference on Learning Representations},
year={2025},
url={https://openreview.net/forum?id=jw2fC6REUB}
}

@inproceedings{
bazgir2025agentichypothesis,
title={AgenticHypothesis: A Survey on Hypothesis Generation Using {LLM} Systems},
author={Adib Bazgir and Rama chandra Praneeth Madugula and Yuwen Zhang},
booktitle={Towards Agentic AI for Science: Hypothesis Generation, Comprehension, Quantification, and Validation},
year={2025},
url={https://openreview.net/forum?id=UeeyfR4CUg}
}

@article{sprueill2024chemreasoner,
  title={ChemReasoner: Heuristic search over a large language model's knowledge space using quantum-chemical feedback},
  author={Sprueill, Henry W and Edwards, Carl and Agarwal, Khushbu and Olarte, Mariefel V and Sanyal, Udishnu and Johnston, Conrad and Liu, Hongbin and Ji, Heng and Choudhury, Sutanay},
  journal={arXiv preprint arXiv:2402.10980},
  year={2024}
}

@article{xiong2024artificial,
  title={From Artificial Needles to Real Haystacks: Improving Retrieval Capabilities in LLMs by Finetuning on Synthetic Data},
  author={Xiong, Zheyang and Papageorgiou, Vasilis and Lee, Kangwook and Papailiopoulos, Dimitris},
  journal={arXiv preprint arXiv:2406.19292},
  year={2024}
}

@inproceedings{shaham2023zeroscrolls,
  title={ZeroSCROLLS: A Zero-Shot Benchmark for Long Text Understanding},
  author={Shaham, Uri and Ivgi, Maor and Efrat, Avia and Berant, Jonathan and Levy, Omer},
  booktitle={Findings of the Association for Computational Linguistics: EMNLP 2023},
  pages={7977--7989},
  year={2023}
}

@article{kamradt2023needle,
  author={Kamradt, Greg},
  title={Needle In A Haystack - Pressure Testing LLMs},
  journal={GitHub},
  year={2023},
  url={https://github.com/gkamradt/LLMTest_NeedleInAHaystack},
}

@article{hsieh2024ruler,
  title={RULER: What's the Real Context Size of Your Long-Context Language Models?},
  author={Hsieh, Cheng-Ping and Sun, Simeng and Kriman, Samuel and Acharya, Shantanu and Rekesh, Dima and Jia, Fei and Ginsburg, Boris},
  journal={arXiv preprint arXiv:2404.06654},
  year={2024}
}

@article{yen2024helmet,
  title={HELMET: How to Evaluate Long-Context Language Models Effectively and Thoroughly},
  author={Yen, Howard and Gao, Tianyu and Hou, Minmin and Ding, Ke and Fleischer, Daniel and Izasak, Peter and Wasserblat, Moshe and Chen, Danqi},
  journal={arXiv preprint arXiv:2410.02694},
  year={2024}
}

@inproceedings{shaham2022scrolls,
    title = "{SCROLLS}: Standardized {C}ompa{R}ison Over Long Language Sequences",
    author = "Shaham, Uri  and
      Segal, Elad  and
      Ivgi, Maor  and
      Efrat, Avia  and
      Yoran, Ori  and
      Haviv, Adi  and
      Gupta, Ankit  and
      Xiong, Wenhan  and
      Geva, Mor  and
      Berant, Jonathan  and
      Levy, Omer",
    booktitle = "Proceedings of the 2022 Conference on Empirical Methods in Natural Language Processing",
    url = "https://aclanthology.org/2022.emnlp-main.823/"
}

@article{liu2024lost,
    title = "Lost in the Middle: How Language Models Use Long Contexts",
    author = "Liu, Nelson F.  and
      Lin, Kevin  and
      Hewitt, John  and
      Paranjape, Ashwin  and
      Bevilacqua, Michele  and
      Petroni, Fabio  and
      Liang, Percy",
    journal = "Transactions of the Association for Computational Linguistics",
    url = "https://aclanthology.org/2024.tacl-1.9/",
}

@inproceedings{bai2024longbench,
    title = "{L}ong{B}ench: A Bilingual, Multitask Benchmark for Long Context Understanding",
    author = "Bai, Yushi  and
      Lv, Xin  and
      Zhang, Jiajie  and
      Lyu, Hongchang  and
      Tang, Jiankai  and
      Huang, Zhidian  and
      Du, Zhengxiao  and
      Liu, Xiao  and
      Zeng, Aohan  and
      Hou, Lei  and
      Dong, Yuxiao  and
      Tang, Jie  and
      Li, Juanzi",
    booktitle = "Proceedings of the 62nd Annual Meeting of the Association for Computational Linguistics (Volume 1: Long Papers)",
    url = "https://aclanthology.org/2024.acl-long.172/",
}

@inproceedings{li2024loogle,
    title = "{L}oo{GLE}: Can Long-Context Language Models Understand Long Contexts?",
    author = "Li, Jiaqi  and
      Wang, Mengmeng  and
      Zheng, Zilong  and
      Zhang, Muhan",
    booktitle = "Proceedings of the 62nd Annual Meeting of the Association for Computational Linguistics (Volume 1: Long Papers)",
    url = "https://aclanthology.org/2024.acl-long.859/",
}

@misc{adams2024longhealth,
      title={LongHealth: A Question Answering Benchmark with Long Clinical Documents}, 
      author={Lisa Adams and Felix Busch and Tianyu Han and Jean-Baptiste Excoffier and Matthieu Ortala and Alexander Löser and Hugo JWL. Aerts and Jakob Nikolas Kather and Daniel Truhn and Keno Bressem},
      year={2024},
      url={https://arxiv.org/abs/2401.14490}, 
}

@inproceedings{veselipositional,
  title={Positional Biases Shift as Inputs Approach Context Window Limits},
  author={Veseli, Blerta and Chibane, Julian and Toneva, Mariya and Koller, Alexander},
  booktitle={Second Conference on Language Modeling}, 
  year={2025}
}

@inproceedings{levy2024task,
    title = "Same Task, More Tokens: the Impact of Input Length on the Reasoning Performance of Large Language Models",
    author = "Levy, Mosh  and
      Jacoby, Alon  and
      Goldberg, Yoav",
    booktitle = "Proceedings of the 62nd Annual Meeting of the Association for Computational Linguistics (Volume 1: Long Papers)",
    url = "https://aclanthology.org/2024.acl-long.818/",
    year= {2024},
}

@article{zheng2024dape,
  title={Dape: Data-adaptive positional encoding for length extrapolation},
  author={Zheng, Chuanyang and Gao, Yihang and Shi, Han and Huang, Minbin and Li, Jingyao and Xiong, Jing and Ren, Xiaozhe and Ng, Michael and Jiang, Xin and Li, Zhenguo and others},
  journal={Advances in Neural Information Processing Systems},
  volume={37},
  pages={26659--26700},
  year={2024}
}

@article{zhang2024found,
  title={Found in the middle: How language models use long contexts better via plug-and-play positional encoding},
  author={Zhang, Zhenyu and Chen, Runjin and Liu, Shiwei and Yao, Zhewei and Ruwase, Olatunji and Chen, Beidi and Wu, Xiaoxia and Wang, Zhangyang},
  journal={arXiv preprint arXiv:2403.04797},
  year={2024}
}

@misc{gao2025uniah,
      title={U-NIAH: Unified RAG and LLM Evaluation for Long Context Needle-In-A-Haystack}, 
      author={Yunfan Gao and Yun Xiong and Wenlong Wu and Zijing Huang and Bohan Li and Haofen Wang},
      url={https://arxiv.org/abs/2503.00353}, 
}

@misc{zhuang2025docpuzz,
      title={DocPuzzle: A Process-Aware Benchmark for Evaluating Realistic Long-Context Reasoning Capabilities}, 
      author={Tianyi Zhuang and Chuqiao Kuang and Xiaoguang Li and Yihua Teng and Jihao Wu and Yasheng Wang and Lifeng Shang},
      url={https://arxiv.org/abs/2502.17807}, 
}

@article{zhang2024longcite,
  title={Longcite: Enabling llms to generate fine-grained citations in long-context qa},
  author={Zhang, Jiajie and Bai, Yushi and Lv, Xin and Gu, Wanjun and Liu, Danqing and Zou, Minhao and Cao, Shulin and Hou, Lei and Dong, Yuxiao and Feng, Ling and others},
  journal={arXiv preprint arXiv:2409.02897},
  year={2024}
}

@article{zhao2024understanding,
  title={Understanding Synthetic Context Extension via Retrieval Heads},
  author={Zhao, Xinyu and Yin, Fangcong and Durrett, Greg},
  journal={arXiv preprint arXiv:2410.22316},
  year={2024}
}

@article{cao2025infiniteicl,
  title={InfiniteICL: Breaking the Limit of Context Window Size via Long Short-term Memory Transformation},
  author={Cao, Bowen and Cai, Deng and Lam, Wai},
  journal={arXiv preprint arXiv:2504.01707},
  year={2025}
}

@misc{li2025larabench,
      title={LaRA: Benchmarking Retrieval-Augmented Generation and Long-Context LLMs -- No Silver Bullet for LC or RAG Routing}, 
      author={Kuan Li and Liwen Zhang and Yong Jiang and Pengjun Xie and Fei Huang and Shuai Wang and Minhao Cheng},
      url={https://arxiv.org/abs/2502.09977}, 
}

@misc{modarressi2025nolima,
      title={NoLiMa: Long-Context Evaluation Beyond Literal Matching}, 
      author={Ali Modarressi and Hanieh Deilamsalehy and Franck Dernoncourt and Trung Bui and Ryan A. Rossi and Seunghyun Yoon and Hinrich Schütze},
      year={2025},
      eprint={2502.05167},
      archivePrefix={arXiv},
      primaryClass={cs.CL},
      url={https://arxiv.org/abs/2502.05167}, 
}

@misc{ling2025longreason,
      title={LongReason: A Synthetic Long-Context Reasoning Benchmark via Context Expansion}, 
      author={Zhan Ling and Kang Liu and Kai Yan and Yifan Yang and Weijian Lin and Ting-Han Fan and Lingfeng Shen and Zhengyin Du and Jiecao Chen},
      year={2025},
      eprint={2501.15089},
      archivePrefix={arXiv},
      primaryClass={cs.CL},
      url={https://arxiv.org/abs/2501.15089}, 
}

@misc{katsis2025mtrag,
      title={MTRAG: A Multi-Turn Conversational Benchmark for Evaluating Retrieval-Augmented Generation Systems}, 
      author={Yannis Katsis and Sara Rosenthal and Kshitij Fadnis and Chulaka Gunasekara and Young-Suk Lee and Lucian Popa and Vraj Shah and Huaiyu Zhu and Danish Contractor and Marina Danilevsky},
      url={https://arxiv.org/abs/2501.03468}, 
}

@inproceedings{
wu2025longmemeval,
title={LongMemEval: Benchmarking Chat Assistants on Long-Term Interactive Memory},
author={Di Wu and Hongwei Wang and Wenhao Yu and Yuwei Zhang and Kai-Wei Chang and Dong Yu},
booktitle={The Thirteenth International Conference on Learning Representations},
url={https://openreview.net/forum?id=pZiyCaVuti}
}

@inproceedings{zhang2024toolbehonest,
    title = "{T}ool{B}e{H}onest: A Multi-level Hallucination Diagnostic Benchmark for Tool-Augmented Large Language Models",
    author = "Zhang, Yuxiang  and
      Chen, Jing  and
      Wang, Junjie  and
      Liu, Yaxin  and
      Yang, Cheng  and
      Shi, Chufan  and
      Zhu, Xinyu  and
      Lin, Zihao  and
      Wan, Hanwen  and
      Yang, Yujiu  and
      Sakai, Tetsuya  and
      Feng, Tian  and
      Yamana, Hayato",
    booktitle = "Proceedings of the 2024 Conference on Empirical Methods in Natural Language Processing",
    url = "https://aclanthology.org/2024.emnlp-main.637/",
}

@article{liu2023repobench,
  title={Repobench: Benchmarking repository-level code auto-completion systems},
  author={Liu, Tianyang and Xu, Canwen and McAuley, Julian},
  journal={arXiv preprint arXiv:2306.03091},
  year={2023}
}

@article{zhang2023repocoder,
  title={Repocoder: Repository-level code completion through iterative retrieval and generation},
  author={Zhang, Fengji and Chen, Bei and Zhang, Yue and Keung, Jacky and Liu, Jin and Zan, Daoguang and Mao, Yi and Lou, Jian-Guang and Chen, Weizhu},
  journal={arXiv preprint arXiv:2303.12570},
  year={2023}
}

@article{bogomolov2024long,
  title={Long code arena: a set of benchmarks for long-context code models},
  author={Bogomolov, Egor and Eliseeva, Aleksandra and Galimzyanov, Timur and Glukhov, Evgeniy and Shapkin, Anton and Tigina, Maria and Golubev, Yaroslav and Kovrigin, Alexander and van Deursen, Arie and Izadi, Maliheh and others},
  journal={arXiv preprint arXiv:2406.11612},
  year={2024}
}

@misc{biomedsql_nih_card_2025,
  title        = {BiomedSQL},
  author       = {Center for Alzheimer’s and Related Dementias (CARD)},
  year         = {2025},
  howpublished = {\url{https://huggingface.co/datasets/NIH-CARD/BiomedSQL}},
  note         = {Accessed: 2025-05-14},
  publisher    = {Hugging Face},
  license      = {Apache 2.0},
}

@misc{petroni2021kiltbenchmarkknowledgeintensive,
      title={KILT: a Benchmark for Knowledge Intensive Language Tasks}, 
      author={Fabio Petroni and Aleksandra Piktus and Angela Fan and Patrick Lewis and Majid Yazdani and Nicola De Cao and James Thorne and Yacine Jernite and Vladimir Karpukhin and Jean Maillard and Vassilis Plachouras and Tim Rocktäschel and Sebastian Riedel},
      year={2021},
      eprint={2009.02252},
      archivePrefix={arXiv},
      primaryClass={cs.CL},
      url={https://arxiv.org/abs/2009.02252}, 
}

@misc{yen2025helmetevaluatelongcontextlanguage,
      title={HELMET: How to Evaluate Long-Context Language Models Effectively and Thoroughly}, 
      author={Howard Yen and Tianyu Gao and Minmin Hou and Ke Ding and Daniel Fleischer and Peter Izsak and Moshe Wasserblat and Danqi Chen},
      year={2025},
      eprint={2410.02694},
      archivePrefix={arXiv},
      primaryClass={cs.CL},
      url={https://arxiv.org/abs/2410.02694}, 
}

@misc{openr1math220k,
  title        = {OpenR1-Math-220k},
  author       = {Open R1},
  year         = {2024},
  howpublished = {\url{https://huggingface.co/datasets/open-r1/OpenR1-Math-220k}},
  note         = {Accessed: 2025-05-15}
}

@misc{deepseekai2025deepseekr1incentivizingreasoningcapability,
      title={DeepSeek-R1: Incentivizing Reasoning Capability in LLMs via Reinforcement Learning}, 
      author={DeepSeek-AI and Daya Guo and Dejian Yang and Haowei Zhang and Junxiao Song and Ruoyu Zhang and Runxin Xu and Qihao Zhu and Shirong Ma and Peiyi Wang and Xiao Bi and Xiaokang Zhang and Xingkai Yu and Yu Wu and Z. F. Wu and Zhibin Gou and Zhihong Shao and Zhuoshu Li and Ziyi Gao and Aixin Liu and Bing Xue and Bingxuan Wang and Bochao Wu and Bei Feng and Chengda Lu and Chenggang Zhao and Chengqi Deng and Chenyu Zhang and Chong Ruan and Damai Dai and Deli Chen and Dongjie Ji and Erhang Li and Fangyun Lin and Fucong Dai and Fuli Luo and Guangbo Hao and Guanting Chen and Guowei Li and H. Zhang and Han Bao and Hanwei Xu and Haocheng Wang and Honghui Ding and Huajian Xin and Huazuo Gao and Hui Qu and Hui Li and Jianzhong Guo and Jiashi Li and Jiawei Wang and Jingchang Chen and Jingyang Yuan and Junjie Qiu and Junlong Li and J. L. Cai and Jiaqi Ni and Jian Liang and Jin Chen and Kai Dong and Kai Hu and Kaige Gao and Kang Guan and Kexin Huang and Kuai Yu and Lean Wang and Lecong Zhang and Liang Zhao and Litong Wang and Liyue Zhang and Lei Xu and Leyi Xia and Mingchuan Zhang and Minghua Zhang and Minghui Tang and Meng Li and Miaojun Wang and Mingming Li and Ning Tian and Panpan Huang and Peng Zhang and Qiancheng Wang and Qinyu Chen and Qiushi Du and Ruiqi Ge and Ruisong Zhang and Ruizhe Pan and Runji Wang and R. J. Chen and R. L. Jin and Ruyi Chen and Shanghao Lu and Shangyan Zhou and Shanhuang Chen and Shengfeng Ye and Shiyu Wang and Shuiping Yu and Shunfeng Zhou and Shuting Pan and S. S. Li and Shuang Zhou and Shaoqing Wu and Shengfeng Ye and Tao Yun and Tian Pei and Tianyu Sun and T. Wang and Wangding Zeng and Wanjia Zhao and Wen Liu and Wenfeng Liang and Wenjun Gao and Wenqin Yu and Wentao Zhang and W. L. Xiao and Wei An and Xiaodong Liu and Xiaohan Wang and Xiaokang Chen and Xiaotao Nie and Xin Cheng and Xin Liu and Xin Xie and Xingchao Liu and Xinyu Yang and Xinyuan Li and Xuecheng Su and Xuheng Lin and X. Q. Li and Xiangyue Jin and Xiaojin Shen and Xiaosha Chen and Xiaowen Sun and Xiaoxiang Wang and Xinnan Song and Xinyi Zhou and Xianzu Wang and Xinxia Shan and Y. K. Li and Y. Q. Wang and Y. X. Wei and Yang Zhang and Yanhong Xu and Yao Li and Yao Zhao and Yaofeng Sun and Yaohui Wang and Yi Yu and Yichao Zhang and Yifan Shi and Yiliang Xiong and Ying He and Yishi Piao and Yisong Wang and Yixuan Tan and Yiyang Ma and Yiyuan Liu and Yongqiang Guo and Yuan Ou and Yuduan Wang and Yue Gong and Yuheng Zou and Yujia He and Yunfan Xiong and Yuxiang Luo and Yuxiang You and Yuxuan Liu and Yuyang Zhou and Y. X. Zhu and Yanhong Xu and Yanping Huang and Yaohui Li and Yi Zheng and Yuchen Zhu and Yunxian Ma and Ying Tang and Yukun Zha and Yuting Yan and Z. Z. Ren and Zehui Ren and Zhangli Sha and Zhe Fu and Zhean Xu and Zhenda Xie and Zhengyan Zhang and Zhewen Hao and Zhicheng Ma and Zhigang Yan and Zhiyu Wu and Zihui Gu and Zijia Zhu and Zijun Liu and Zilin Li and Ziwei Xie and Ziyang Song and Zizheng Pan and Zhen Huang and Zhipeng Xu and Zhongyu Zhang and Zhen Zhang},
      year={2025},
      eprint={2501.12948},
      archivePrefix={arXiv},
      primaryClass={cs.CL},
      url={https://arxiv.org/abs/2501.12948}, 
}

@misc{openai2024gpt4ocard,
      title={GPT-4o System Card}, 
      author={OpenAI and : and Aaron Hurst and Adam Lerer and Adam P. Goucher and Adam Perelman and Aditya Ramesh and Aidan Clark and AJ Ostrow and Akila Welihinda and Alan Hayes and Alec Radford and Aleksander Mądry and Alex Baker-Whitcomb and Alex Beutel and Alex Borzunov and Alex Carney and Alex Chow and Alex Kirillov and Alex Nichol and Alex Paino and Alex Renzin and Alex Tachard Passos and Alexander Kirillov and Alexi Christakis and Alexis Conneau and Ali Kamali and Allan Jabri and Allison Moyer and Allison Tam and Amadou Crookes and Amin Tootoochian and Amin Tootoonchian and Ananya Kumar and Andrea Vallone and Andrej Karpathy and Andrew Braunstein and Andrew Cann and Andrew Codispoti and Andrew Galu and Andrew Kondrich and Andrew Tulloch and Andrey Mishchenko and Angela Baek and Angela Jiang and Antoine Pelisse and Antonia Woodford and Anuj Gosalia and Arka Dhar and Ashley Pantuliano and Avi Nayak and Avital Oliver and Barret Zoph and Behrooz Ghorbani and Ben Leimberger and Ben Rossen and Ben Sokolowsky and Ben Wang and Benjamin Zweig and Beth Hoover and Blake Samic and Bob McGrew and Bobby Spero and Bogo Giertler and Bowen Cheng and Brad Lightcap and Brandon Walkin and Brendan Quinn and Brian Guarraci and Brian Hsu and Bright Kellogg and Brydon Eastman and Camillo Lugaresi and Carroll Wainwright and Cary Bassin and Cary Hudson and Casey Chu and Chad Nelson and Chak Li and Chan Jun Shern and Channing Conger and Charlotte Barette and Chelsea Voss and Chen Ding and Cheng Lu and Chong Zhang and Chris Beaumont and Chris Hallacy and Chris Koch and Christian Gibson and Christina Kim and Christine Choi and Christine McLeavey and Christopher Hesse and Claudia Fischer and Clemens Winter and Coley Czarnecki and Colin Jarvis and Colin Wei and Constantin Koumouzelis and Dane Sherburn and Daniel Kappler and Daniel Levin and Daniel Levy and David Carr and David Farhi and David Mely and David Robinson and David Sasaki and Denny Jin and Dev Valladares and Dimitris Tsipras and Doug Li and Duc Phong Nguyen and Duncan Findlay and Edede Oiwoh and Edmund Wong and Ehsan Asdar and Elizabeth Proehl and Elizabeth Yang and Eric Antonow and Eric Kramer and Eric Peterson and Eric Sigler and Eric Wallace and Eugene Brevdo and Evan Mays and Farzad Khorasani and Felipe Petroski Such and Filippo Raso and Francis Zhang and Fred von Lohmann and Freddie Sulit and Gabriel Goh and Gene Oden and Geoff Salmon and Giulio Starace and Greg Brockman and Hadi Salman and Haiming Bao and Haitang Hu and Hannah Wong and Haoyu Wang and Heather Schmidt and Heather Whitney and Heewoo Jun and Hendrik Kirchner and Henrique Ponde de Oliveira Pinto and Hongyu Ren and Huiwen Chang and Hyung Won Chung and Ian Kivlichan and Ian O'Connell and Ian O'Connell and Ian Osband and Ian Silber and Ian Sohl and Ibrahim Okuyucu and Ikai Lan and Ilya Kostrikov and Ilya Sutskever and Ingmar Kanitscheider and Ishaan Gulrajani and Jacob Coxon and Jacob Menick and Jakub Pachocki and James Aung and James Betker and James Crooks and James Lennon and Jamie Kiros and Jan Leike and Jane Park and Jason Kwon and Jason Phang and Jason Teplitz and Jason Wei and Jason Wolfe and Jay Chen and Jeff Harris and Jenia Varavva and Jessica Gan Lee and Jessica Shieh and Ji Lin and Jiahui Yu and Jiayi Weng and Jie Tang and Jieqi Yu and Joanne Jang and Joaquin Quinonero Candela and Joe Beutler and Joe Landers and Joel Parish and Johannes Heidecke and John Schulman and Jonathan Lachman and Jonathan McKay and Jonathan Uesato and Jonathan Ward and Jong Wook Kim and Joost Huizinga and Jordan Sitkin and Jos Kraaijeveld and Josh Gross and Josh Kaplan and Josh Snyder and Joshua Achiam and Joy Jiao and Joyce Lee and Juntang Zhuang and Justyn Harriman and Kai Fricke and Kai Hayashi and Karan Singhal and Katy Shi and Kavin Karthik and Kayla Wood and Kendra Rimbach and Kenny Hsu and Kenny Nguyen and Keren Gu-Lemberg and Kevin Button and Kevin Liu and Kiel Howe and Krithika Muthukumar and Kyle Luther and Lama Ahmad and Larry Kai and Lauren Itow and Lauren Workman and Leher Pathak and Leo Chen and Li Jing and Lia Guy and Liam Fedus and Liang Zhou and Lien Mamitsuka and Lilian Weng and Lindsay McCallum and Lindsey Held and Long Ouyang and Louis Feuvrier and Lu Zhang and Lukas Kondraciuk and Lukasz Kaiser and Luke Hewitt and Luke Metz and Lyric Doshi and Mada Aflak and Maddie Simens and Madelaine Boyd and Madeleine Thompson and Marat Dukhan and Mark Chen and Mark Gray and Mark Hudnall and Marvin Zhang and Marwan Aljubeh and Mateusz Litwin and Matthew Zeng and Max Johnson and Maya Shetty and Mayank Gupta and Meghan Shah and Mehmet Yatbaz and Meng Jia Yang and Mengchao Zhong and Mia Glaese and Mianna Chen and Michael Janner and Michael Lampe and Michael Petrov and Michael Wu and Michele Wang and Michelle Fradin and Michelle Pokrass and Miguel Castro and Miguel Oom Temudo de Castro and Mikhail Pavlov and Miles Brundage and Miles Wang and Minal Khan and Mira Murati and Mo Bavarian and Molly Lin and Murat Yesildal and Nacho Soto and Natalia Gimelshein and Natalie Cone and Natalie Staudacher and Natalie Summers and Natan LaFontaine and Neil Chowdhury and Nick Ryder and Nick Stathas and Nick Turley and Nik Tezak and Niko Felix and Nithanth Kudige and Nitish Keskar and Noah Deutsch and Noel Bundick and Nora Puckett and Ofir Nachum and Ola Okelola and Oleg Boiko and Oleg Murk and Oliver Jaffe and Olivia Watkins and Olivier Godement and Owen Campbell-Moore and Patrick Chao and Paul McMillan and Pavel Belov and Peng Su and Peter Bak and Peter Bakkum and Peter Deng and Peter Dolan and Peter Hoeschele and Peter Welinder and Phil Tillet and Philip Pronin and Philippe Tillet and Prafulla Dhariwal and Qiming Yuan and Rachel Dias and Rachel Lim and Rahul Arora and Rajan Troll and Randall Lin and Rapha Gontijo Lopes and Raul Puri and Reah Miyara and Reimar Leike and Renaud Gaubert and Reza Zamani and Ricky Wang and Rob Donnelly and Rob Honsby and Rocky Smith and Rohan Sahai and Rohit Ramchandani and Romain Huet and Rory Carmichael and Rowan Zellers and Roy Chen and Ruby Chen and Ruslan Nigmatullin and Ryan Cheu and Saachi Jain and Sam Altman and Sam Schoenholz and Sam Toizer and Samuel Miserendino and Sandhini Agarwal and Sara Culver and Scott Ethersmith and Scott Gray and Sean Grove and Sean Metzger and Shamez Hermani and Shantanu Jain and Shengjia Zhao and Sherwin Wu and Shino Jomoto and Shirong Wu and Shuaiqi and Xia and Sonia Phene and Spencer Papay and Srinivas Narayanan and Steve Coffey and Steve Lee and Stewart Hall and Suchir Balaji and Tal Broda and Tal Stramer and Tao Xu and Tarun Gogineni and Taya Christianson and Ted Sanders and Tejal Patwardhan and Thomas Cunninghman and Thomas Degry and Thomas Dimson and Thomas Raoux and Thomas Shadwell and Tianhao Zheng and Todd Underwood and Todor Markov and Toki Sherbakov and Tom Rubin and Tom Stasi and Tomer Kaftan and Tristan Heywood and Troy Peterson and Tyce Walters and Tyna Eloundou and Valerie Qi and Veit Moeller and Vinnie Monaco and Vishal Kuo and Vlad Fomenko and Wayne Chang and Weiyi Zheng and Wenda Zhou and Wesam Manassra and Will Sheu and Wojciech Zaremba and Yash Patil and Yilei Qian and Yongjik Kim and Youlong Cheng and Yu Zhang and Yuchen He and Yuchen Zhang and Yujia Jin and Yunxing Dai and Yury Malkov},
      year={2024},
      eprint={2410.21276},
      archivePrefix={arXiv},
      primaryClass={cs.CL},
      url={https://arxiv.org/abs/2410.21276}, 
}

@misc{openai2024gpt4omini,
  author       = {OpenAI},
  title        = {GPT-4o mini: advancing cost-efficient intelligence},
  year         = {2024},
  month        = {July},
  note         = {\url{https://openai.com/research/gpt-4o-mini-advancing-cost-efficient-intelligence}},
  institution  = {OpenAI},
  howpublished = {\url{https://openai.com/research/gpt-4o-mini-advancing-cost-efficient-intelligence}}
}

@misc{gemini2025flash,
  author       = {Shrestha Basu Mallick and Logan Kilpatrick},
  title        = {Gemini 2.0: Flash, Flash-Lite and Pro},
  howpublished = {\url{https://developers.google.com/updates/gemini-2-0-flash-flash-lite-pro}},
  note         = {Google for Developers Blog, February 5, 2025},
  year         = {2025},
  month        = feb,
  institution  = {Google},
}

@inproceedings{macavaney:sigir2021-irds,
  author = {MacAvaney, Sean and Yates, Andrew and Feldman, Sergey and Downey, Doug and Cohan, Arman and Goharian, Nazli},
  title = {Simplified Data Wrangling with ir\_datasets},
  year = {2021},
  booktitle = {SIGIR}
}

@misc{tiktoken,
  author       = {OpenAI},
  title        = {tiktoken: A fast BPE tokenizer for use with OpenAI's models},
  year         = {2023},
  howpublished = {\url{https://github.com/openai/tiktoken}},
  note         = {Accessed: 2025-05-15}
}

@article{pubtator3,
    author = {Wei, Chih-Hsuan and Allot, Alexis and Lai, Po-Ting and Leaman, Robert and Tian, Shubo and Luo, Ling and Jin, Qiao and Wang, Zhizheng and Chen, Qingyu and Lu, Zhiyong},
    title = {PubTator 3.0: an AI-powered literature resource for unlocking biomedical knowledge},
    journal = {Nucleic Acids Research},
    volume = {52},
    number = {W1},
    pages = {W540-W546},
    year = {2024},
    month = {04},
    issn = {0305-1048},
    doi = {10.1093/nar/gkae235},
    url = {https://doi.org/10.1093/nar/gkae235},
    eprint = {https://academic.oup.com/nar/article-pdf/52/W1/W540/58436124/gkae235.pdf},
}

@article{Stenson2020HGMD,
  title        = {The Human Gene Mutation Database (HGMD®): optimizing its use in a clinical diagnostic or research setting},
  author       = {Stenson, Peter D. and Mort, Matthew and Ball, Edward V. and Chapman, Molly and Evans, Katy and Azevedo, Luisa and Hayden, Matthew and Heywood, Sally and Millar, David S. and Phillips, Andrew D. and Cooper, David N.},
  journal      = {Human Genetics},
  volume       = {139},
  number       = {10},
  pages        = {1197--1207},
  year         = {2020},
  doi          = {10.1007/s00439-020-02199-3},
  url          = {https://doi.org/10.1007/s00439-020-02199-3}
}

@misc{ncbi2025,
  author       = {{National Center for Biotechnology Information (NCBI)}},
  title        = {NCBI [Internet]},
  howpublished = {\url{https://www.ncbi.nlm.nih.gov/}},
  note         = {Bethesda (MD): National Library of Medicine (US), National Center for Biotechnology Information; [1988]– [cited 2025 May 20]},
  year         = {1988}
}

@misc{biowulf,
  author = {{NIH Biowulf}},
  title = {NIH High-Performance Computing (HPC) Biowulf Cluster},
  year = {2024},
  howpublished = {\url{https://hpc.nih.gov}},
  note = {Accessed May 2025}
}

@software{huggingface2025mathverify,
  author       = {Hugging Face},
  title        = {Math-Verify: A Robust Mathematical Expression Evaluation System},
  year         = {2025},
  publisher    = {GitHub},
  url          = {https://github.com/huggingface/Math-Verify},
  version      = {0.6.2}
}

@misc{dai2024deniahlincontextfeaturesinfluence,
      title={DENIAHL: In-Context Features Influence LLM Needle-In-A-Haystack Abilities}, 
      author={Hui Dai and Dan Pechi and Xinyi Yang and Garvit Banga and Raghav Mantri},
      year={2024},
      eprint={2411.19360},
      archivePrefix={arXiv},
      primaryClass={cs.CL},
      url={https://arxiv.org/abs/2411.19360}, 
}

@misc{abdin2025phi4reasoningtechnicalreport,
      title={Phi-4-reasoning Technical Report}, 
      author={Marah Abdin and Sahaj Agarwal and Ahmed Awadallah and Vidhisha Balachandran and Harkirat Behl and Lingjiao Chen and Gustavo de Rosa and Suriya Gunasekar and Mojan Javaheripi and Neel Joshi and Piero Kauffmann and Yash Lara and Caio César Teodoro Mendes and Arindam Mitra and Besmira Nushi and Dimitris Papailiopoulos and Olli Saarikivi and Shital Shah and Vaishnavi Shrivastava and Vibhav Vineet and Yue Wu and Safoora Yousefi and Guoqing Zheng},
      year={2025},
      eprint={2504.21318},
      archivePrefix={arXiv},
      primaryClass={cs.AI},
      url={https://arxiv.org/abs/2504.21318}, 
}
\bibliographystyle{iclr2026_conference}

\newpage

\appendix
\section{Appendix}
\appendix

\section{Extended Experimental Details}

We provide extended experimental details on benchmark construction, model configuration, and evaluation methodology to support the reproducibility and interpretability of our results.

\subsection{Original Benchmarks}
\label{app:benchmarks}

We describe the sources, licenses, and preprocessing procedures for each of the three adapted benchmarks used in our experiments. All experiments were run on a sampled subset of 250 examples per benchmark. See the code repository for exact methodology.

\textbf{CARDBiomedBench}
\begin{itemize}
    \item \textbf{Source:} CARDBiomedBench on \href{https://huggingface.co/datasets/NIH-CARD/CARDBiomedBench}{Hugging Face} and its BiomedSQL variant on \href{https://huggingface.co/datasets/NIH-CARD/BiomedSQL}{Hugging Face}. Distractor documents were retrieved using a multi-agent retrieval system, which retrieves content from: (1) Google search over NIH domains, (2) PubTator3.0~\citep{pubtator3}, (3) the Human Gene Mutation Database (HGMD)~\citep{Stenson2020HGMD}, and (4) NCBI gene and variant pages~\citep{ncbi2025}.
    \item \textbf{License:} Apache 2.0 for benchmark code and data. Some distractor sources (e.g., HGMD) are not redistributable but are publicly accessible on their respective platforms.
    \item \textbf{Preprocessing:} None, the distractor and gold contexts are as-is from the retriever.
\end{itemize}

\textbf{NaturalQuestions}
\begin{itemize}
    \item \textbf{Source:} NQ with evidence spans aligned to Knowledge Intensive Language Tasks (KILT) on \href{https://huggingface.co/datasets/facebook/kilt_tasks}{Hugging Face}. Gold contexts were loaded using the Ai2 ir\_datasets python package~\citep{macavaney:sigir2021-irds} and distractors were sourced from HELMET on \href{https://huggingface.co/datasets/princeton-nlp/HELMET}{Hugging Face}.
    \item \textbf{License:} Creative Commons Share-Alike 3.0 (NQ), MIT (KILT \& HELMET), and Apache 2.0 (ir\_datasets).
    \item \textbf{Preprocessing:} We filtered for validation examples that had matching HELMET distractors. Examples with missing KILT provenance, absent or unresolvable answer spans, or malformed metadata were excluded. Gold and distractor documents included the title of the article \textit{`Title: \{title\} Document: \{gold\_document\}'} to give them context.
\end{itemize}

\textbf{NuminaMath1.5}
\begin{itemize}
    \item \textbf{Source:} NuminaMath1.5 (NM) and its OpenR1Math (OR1M) variant on \href{https://huggingface.co/datasets/open-r1/OpenR1-Math-220k}{Hugging Face}, that contains DeepSeekR1 reasoning chains.
    \item \textbf{License:} Apache 2.0. (NM and OR1M).
    \item \textbf{Preprocessing:} Filtered to retain only examples with `complete' and `verified' fields for question, final answer, structured solution, and long-form generation. DeepSeekR1 generations were truncated to the final 5,000 tokens using GPT-4o tiktoken~\citep{tiktoken} tokenization to normalize document length across tasks. Distractors sampling was among the other questions and excluded duplicates. Sizes of gold and distractors were strung into a pseudo-document by including \textit{`The answer to \{question\} is \{gold\_document\}'} to give them context.
\end{itemize}

\newpage

\subsection{Task Creation}
\label{appendix:task:creation}

\begin{figure}[th]
    \centering
    \small
    \includegraphics[width=1.0\textwidth,trim=0cm 0cm 0cm 0cm]{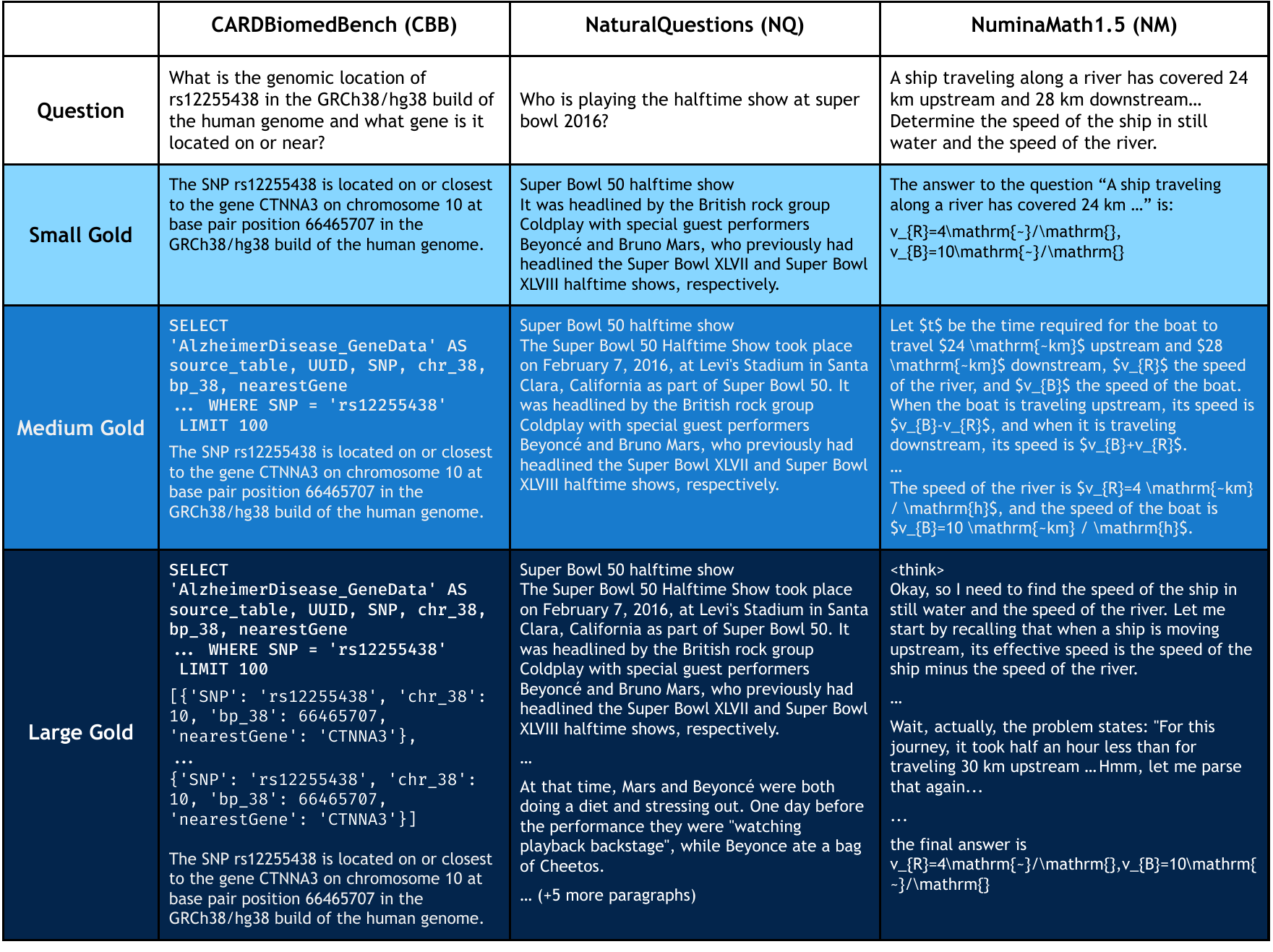}
    
    \caption{Gold context construction across benchmarks. The \lightbluetext{``small''} gold context is minimally sufficient to answer the question; \mediumbluetext{``medium''} and \darkbluetext{``large''} add further relevant information. In CARDBiomedBench (left), this includes SQL and result rows; in NQ (center), adjacent Wikipedia paragraphs; in NM (right), full solution traces and DeepSeekR1 reasoning chain.}
    \label{fig:gold_sizes_examples}
    
\end{figure}

\changed{Here we provide details on the gold context construction as discussed in \S\ref{subsec:task:construction}. Specifically, in Figure~\ref{fig:gold_sizes_examples} we show examples of the gold contexts, of different sizes, for each dataset.}


\subsection{LLM Configuration}
\label{app:llm-config}

We evaluated seven LLMs, each configured via provider-specific APIs. All evaluations were conducted as deterministically as possible.

\textbf{API Providers.}
We used the following service providers for model access:

\begin{itemize}
    \item \textbf{GPT models} (o3-mini, GPT-4o, GPT-4o-mini) were accessed via the \texttt{Azure OpenAI} service.
    \item \textbf{Gemini models} (Gemini-2.5-Flash, Gemini-2.0-Flash, Gemini-2.0-Flash-Lite) were accessed via the \texttt{Google AI GenAI SDK}, using the official \texttt{genai} Python client.
    \item \textbf{DeepSeek-R1 and Phi-4-reasoning} were accessed via the \texttt{Azure AI Inference} service.
    \item \textbf{LLaMA models >= 70b params} (Meta-LLaMA-3.1-405B-Instruct, LLaMA-3.3-70B-Instruct) were accessed via the \texttt{Azure AI Inference} service.
    \item \textbf{LLaMA model < 70B parameters} (LLaMA-3.1-8B-Instruct) was evaluated locally using the \texttt{meta-llama/Llama-3.1-8B-Instruct} checkpoint, loaded via Hugging Face \texttt{transformers}. All local evaluations were conducted on the NIH High-Performance Computing (HPC) Biowulf cluster~\citep{biowulf}, leveraging GPU nodes for inference.
\end{itemize}

\textbf{Prompting and Evaluation Configuration.}
Prompts were benchmark-specific and standardized across model types. All non-reasoning models were queried with \texttt{max\_tokens=256} and \texttt{temperature=0.0}. Provider-specific configurations (e.g., safety settings for Google GenAI, and device mapping for HuggingFace) were handled automatically during model initialization. See the code and YAML config files for full details. Reasoning models were queried with \texttt{max\_tokens=2048} and their default generation params, to allow for reasoning. Reasoning models were additionally given instructions to encourage grounding their answer in the retrieved documents, to prevent relying on internal knowledge.

\textbf{Grading LLMs.}
For CARDBiomedBench, an additional grading LLM was used to assess answer correctness via BioScore using GPT-4o, as done by the authors. It was instantiated using the same infrastructure and configurations as the primary LLMs, with \texttt{max\_tokens=10}.

\subsection{Metrics}
\label{app:metrics}

We used evaluation metrics that align with the original datasets' scoring protocols:

\textbf{Quality Rate.} We evaluate responses to the CBB tasks following their proposed \textbf{BioScore} framework, an LLM-as-a-judge metric implemented with GPT-4o. Each response is scored on a 3-point scale according to the BioScore prompt \ref{prompt:bioscore}, and a score $\geq 2$ is considered factually correct. The \textbf{Quality Rate} is computed as the proportion of responses meeting this threshold.

Formally, given a reference set $Resp$ of expert-annotated responses and a corresponding set $\hat{Resp}$ of model-generated responses for $n$ questions:

\begin{equation}
\text{Quality Rate} = \frac{1}{N} \sum_{n=1}^{N} Correct(r_n, \hat{r}_n)
\end{equation}

\begin{equation}
\text{where} \quad Correct(r_n, \hat{r}_n) =
\begin{cases}
1, & \text{if} \quad \text{BioScore}(r_n, \hat{r}_n) \geq 2 \\
0, & \text{otherwise}
\end{cases}
\quad \text{and} \quad r_n \in Resp, \hat{r}_n \in \hat{Resp}
\end{equation}

\textbf{SubEM.} For NQ we utilized substring exact match, which assigns a score of 1.0 if any normalized ground truth string is a subspan of the model's response (after normalization), and 0 otherwise. This is a correctness signal used by previous work on this data.

\textbf{math-verify.} Evaluated with math-verify~\citep{huggingface2025mathverify}, a symbolic equivalence checker that parses LaTeX boxed answers and verifies correctness through structured math expression comparison. Parsing and verification are done using an extraction and comparison pipeline derived from the Math-Verify toolkit.

\textbf{Error Bars.} All plots showing aggregate scores (e.g., Figure~\ref{fig:two_by_three_grid}) report 90\% confidence intervals (CIs) estimated via non-parametric bootstrapping over tasks. Given $N$ scores, we resample with replacement 1,000 times and compute the middle 90\% interval from the resulting bootstrap distribution.

\newpage

\subsection{Prompts.}

We show prompts used to collect results from the models and the BioScore grading prompt. There is a unique prompt for each benchmark, which is used on every model. \{Variables\} are in curly braces which are formatted with task data (question and documents). We encourage models to ground their answers in the context and abstain if unable to answer.

\begin{promptbox}
You are a highly knowledgeable and experienced expert in the healthcare and biomedical field, 
possessing extensive medical knowledge and practical expertise. Create an answer to the question 
using only the provided documents (some of which might be irrelevant). If you cannot answer the 
question based on the documents, explicitly state that you do not know.
Question: {question}
Documents: {documents}
\end{promptbox}
\begingroup
  \setlength{\abovecaptionskip}{6pt}
  \setlength{\belowcaptionskip}{10pt}
  \captionof{prompt}{The CARDBiomedBench prompt is adapted from the original paper's experimental methods and includes mention of biomedical expertise.}
  \label{prompt:cbb_prompt}
\endgroup

\begin{promptbox}
Create an answer to the question using only the provided documents (some of which might be
irrelevant). If you cannot answer the question based on the documents, explicitly state that 
you do not know.
Question: {question}
Documents: {documents}
\end{promptbox}
\begingroup
  \setlength{\abovecaptionskip}{6pt}
  \setlength{\belowcaptionskip}{10pt}
  \captionof{prompt}{The NaturalQuestions prompt is adapted from previous work's experimental methods~\citep{liu2024lost, yen2025helmetevaluatelongcontextlanguage}.}
  \label{prompt:nq_prompt}
\endgroup

\begin{promptbox}
Create an ANSWER to the QUESTION using only the provided DOCUMENTS (some of which might be
irrelevant). Write nothing but your final answer in LaTeX within \\boxed{}. If you do not 
know the answer to a question, explicitly state so in \\boxed{I don't know}. 
QUESTION: {question}
DOCUMENTS: {documents}
QUESTION: {question}
ANSWER:
\end{promptbox}
\begingroup
  \setlength{\abovecaptionskip}{6pt}
  \setlength{\belowcaptionskip}{10pt}
  \captionof{prompt}{The NuminaMath1.5 prompt uniquely repeats the question and has formatting guidelines, to comply with the math-verify metric. Without repeating the question models exhibited extremely poor performance in every configuration.}
  \label{prompt:nm_prompt}
\endgroup

\clearpage

\begin{promptbox}
You are a highly knowledgeable and experienced expert in the healthcare and biomedical field, possessing 
extensive medical knowledge and practical expertise.
### Scoring Instructions for Evaluating Analyst Responses

**Objective:** Evaluate an analyst's response against a gold standard.

**Scoring Criteria:**
    - **Exact Match:** 3 points for an exact or equally accurate response.
    - **Close Match:** 2 points for a very close response with minor inaccuracies.
    - **Partial Match:** 1 point for a partially accurate response with significant omissions.
    - **Irrelevant Information (Harmless):** Deduct 0.5 points for harmless irrelevant information.
    - **Irrelevant Information (Distracting):** Deduct 1 point for distracting irrelevant information.
    - **No Match:** 0 points for no match.
    - **Not Knowing Response:** -1 point for stating lack of knowledge or abstaining. An example
    of this scenario is when Analyst Response says \'There are various studies, resources or
    databases on this topic that you can check ... but I do not have enough information on this topic.

**Scoring Process:**
    1. **Maximum Score:** 3 points per question.
    2. **Calculate Score:** Apply criteria to evaluate the response.
    
**Question:** {question}
**Golden Answer:** {gold_ans}
**Analyst Response:** {pred_ans}

## Your grading
Using the scoring instructions above, grade the Analyst Response return only the numeric score
on a scale from 0.0-3.0. If the response is stating lack of knowledge or abstaining, give it
-1.0.
\end{promptbox}
\begingroup
  \setlength{\abovecaptionskip}{6pt}
  \setlength{\belowcaptionskip}{10pt}
  \captionof{prompt}{BioScore grading prompt for LLM-as-a-judge on CBB tasks, awarding points for correct information and deducting points for incorrect information. It differentiates an abstention (-1) from an incorrect answer (0 or 1).}
  \label{prompt:bioscore}
\endgroup

\newpage

\section{Extended Results}
\label{appendix:extended}

\changed{As discussed in \S\ref{subsec:baselines} we evaluated \textit{gold-only} baselines where each gold context size (small, medium, large) was presented alone, without distractors. This verified that all variants were independently sufficient to solve the task and that downstream performance drops are due to aggregation effects (e.g., distractor interference or gold placement). We provide  baselines for all benchmarks in \ref{apdx:baselines} Figure~\ref{fig:baseline_heatmap}.
} Additionally, we provide full results for CBB in Figure~\ref{fig:cbb_heatmap}, for NQ in Figure~\ref{fig:nq_heatmap}, and for NM in Figure~\ref{fig:nm_heatmap}. Finally, we provide positional curves for all models in Figure~\ref{fig:byposition_all_models_supps}.

\clearpage

\clearpage
\subsection{Performance by Position}
\label{appdx:by_position}

\begin{figure}[ht]
    \centering
    \tiny
    \setlength{\tabcolsep}{2pt}
    \begin{tabular}{cccc}
         \lightpurpletext{CARDBiomedBench} & \lightpurpletext{NaturalQuestions} & \lightpurpletext{NuminaMath1.5} & 
    \\
    \begin{subfigure}{0.329\textwidth}
        \includegraphics[width=0.9\linewidth]{images/cbb/gemini-2.0-flash/gemini-2.0-flash_CBB_QR.png}
    \end{subfigure} 
    & 
    \begin{subfigure}{0.329\textwidth}
        \includegraphics[width=0.9\linewidth]{images/nq/gemini-2.0-flash/gemini-2.0-flash_NQ_subEM_200distractors.png}
    \end{subfigure}
    & 
    \begin{subfigure}{0.329\textwidth}
        \includegraphics[width=0.9\linewidth]{images/nm/gemini-2.0-flash/gemini-2.0-flash_NM_math-verify_5distractors.png}
    \end{subfigure}
    & 
    \hspace{-0.8cm} \raisebox{13ex}{\rotatebox[origin=c]{270}{\lightpurpletext{Gemini-2.0-Flash}}} 
    \\  
    \begin{subfigure}{0.329\textwidth}
        \includegraphics[width=0.9\linewidth]{images/cbb/gemini-2.0-flash-lite/gemini-2.0-flash-lite_CBB_QR.png}
    \end{subfigure} 
    & 
    \begin{subfigure}{0.329\textwidth}
        \includegraphics[width=0.9\linewidth]{images/nq/gemini-2.0-flash-lite/gemini-2.0-flash-lite_NQ_subEM_200distractors.png}
    \end{subfigure}
    & 
    \begin{subfigure}{0.329\textwidth}
        \includegraphics[width=0.9\linewidth]{images/nm/gemini-2.0-flash-lite/gemini-2.0-flash-lite_NM_math-verify_5distractors.png}
    \end{subfigure}
    & 
    \hspace{-0.8cm} \raisebox{13ex}{\rotatebox[origin=c]{270}{\lightpurpletext{Gemini-2.0-Flash-Lite}}} 
    \\
    \begin{subfigure}{0.329\textwidth}
        \includegraphics[width=0.9\linewidth]{images/cbb/gpt-4o/gpt-4o_CBB_QR.png}
    \end{subfigure}     
     & 
    \begin{subfigure}{0.329\textwidth}
        \includegraphics[width=0.9\linewidth]{images/nq/gpt-4o/gpt-4o_NQ_subEM_200distractors.png}
    \end{subfigure}
     & 
    \begin{subfigure}{0.329\textwidth}
        \includegraphics[width=0.9\linewidth]{images/nm/gpt-4o/gpt-4o_NM_math-verify_5distractors.png}
    \end{subfigure}
    & \hspace{-0.8cm} \raisebox{9ex}{\rotatebox[origin=r]{270}{\lightpurpletext{GPT-4o}}}
    \\
    \begin{subfigure}{0.329\textwidth}
        \includegraphics[width=0.9\linewidth]{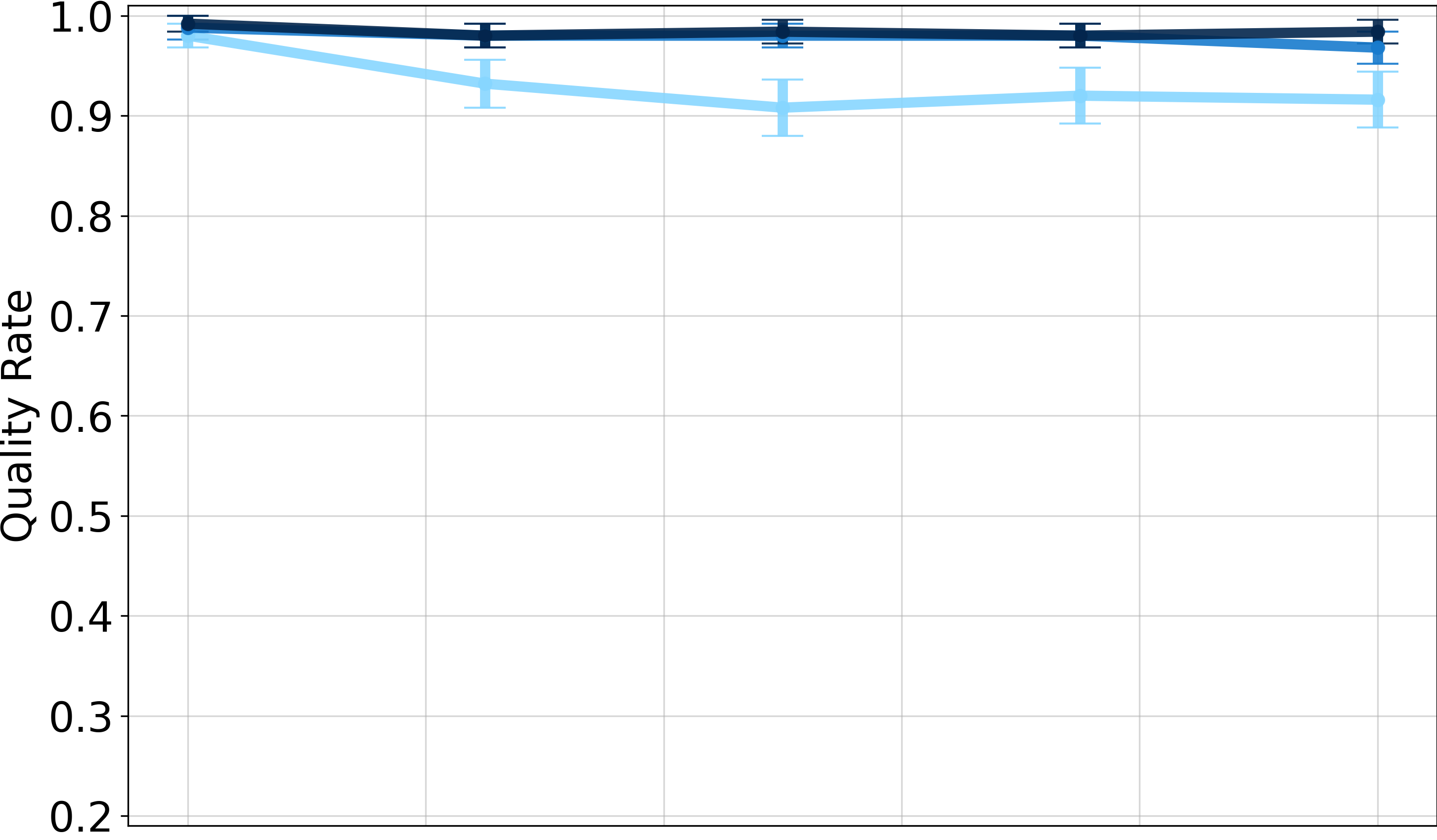}
    \end{subfigure}     
     & 
    \begin{subfigure}{0.329\textwidth}
        \includegraphics[width=0.9\linewidth]{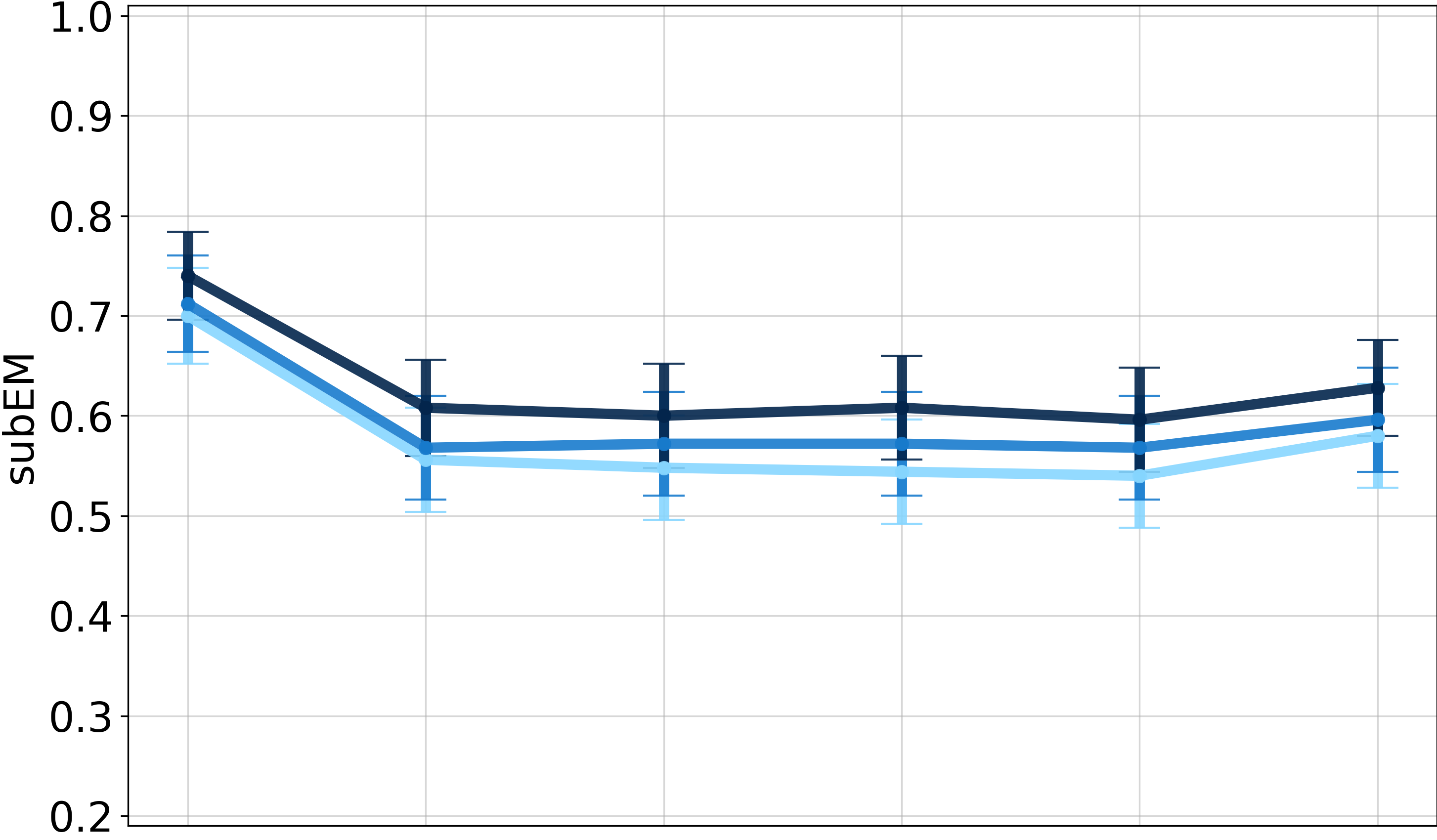}
    \end{subfigure}
     & 
    \begin{subfigure}{0.329\textwidth}
        \includegraphics[width=0.9\linewidth]{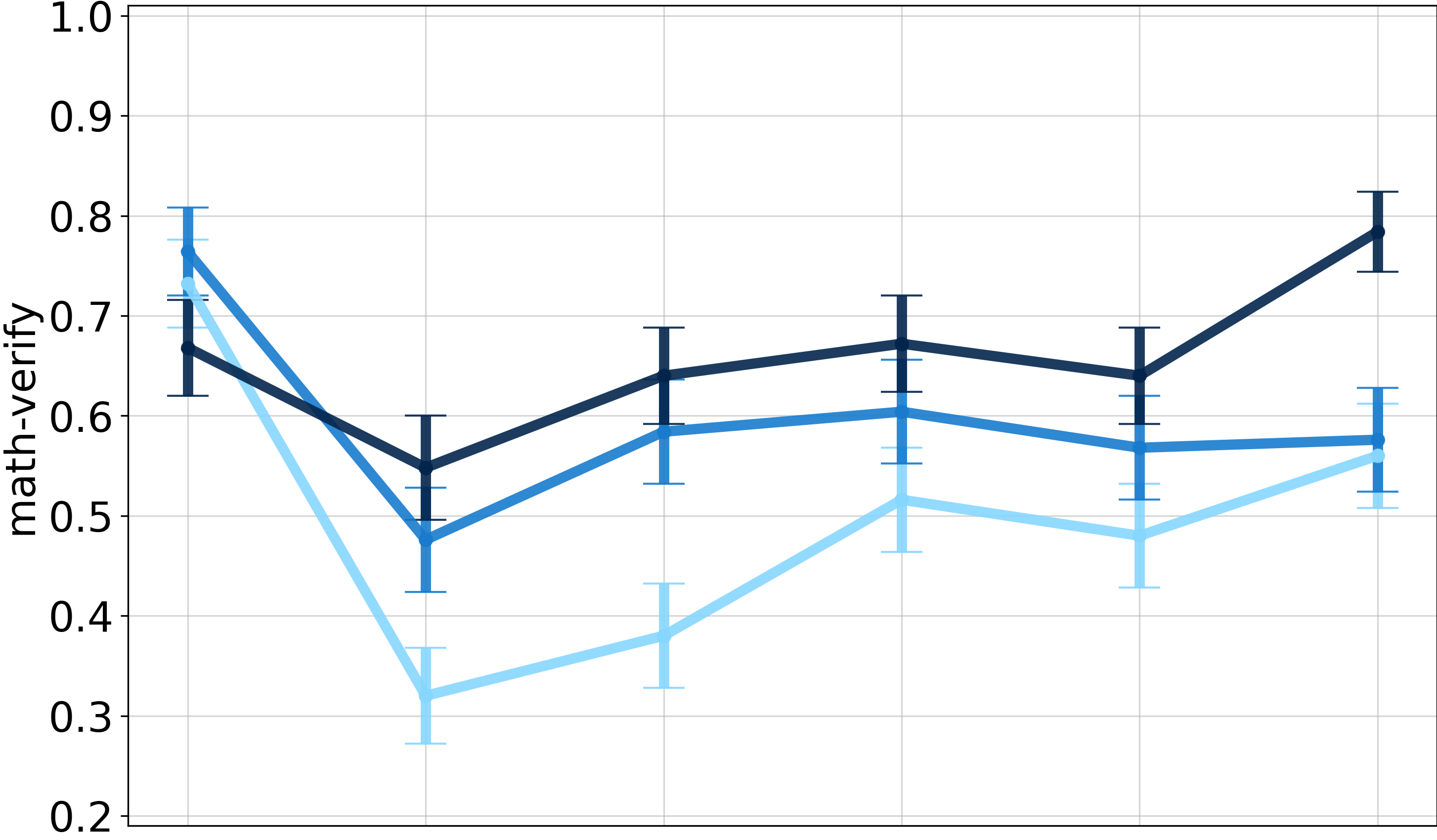}
    \end{subfigure}
    & \hspace{-0.8cm} \raisebox{9ex}{\rotatebox[origin=r]{270}{\lightpurpletext{GPT-4o-Mini}}}
     \\ 
    \begin{subfigure}{0.329\textwidth}
        \includegraphics[width=0.9\linewidth]{images/cbb/Meta-Llama-3.1-405B-Instruct/Meta-Llama-3.1-405B-Instruct_CBB_QR.png}
    \end{subfigure}
    & 
    \begin{subfigure}{0.329\textwidth}
        \includegraphics[width=0.9\linewidth]{images/nq/Meta-Llama-3.1-405B-Instruct/Meta-Llama-3.1-405B-Instruct_NQ_subEM_200distractors.png}
    \end{subfigure}
    &  
    \begin{subfigure}{0.329\textwidth}
        \includegraphics[width=0.9\linewidth]{images/nm/Meta-Llama-3.1-405B-Instruct/Meta-Llama-3.1-405B-Instruct_NM_math-verify_5distractors.png}
    \end{subfigure}
    & \hspace{-0.8cm} \raisebox{3ex}{\rotatebox[origin=r]{270}{\lightpurpletext{LLaMA-3.1-405B}}}   
    \\  
    \begin{subfigure}{0.329\textwidth}
        \includegraphics[width=0.9\linewidth]{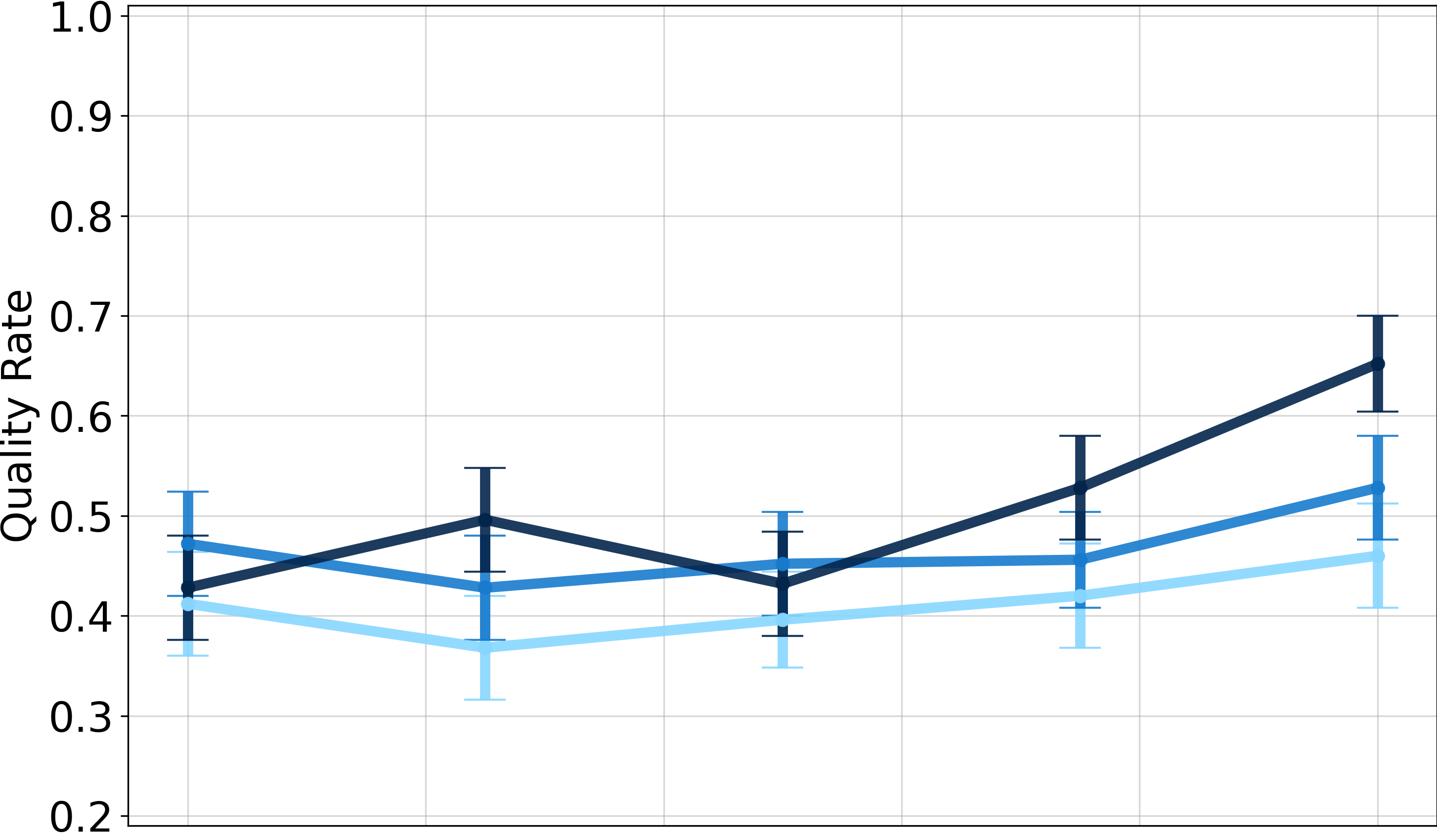}
    \end{subfigure}
    & 
    \begin{subfigure}{0.329\textwidth}
        \includegraphics[width=0.9\linewidth]{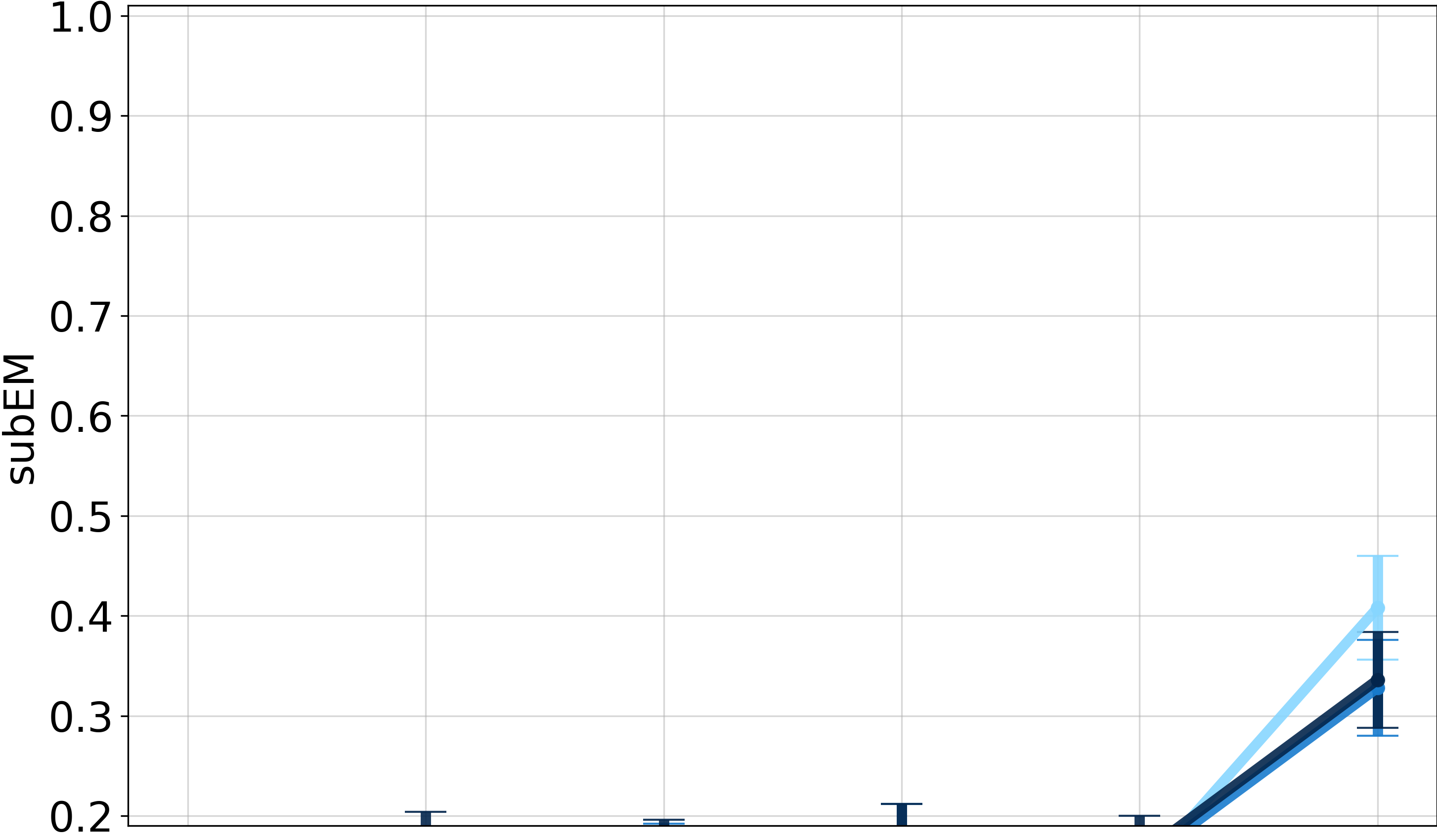}
    \end{subfigure}
    & 
    \begin{subfigure}{0.329\textwidth}
        \includegraphics[width=0.9\linewidth]{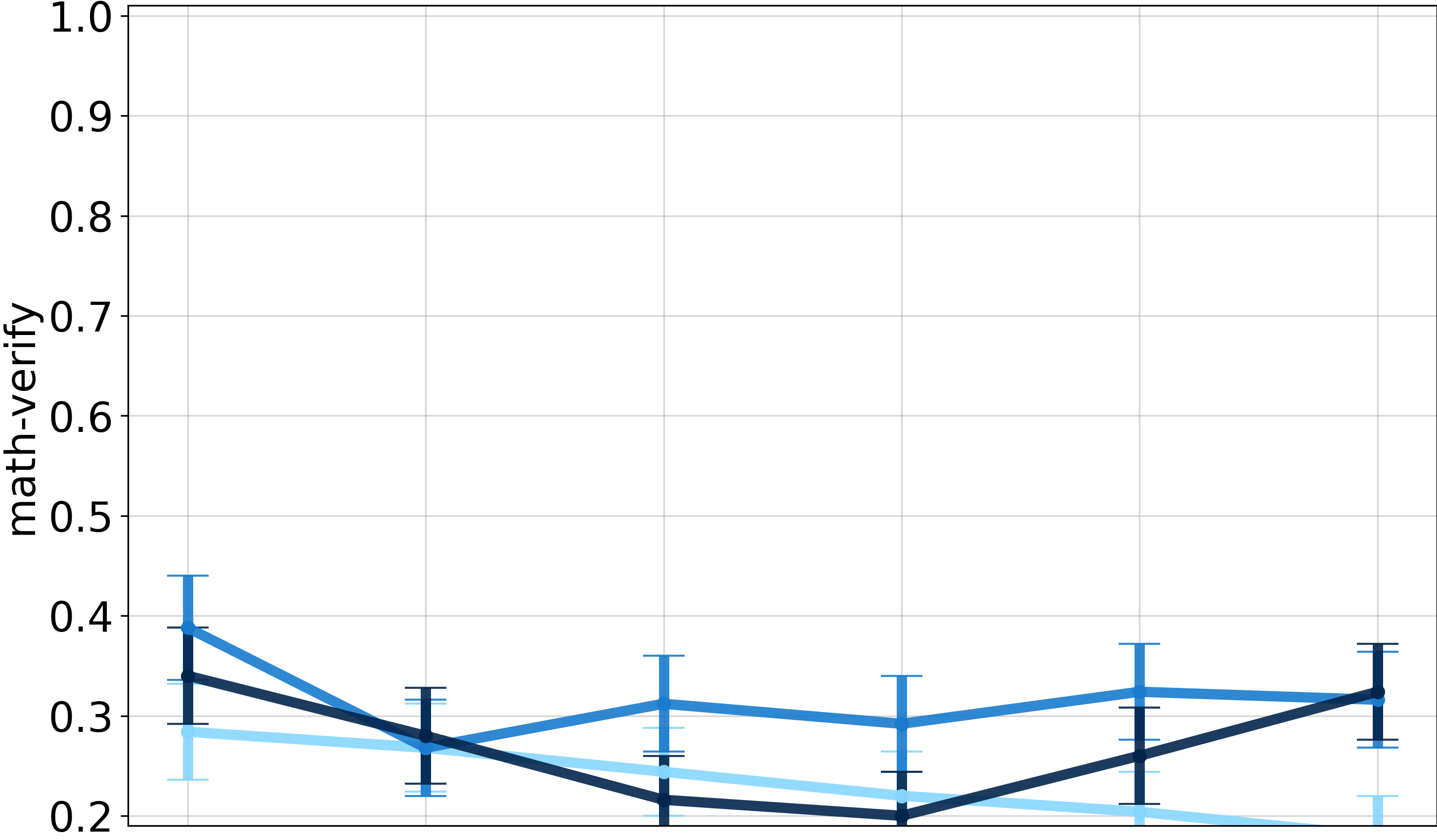}
    \end{subfigure}
    & 
    \hspace{-0.8cm} \raisebox{7ex}{\rotatebox[origin=r]{270}{\lightpurpletext{LLaMA-3.1-8B}}} 
    \\  
    \begin{subfigure}{0.329\textwidth}
        \includegraphics[width=0.9\linewidth]{images/cbb/Llama-3.3-70B-Instruct/Llama-3.3-70B-Instruct_CBB_QR.png}
    \end{subfigure}
    & 
    \begin{subfigure}{0.329\textwidth}
        \includegraphics[width=0.9\linewidth]{images/nq/Llama-3.3-70B-Instruct/Llama-3.3-70B-Instruct_NQ_subEM_200distractors.png}
    \end{subfigure}
    & 
    \begin{subfigure}{0.329\textwidth}
        \includegraphics[width=0.9\linewidth]{images/nm/Llama-3.3-70B-Instruct/Llama-3.3-70B-Instruct_NM_math-verify_5distractors.png}
    \end{subfigure}
    & 
    \hspace{-0.8cm} \raisebox{7ex}{\rotatebox[origin=r]{270}{\lightpurpletext{LLaMA-3.3-70B}}}
     
    \end{tabular}    
    \caption{
    Model performance by gold context position (early to late in input), higher is better and error bars are 90\% CIs. Each row is a model, columns are benchmarks. 
    \textbf{Smaller gold contexts exhibit sharper performance degradation with later placement, especially in specialized domains (CBB, NM).}
    Larger contexts mitigate this sensitivity, highlighting the stabilizing effect of richer input. All non-reasoning models, including the ones in Figure~\ref{fig:byposition_all_models}, are here for comparison.
    }
    \label{fig:byposition_all_models_supps}
\end{figure}

\begin{figure}[ht]
    \centering
    \tiny
    \setlength{\tabcolsep}{2pt}
    \begin{tabular}{cccc}
         \lightpurpletext{CARDBiomedBench} & \lightpurpletext{NaturalQuestions} & \lightpurpletext{NuminaMath1.5} & 
    \\
    
    \begin{subfigure}{0.329\textwidth}
        \includegraphics[width=0.9\linewidth]{images/cbb/DeepSeek-R1-0528/DeepSeek-R1-0528_CBB_QR.png}
    \end{subfigure} 
    & 
    \begin{subfigure}{0.329\textwidth}
        \includegraphics[width=0.9\linewidth]{images/nq/DeepSeek-R1-0528/DeepSeek-R1-0528_NQ_subEM_200distractors.png}
    \end{subfigure}
    & 
    \begin{subfigure}{0.329\textwidth}
        \includegraphics[width=0.9\linewidth]{images/nm/DeepSeek-R1-0528/DeepSeek-R1-0528_NM_math-verify_5distractors.png}
    \end{subfigure}
    & 
    \hspace{-0.8cm} \raisebox{13ex}{\rotatebox[origin=c]{270}{\lightpurpletext{DeepSeek-R1}}} 
    \\ 
    
    \begin{subfigure}{0.329\textwidth}
        \includegraphics[width=0.9\linewidth]{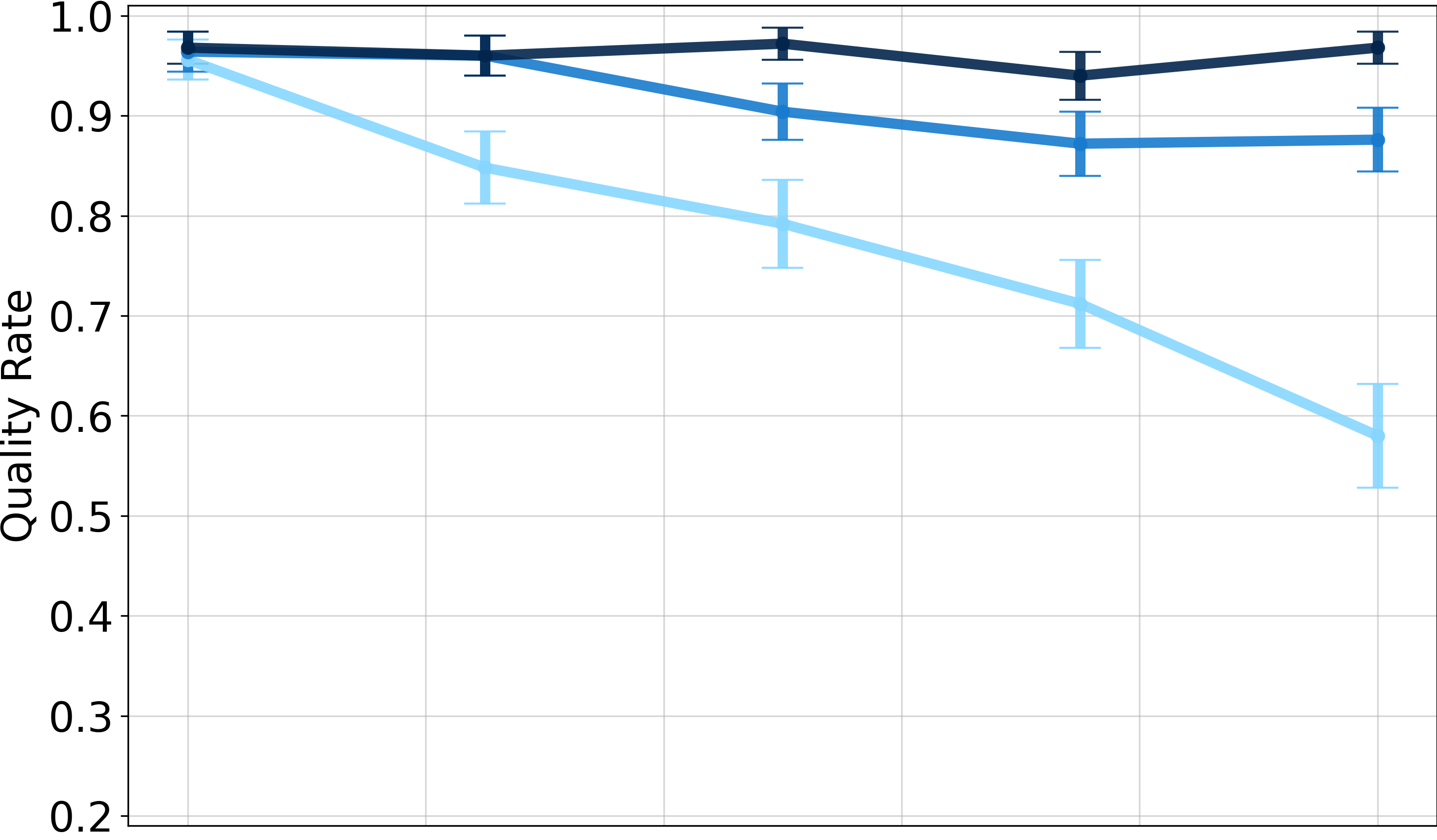}
    \end{subfigure} 
    & 
    \begin{subfigure}{0.329\textwidth}
        \includegraphics[width=0.9\linewidth]{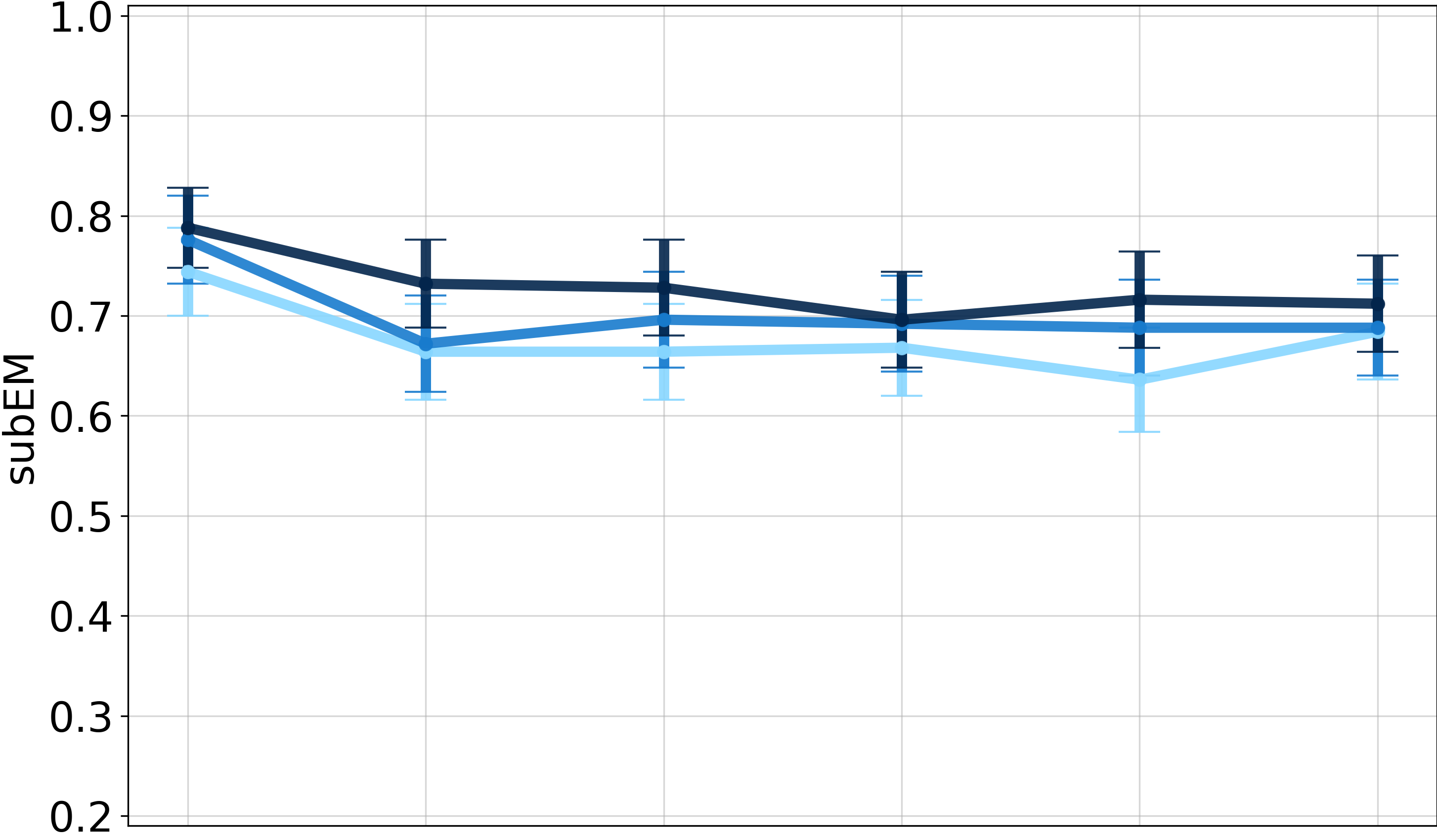}
    \end{subfigure}
    & 
    \begin{subfigure}{0.329\textwidth}
        \includegraphics[width=0.9\linewidth]{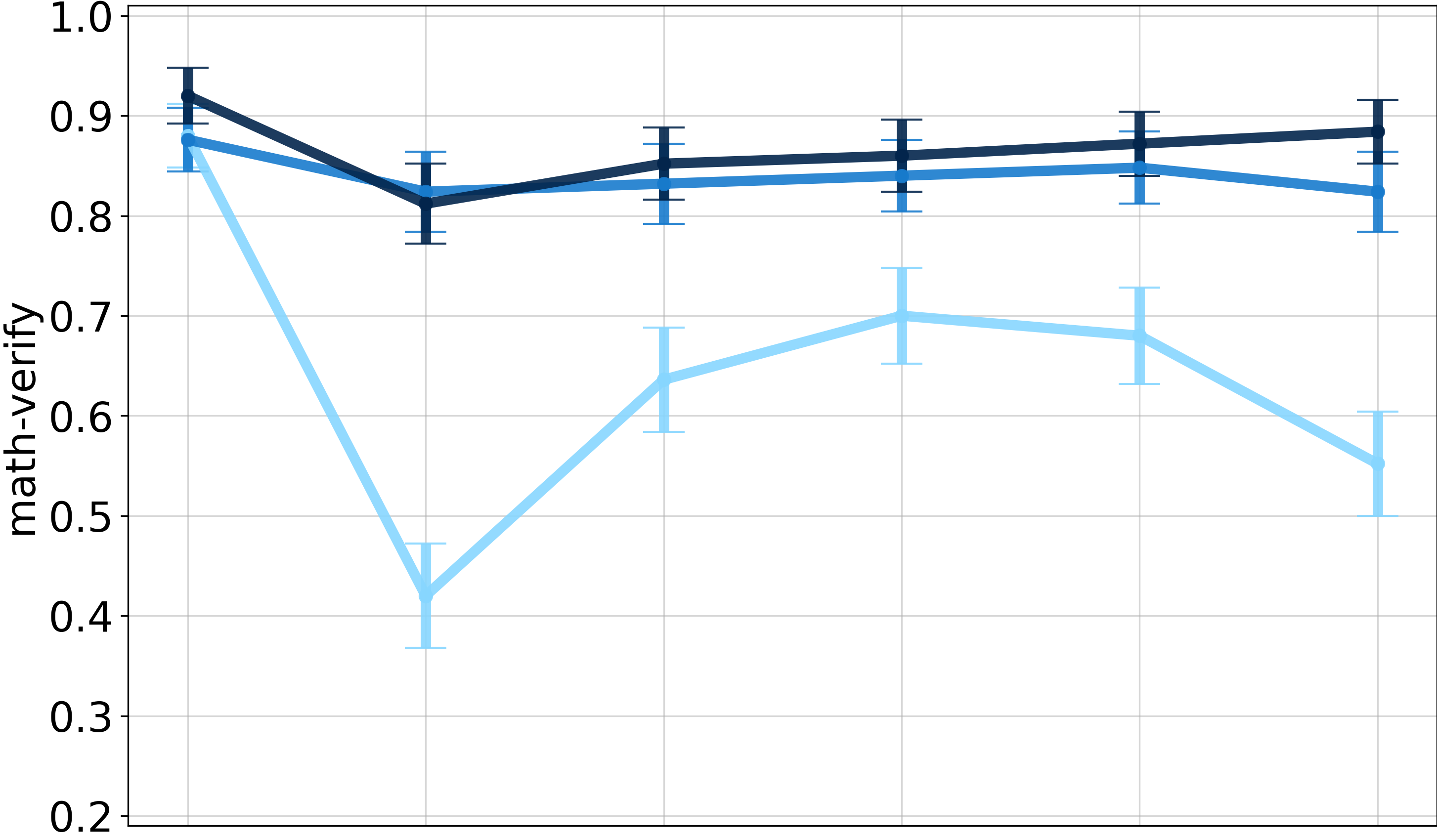}
    \end{subfigure}
    & 
    \hspace{-0.8cm} \raisebox{13ex}{\rotatebox[origin=c]{270}{\lightpurpletext{Gemini-2.5-Flash}}} 
    \\
    \begin{subfigure}{0.329\textwidth}
        \includegraphics[width=0.9\linewidth]{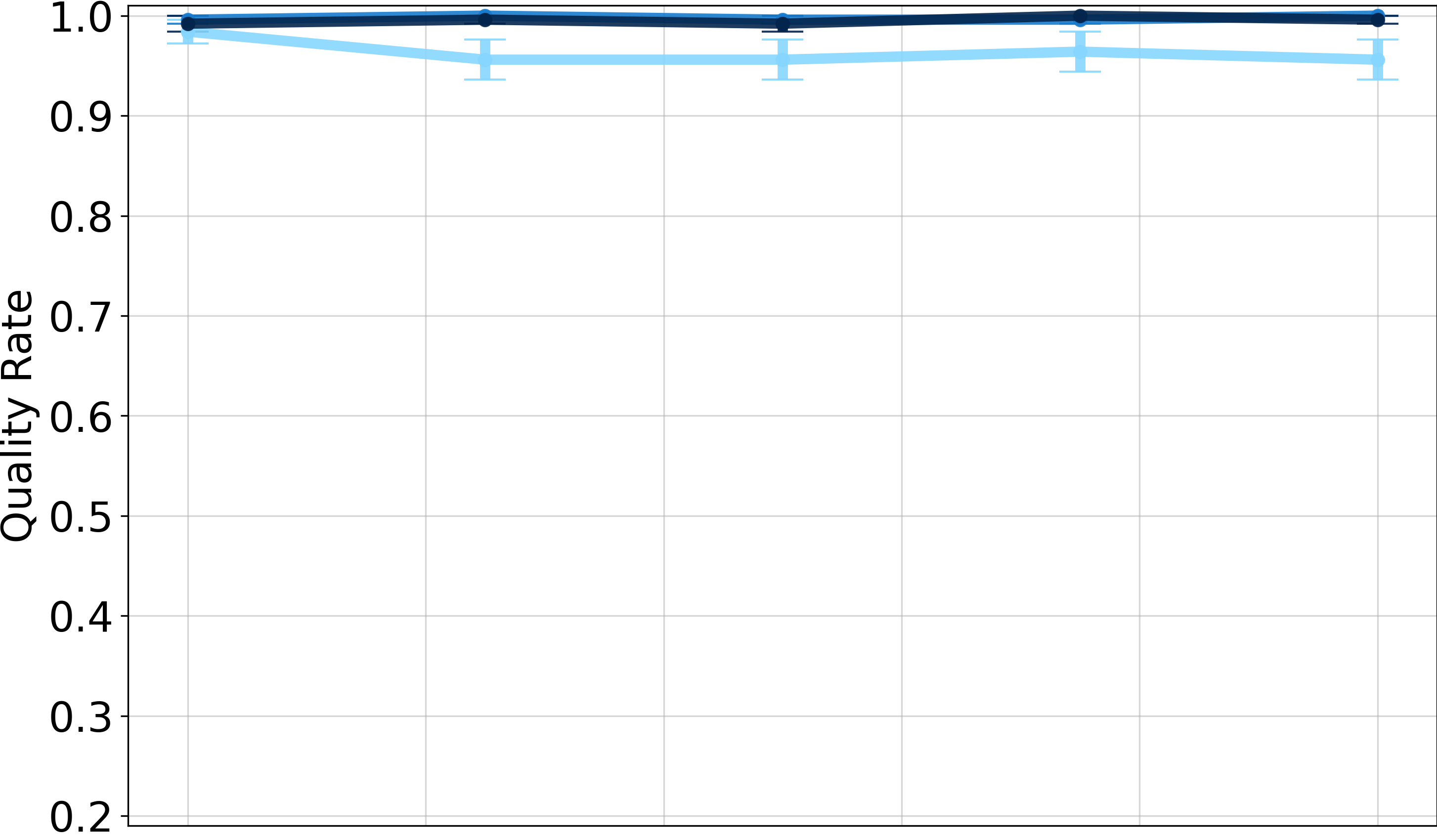}
    \end{subfigure}     
     & 
    \begin{subfigure}{0.329\textwidth}
        \includegraphics[width=0.9\linewidth]{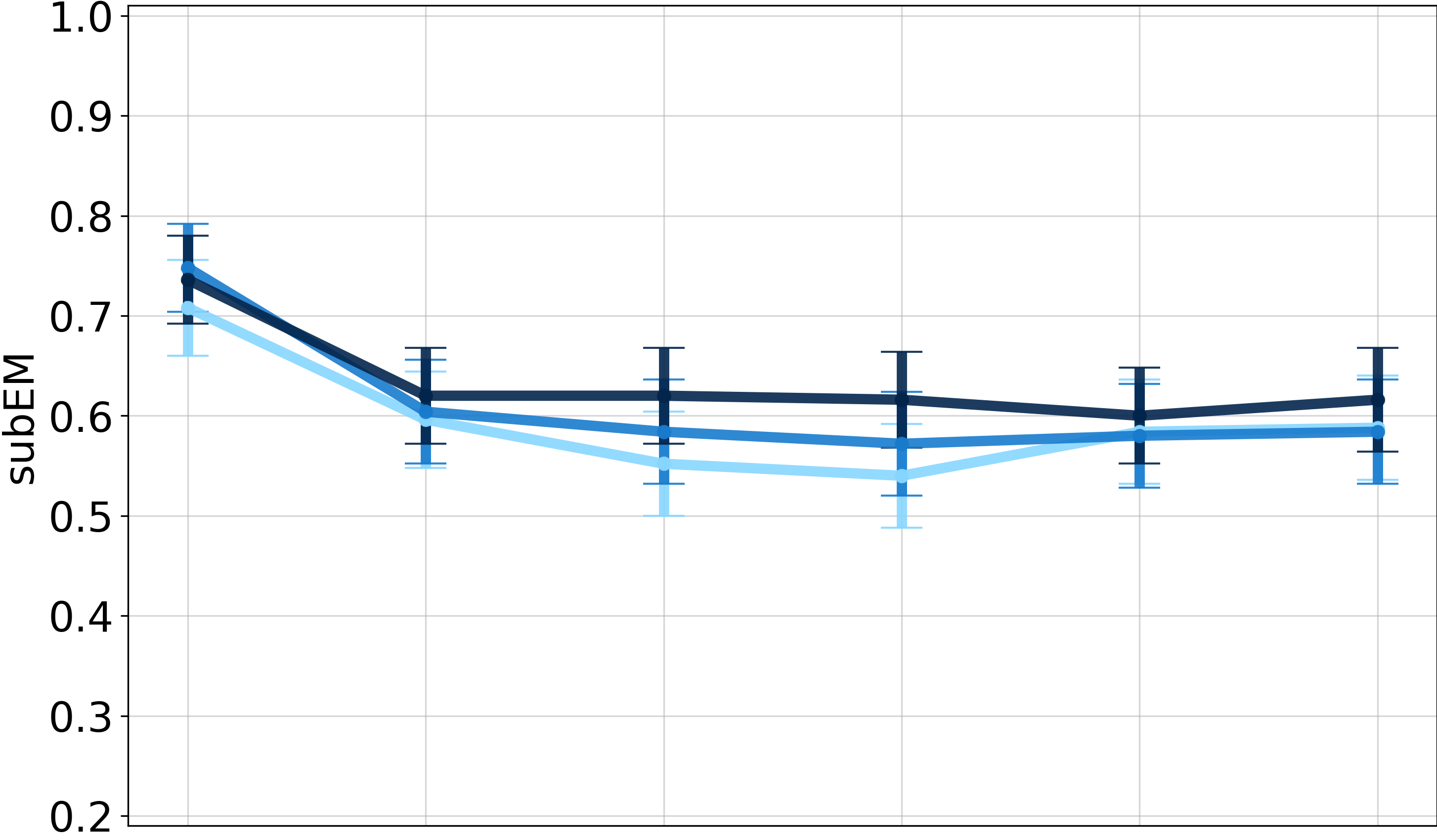}
    \end{subfigure}
     & 
    \begin{subfigure}{0.329\textwidth}
        \includegraphics[width=0.9\linewidth]{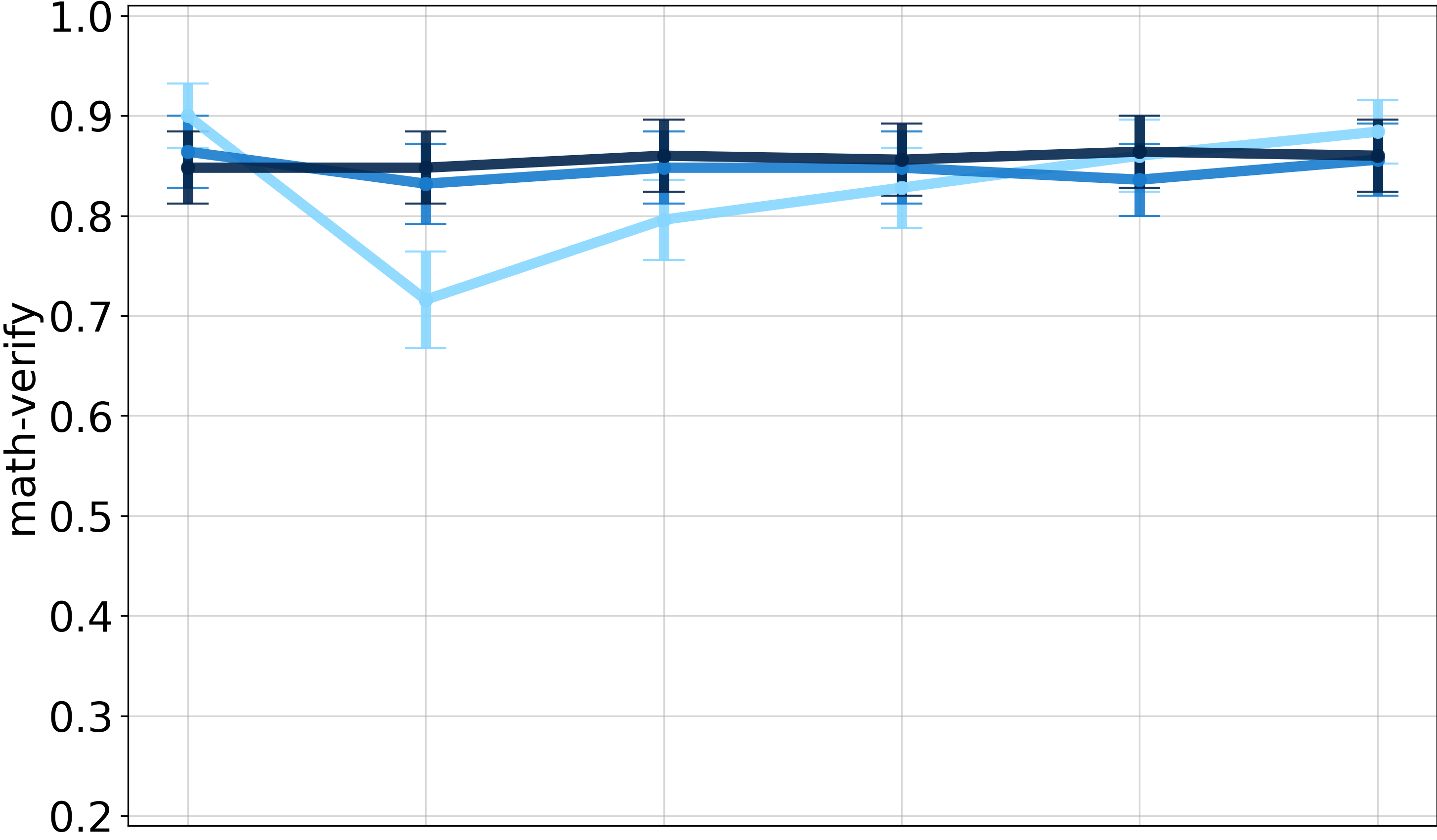}
    \end{subfigure}
    & \hspace{-0.8cm} \raisebox{9ex}{\rotatebox[origin=r]{270}{\lightpurpletext{o3-mini}}}
    \\
    \begin{subfigure}{0.329\textwidth}
        \includegraphics[width=0.9\linewidth]{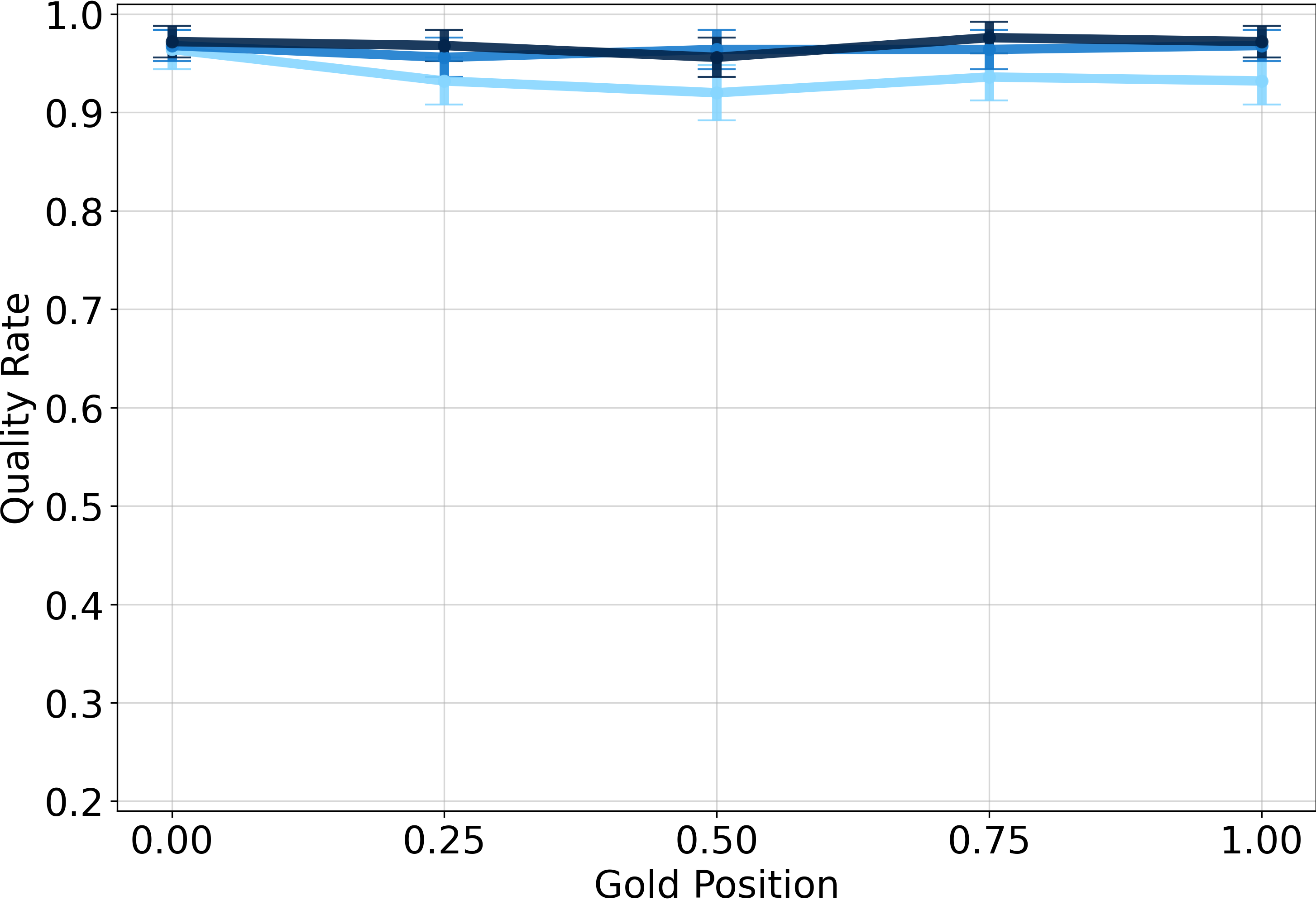}
    \end{subfigure}     
     & 
    \begin{subfigure}{0.329\textwidth}
        \includegraphics[width=0.9\linewidth]{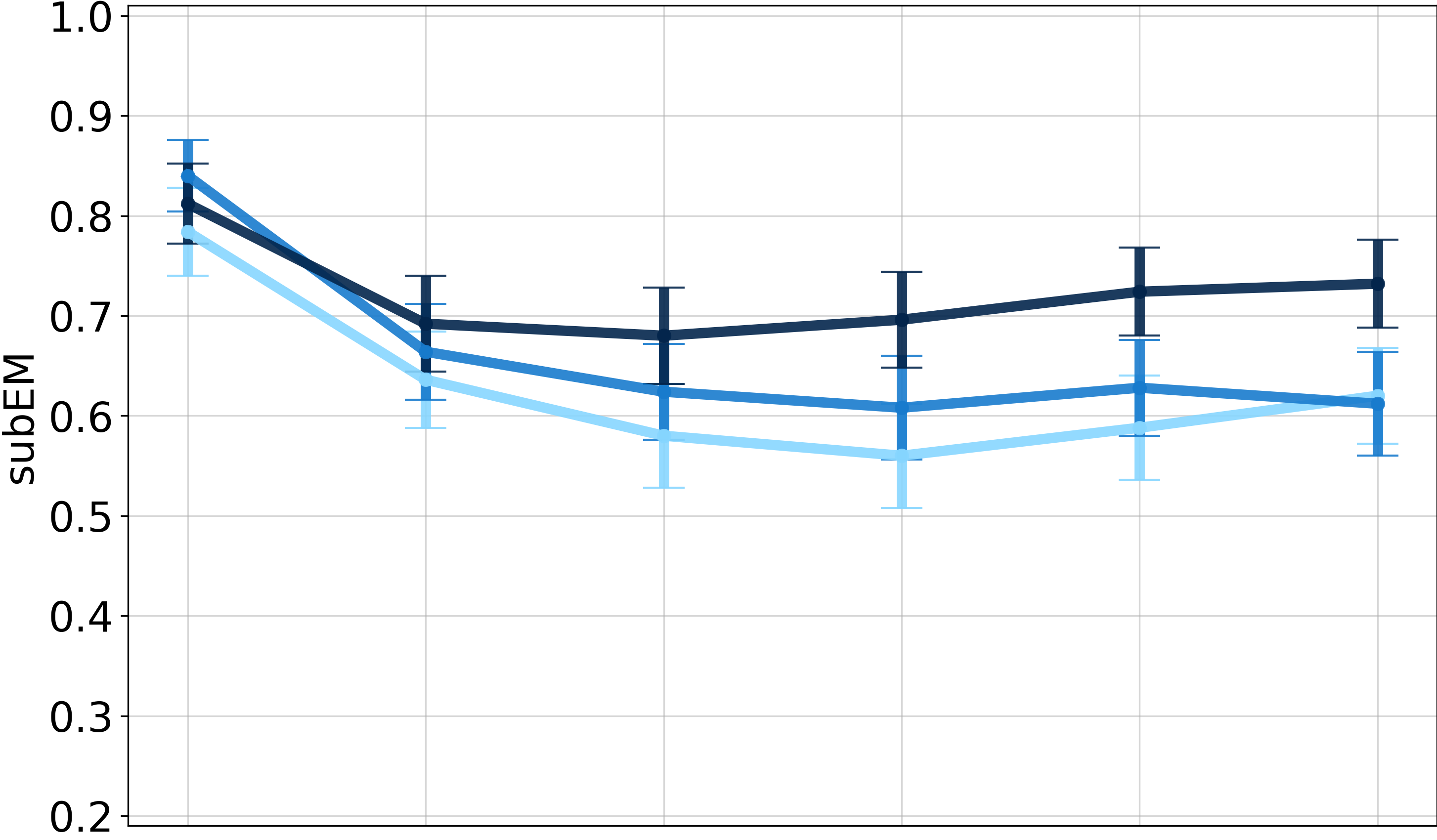}
    \end{subfigure}
     & 
    \begin{subfigure}{0.329\textwidth}
        \includegraphics[width=0.9\linewidth]{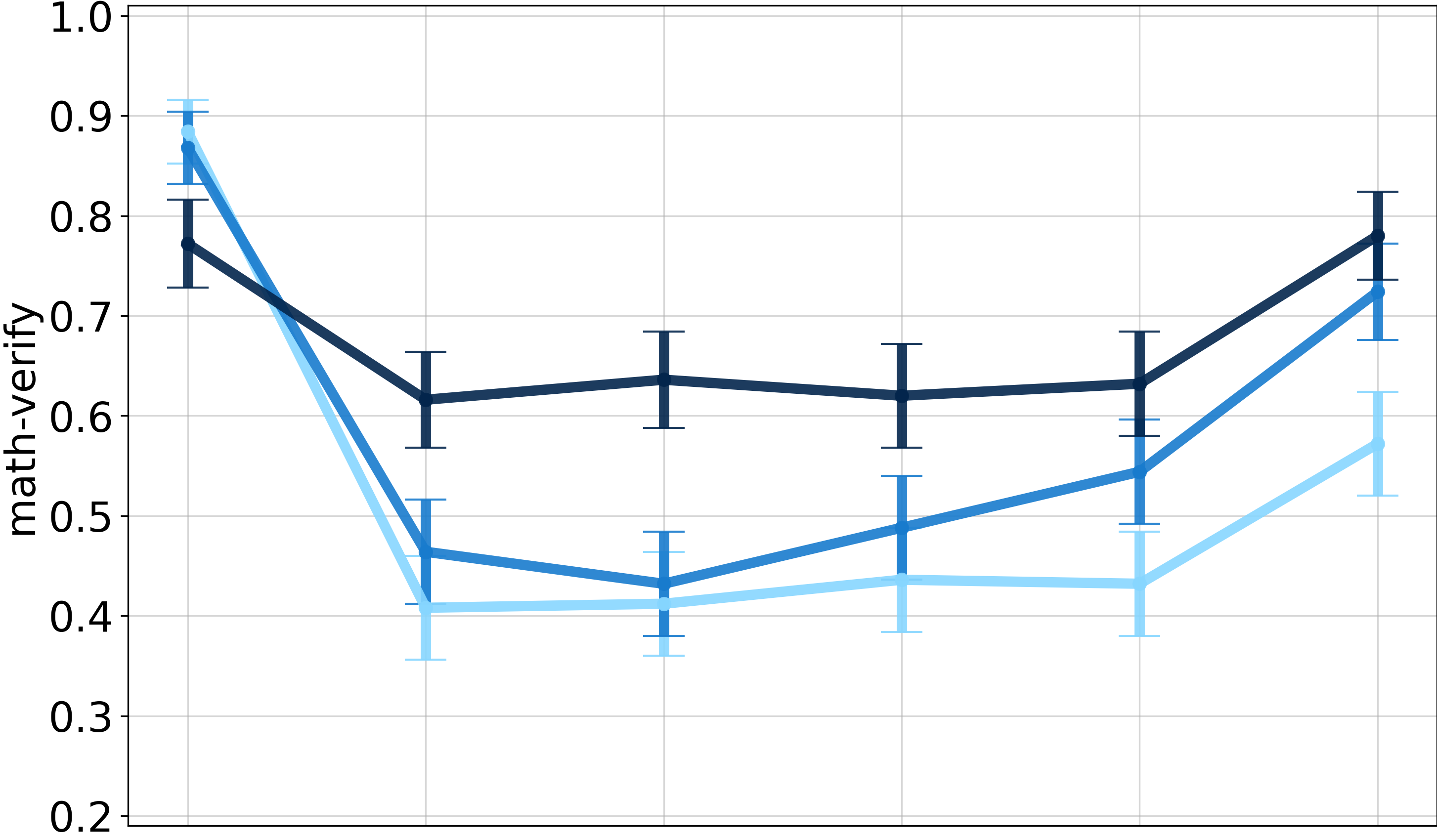}
    \end{subfigure}
    & \hspace{-0.8cm} \raisebox{9ex}{\rotatebox[origin=r]{270}{\lightpurpletext{Phi-4-reasoning}}}
    \end{tabular}    
    \caption{
    Reasoning model performance by gold context position (early to late in input), higher is better and error bars are 90\% CIs. Each row is a model, columns are benchmarks. 
    \textbf{Smaller gold contexts exhibit sharper performance degradation with later placement, especially in specialized domains (CBB, NM).}
    Larger contexts mitigate this sensitivity, highlighting the stabilizing effect of richer input.
    }
    \label{fig:byposition_all_models_supps}
\end{figure}

\clearpage
\subsection{Positional Sensitivity}
\label{subsec:pos_sens_appdx}

\begin{figure}[ht]
    \vspace{-0.1cm}
    \centering

    \begin{subfigure}{\textwidth}
        \centering
        \includegraphics[width=0.55\linewidth,trim=0cm 0.1cm 0cm 0.2cm]{images/cbb/cbb_scatter_range_Quality_Rate.png}
        \caption{CARDBiomedBench}
    \end{subfigure}

    \begin{subfigure}{\textwidth}
         \centering
        \includegraphics[width=0.55\linewidth,trim=0cm 0.1cm 0cm 0.2cm]{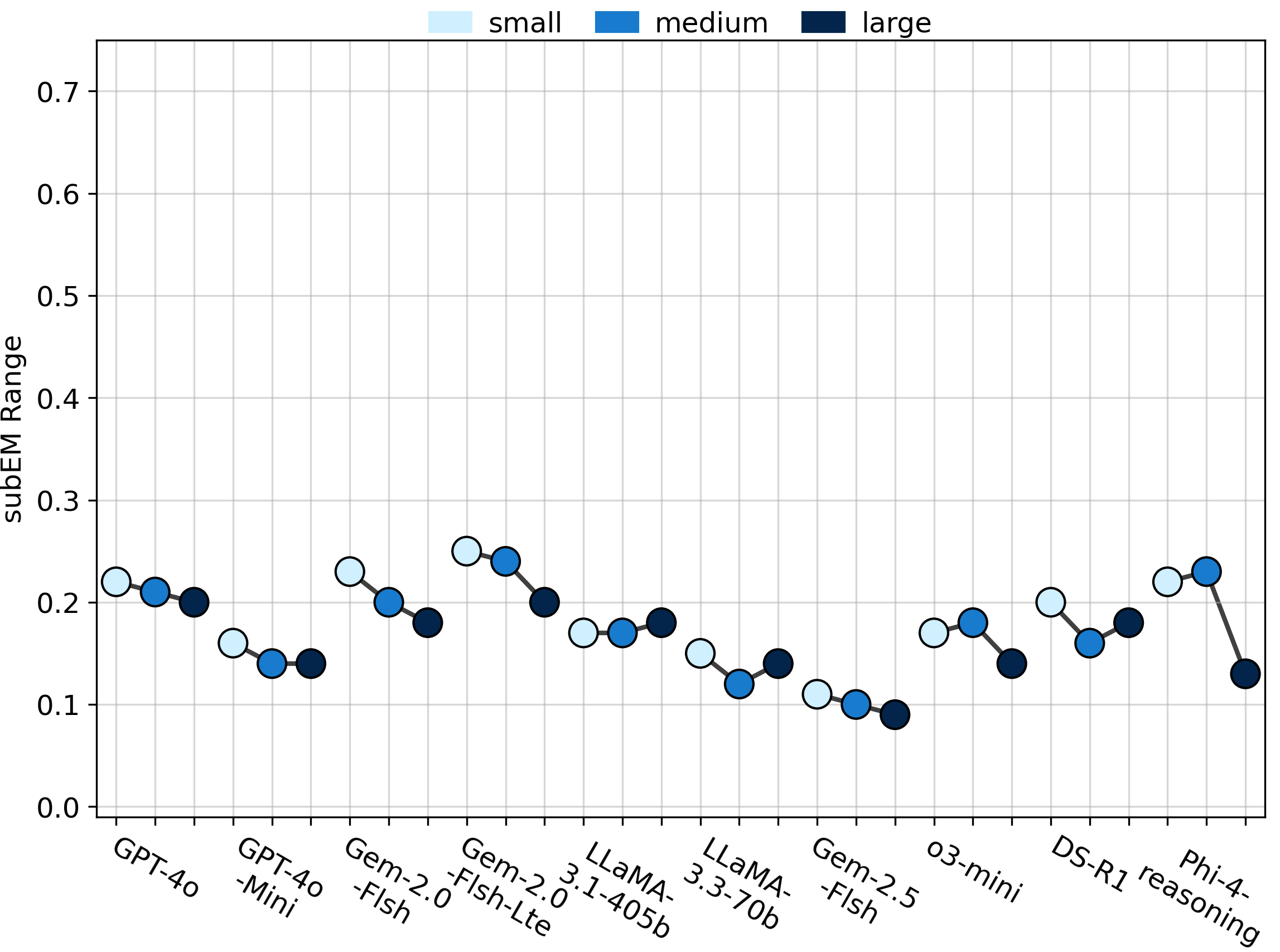}
        \caption{NaturalQuestions}
    \end{subfigure}

    \begin{subfigure}{\textwidth}
         \centering
        \includegraphics[width=0.55\linewidth,trim=0cm 0.1cm 0cm 0.2cm]{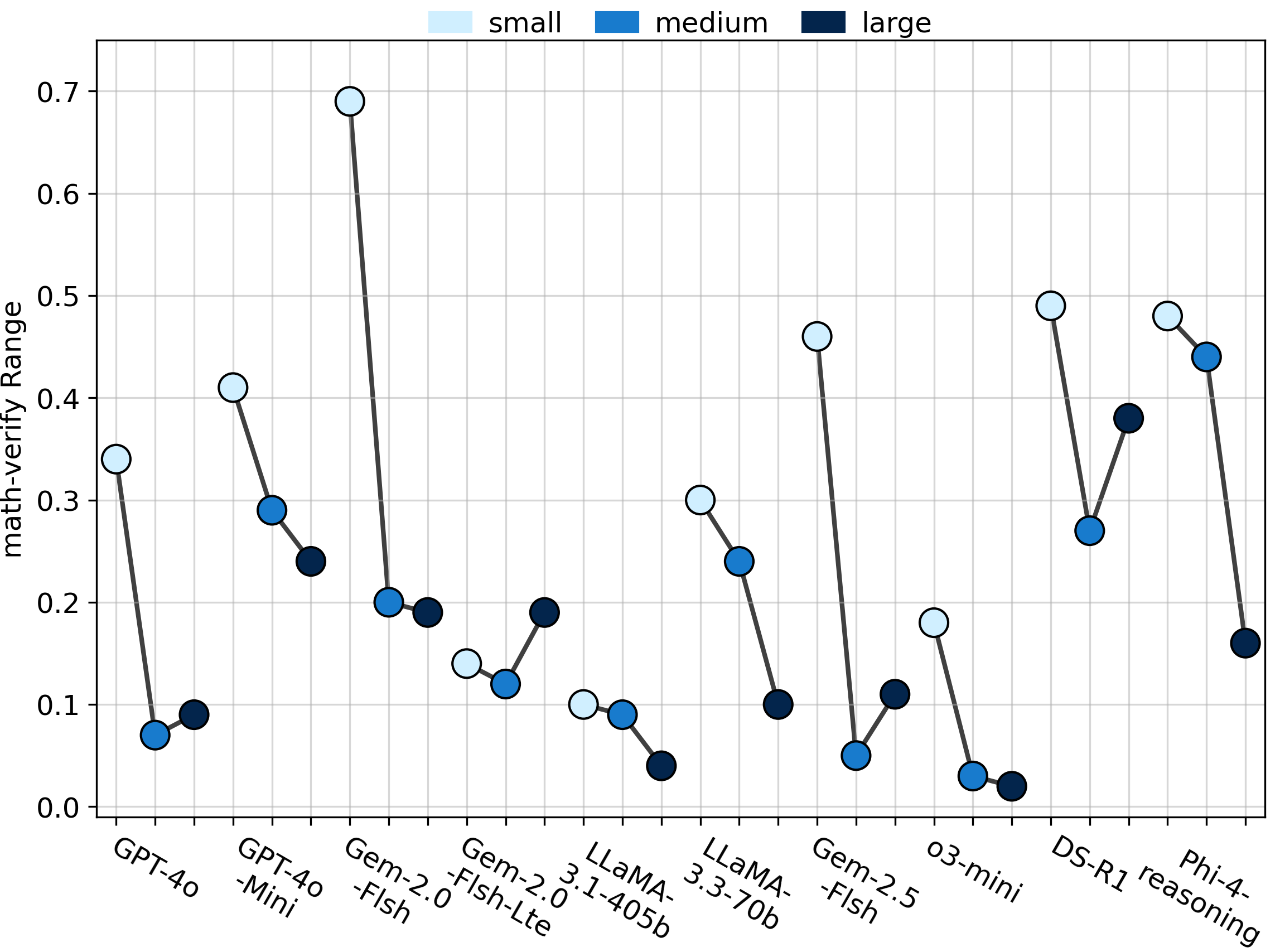}
        \caption{NuminaMath1.5}
    \end{subfigure}

    \caption{
    Positional sensitivity by benchmark. For each model and gold context size, we compute the range (Equation~\ref{eq:range}) of performance across positions.
    \textbf{Smaller gold contexts exhibit much higher sensitivity (larger ranges), especially in domain-specific tasks (CBB, NM).
    Larger gold contexts yield more stable performance across positions.}
    }
    \label{fig:range_by_dataset}
\end{figure}

\clearpage

\section{Confounder Analysis}
\label{apdx:conf}
We provide details, formulas, distributions, and performance across benchmarks when considering the potential confounding variables.

\subsection{Measuring Gold Context Ratios and Answer Overlap}
\label{apdx:conf_formulas}

We define several metrics to quantify the repetition of the answer across gold contexts, as well as the gold-to-distractor ratio.

\textbf{Gold-to-Distractor Ratio.}  
To measure the ratio of gold to distractor tokens, we define $T(x)$ as the tokens of passage $x$ and $|T(x)|$ is the total number of tokens in $x$:
\begin{equation}
\text{Gold-to-Distractor Ratio}(g, D) = \frac{|T(g)|}{\sum_{d \in D} |T(d)|}
\end{equation}

\textbf{Exact Mentions.}  
We count exact string occurrences of the answer in the context, case-insensitive and word-bounded:
\begin{equation}
\text{ExactMentions}(a, c) = \sum_{a_i \in \mathcal{A}} \; \# \{ \text{occurrences of } a_i \text{ in } c \}
\end{equation}
where $\mathcal{A}$ is the set of provided answer strings and $c$ is the context.

\textbf{Answer Token Hits.}  
At the token level, we measure how many context tokens match any token from the answer:
\begin{equation}
\text{AnsTokHits}(a, c) = \sum_{t \in T(c)} \mathbf{1}[t \in T(a)]
\end{equation}
where $T(x)$ is the tokenized version of $x$. This counts duplicates, i.e., repeated matches.

\textbf{Answer Token Repetition.}  

To normalize for answer length, we define repetition as raw answer-token hits per unique answer token:
\begin{equation}
\text{AnsTokRepetition}(a, c) = \frac{\text{AnsTokHits}(a, c)}{|T(a)|^*}
\end{equation}
where $|T(a)|^*$ is the number of unique tokens in the answer. This measures a normalized degree of repetition relative to the answer’s own size.

Exact mentions are rare (sparse), making them an unreliable measure, thus we opt for repetition rate to capture meaningful growth in answer-token recurrence.
\clearpage
\subsection{Confounder Distributions}
\label{apdx:conf_dist}

\begin{figure}[ht]
    \centering
    \includegraphics[width=0.99\linewidth]{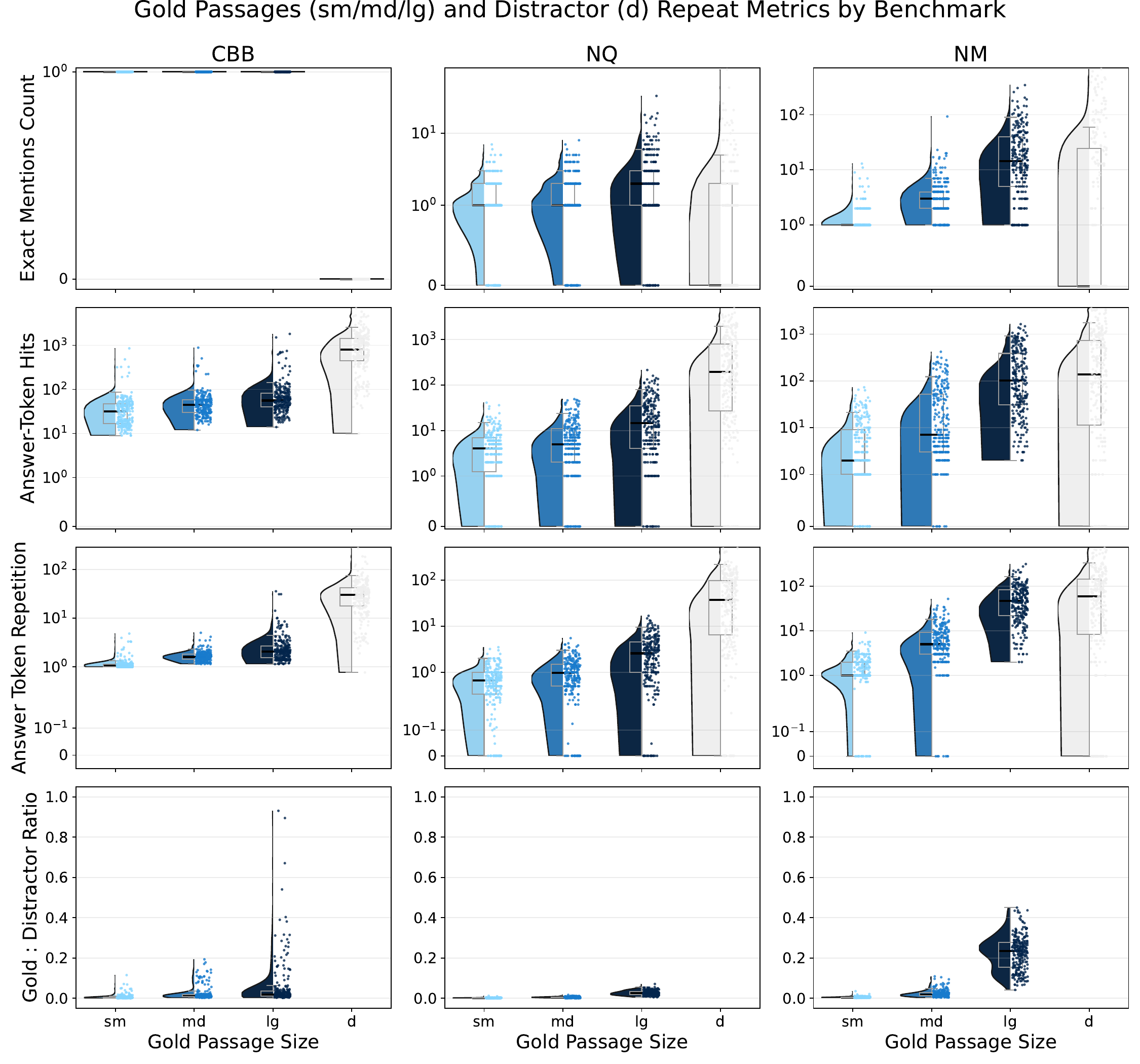}
    \caption{
    Raincloud plots across all benchmarks of Exact Mentions Counts, Answer-Token Hits, Redundancy, and Gold : Distractor Ratio across all sizes of gold and distractor documents for reference. 
    }
    \label{fig:repetition_effect}
\end{figure}

\clearpage
\subsection{Confounder Binned Results}
\label{apdx:conf_bins}

\textbf{Binning Procedure.} To analyze how accuracy varies with a given feature, we partition the feature into binary bins based on the mean value across all examples (small, medium, and large).



\begin{figure}[ht]
    \centering
    {
    \includegraphics[width=0.95\linewidth,trim=0.8cm 2cm 6cm 0cm]{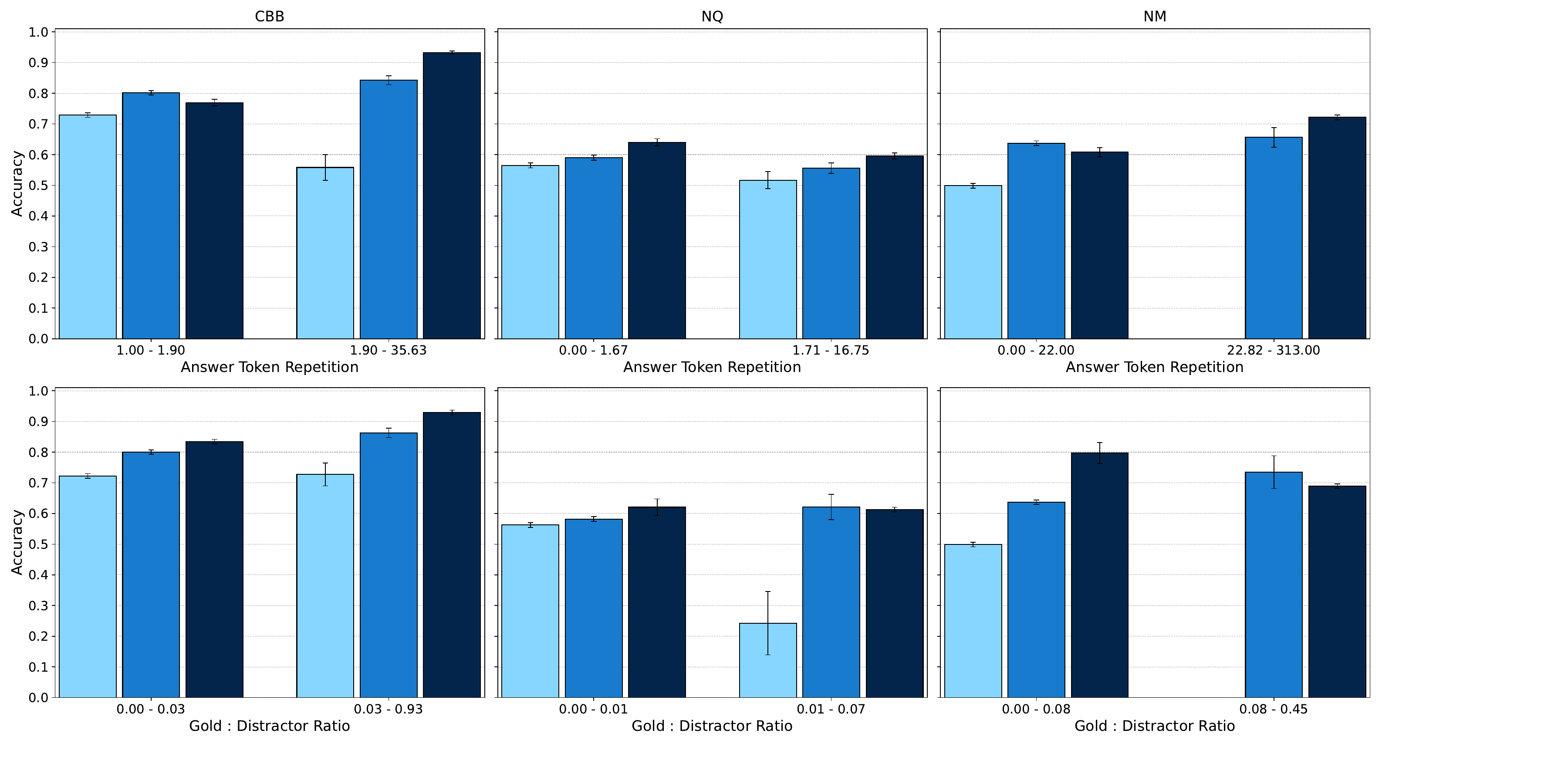}
    }
    \caption{
    \changed{Performance (across all models) when bucketing tasks into binary bins per confounder: below and above the mean value. \lightbluetext{Smaller golds} typically yields lower accuracy compared to \darkbluetext{larger golds}.} Error bars are 95\% confidence intervals, some are larger due to small sample size in that bin.
    }
    \label{fig:conf_bins}
\end{figure}

\clearpage
\changed{\subsection{Confounder Logistic Regression Analysis}}
\label{apdx:logreg}

\changed{To examine the statistical strength of each variable--gold size, answer token repetition, gold-to-distractor ratio, and position--we fit a multivariate logistic regression to predict correctness across all tasks and models for each benchmark. While the earlier analyses attempt to isolate each factor independently, these variables often co-vary in realistic long-context settings. The regression therefore provides a unified framework to estimate the partial contribution of each factor and determine whether gold context size remains an independent driver of performance.}

\subsubsection{Setup}

\textbf{Motivation.}

\changed{Smaller gold contexts naturally coincide with several conditions that make retrieval harder: they tend to repeat answer tokens less often, occupy a smaller share of the input window, and are more sensitive to position. These correlations can obscure the true source of the performance gap. By modeling correctness jointly over all variables, we separate genuine size effects from those attributable to repetition, ratio, or positional bias.}

\textbf{Model Specification.}

\changed{For each benchmark (CBB, NQ, NM), we regress the binary correctness outcome on four categorical predictors:}

\begin{equation}
\Pr\!\big(\mathrm{correct}\big)
=
\sigma\!\left(
\beta_0
+ \beta_{\text{size}}\, X_{\text{size}}
+ \beta_{\text{rep}}\, X_{\text{rep}}
+ \beta_{\text{ratio}}\, X_{\text{ratio}}
+ \beta_{\text{pos}}\, X_{\text{pos}}
\right)
\end{equation}

\changed{All predictors are encoded as categorical variables to keep coefficients directly interpretable--particularly for the confounders, where ‘high vs. low’ bins provide a clear contrast between conditions.}

\changed{\textbf{Variables.}
The regression includes four categorical predictors, each represented by a set of indicator variables $X_{\text{size}}$, $X_{\text{rep}}$, $X_{\text{ratio}}$, and $X_{\text{pos}}$ in the model above. Both confounder variables (repetition and ratio) are binarized into \textit{high} vs.\ \textit{low} using within-benchmark means to keep coefficients directly interpretable.}

\begin{enumerate}[leftmargin=*]
    \item \textbf{Gold Size ($X_{\text{size}}$).}
    A categorical factor with \textit{small} gold contexts as the reference level; the indicator variables encode the \textit{medium} and \textit{large} gold conditions. The coefficient $\beta_{\text{size}}$ captures the change in log-odds of correctness when moving from a small gold context to a larger one (small $\rightarrow$ medium and medium $\rightarrow$ large).

    \item \textbf{Answer Token Repetition ($X_{\text{rep}}$).}
    Token-level overlap between the gold context and the answer, binarized at the benchmark mean into \textit{low} (reference) vs.\ \textit{high} repetition. The coefficient $\beta_{\text{rep}}$ represents the difference in log-odds between below-mean and above-mean repetition levels.

    \item \textbf{Gold-to-Distractor Ratio ($X_{\text{ratio}}$).}
    The proportion of gold to distractor tokens in the context window, binarized at the benchmark mean into \textit{low} (reference) vs.\ \textit{high}. The coefficient $\beta_{\text{ratio}}$ quantifies the effect of moving from a lower share of gold tokens to a higher one.

    \item \textbf{Position ($X_{\text{pos}}$).}
    The normalized start position of the gold context, treated categorically with $0.0$ (start of window) as the reference; indicator variables encode all other positions. The coefficient $\beta_{\text{pos}}$ captures how changes in placement within the input window influence correctness relative to appearing first.
\end{enumerate}

\textbf{Results.}

\changed{\textbf{Across all three benchmarks, gold context size remains a significant and independent predictor of correctness after controlling for answer token repetition, gold-to-distractor ratio, and gold context position.} Medium and large gold contexts consistently increase the log-odds of producing the correct answer relative to small gold contexts.}

\clearpage
\subsubsection{CBB Logistic Regression Output}

\begin{table}[h!]
\centering
\small
\setlength{\tabcolsep}{10pt}
\begin{tabular}{lccccc}
    \toprule
    \textbf{Predictor} & \textbf{Coef} & \textbf{Std. Err.} & \textbf{z} & \textbf{P>|z|} & \textbf{95\% CI} \\
    \midrule
    Intercept & 1.6459 & 0.038 & 43.284 & 0.00 & [1.571, 1.720] \\
    \midrule
    \multicolumn{5}{l}{Gold Size} \\
    \quad md vs.\ sm & 0.3933 & 0.029 & 13.340 & 0.00 &  [0.336, 0.451] \\
    \quad lg vs.\ sm  & 0.5354 & 0.036 & 15.034 & 0.00 &  [0.466, 0.605] \\
    \midrule
    \multicolumn{5}{l}{Ans Tok Rep} \\
    \quad high vs.\ low & 0.5924 & 0.041 & 14.309 & 0.00 &  [0.511, 0.674] \\
    \midrule
    \multicolumn{5}{l}{Gold-to-Dis Ratio} \\
    \quad high vs.\ low & 0.3699 & 0.047 & 7.936 & 0.00 &  [0.279, 0.461] \\
    \midrule
    \multicolumn{5}{l}{Position} \\
    \quad 0.25 vs.\ 0.0 & -0.8087 & 0.045 & -18.089 & 0.00 &  [-0.896, -0.721] \\
    \quad 0.50 vs.\ 0.0 & -0.9001 & 0.044 & -20.305 & 0.00 &  [-0.987, -0.813] \\
    \quad 0.75 vs.\ 0.0 & -0.8930 & 0.044 & -20.132 & 0.00 &  [-0.980, -0.806] \\
    \quad 1.00 vs.\ 0.0 & -0.8701 & 0.044 & -19.576 & 0.00 &  [-0.957, -0.783] \\
    \bottomrule
\end{tabular}
\caption{Logistic Regression Results (CBB)}
\label{tab:logreg:cbb}
\end{table}

\changed{Gold context size shows a clear independent effect on CBB: medium and large golds significantly increase correctness relative to small ones, even after adjusting for all confounders. Answer-token repetition and gold-to-distractor ratio also provide meaningful boosts, while position has the strongest negative impact. Performance drops sharply when the gold appears anywhere other than the start of the window. \textbf{CBB performance improves reliably with larger gold contexts, offering gains that cannot be explained by repetition, ratio, or position alone.}}

\clearpage
\subsubsection{NQ Logistic Regression Output}

\begin{table}[h!]
\centering
\small
\setlength{\tabcolsep}{10pt}
\begin{tabular}{lccccc}
    \toprule
    \textbf{Predictor} & \textbf{Coef} & \textbf{Std. Err.} & \textbf{z} & \textbf{P>|z|} & \textbf{95\% CI} \\
    \midrule
    Intercept & 0.7280 & 0.027 & 26.829 & 0.00 & [0.675, 0.781] \\
    \midrule
    \multicolumn{5}{l}{Gold Size} \\
    \quad md vs.\ sm & 0.1129 & 0.023 & 4.992 & 0.00 & [0.069, 0.157] \\
    \quad lg vs.\ sm & 0.2916 & 0.051 & 5.682 & 0.00 & [0.191, 0.392] \\
    \midrule
    \multicolumn{5}{l}{Ans Tok Rep} \\
    \quad high vs.\ low & -0.1737 & 0.023 & -7.395 & 0.00 & [-0.220, -0.128] \\
    \midrule
    \multicolumn{5}{l}{Gold-to-Dis Ratio} \\
    \quad high vs.\ low & 0.0206 & 0.050 & 0.414 & 0.679 & [-0.077, 0.118] \\
    \midrule
    \multicolumn{5}{l}{Position} \\
    \quad 0.2 vs.\ 0.0 & -0.5486 & 0.033 & -16.795 & 0.00 & [-0.613, -0.485] \\
    \quad 0.4 vs.\ 0.0 & -0.5920 & 0.033 & -18.148 & 0.00 & [-0.656, -0.528] \\
    \quad 0.6 vs.\ 0.0 & -0.6279 & 0.033 & -19.264 & 0.00 & [-0.692, -0.564] \\
    \quad 0.8 vs.\ 0.0 & -0.6397 & 0.033 & -19.630 & 0.00 & [-0.704, -0.576] \\
    \quad 1.0 vs.\ 0.0 & -0.3815 & 0.033 & -11.601 & 0.00 & [-0.446, -0.317] \\
    \bottomrule
\end{tabular}
\caption{Logistic Regression Results (NQ)}
\label{tab:logreg:nq}
\end{table}

\changed{Gold context size remains a significant predictor on NQ, though its effect is smaller than in the other benchmarks. Position exerts a substantial influence, with performance declining steadily as the gold moves deeper into the window. High answer-token repetition has a \textit{negative effect on NQ}, and gold-to-distractor ratio shows no meaningful impact. \textbf{NQ is less sensitive to gold size and confounders than the domain-specific benchmarks, but larger gold contexts provide a consistent advantage.}}

\clearpage
\subsubsection{NM Logistic Regression Output}

\begin{table}[h!]
\centering
\small
\setlength{\tabcolsep}{10pt}
\begin{tabular}{lccccc}
    \toprule
    \textbf{Predictor} & \textbf{Coef} & \textbf{Std. Err.} & \textbf{z} & \textbf{P>|z|} & \textbf{95\% CI} \\
    \midrule
    Intercept & 0.5660 & 0.028 & 20.348 & 0.00 & [0.511, 0.621] \\
    \midrule
    \multicolumn{5}{l}{Gold Size} \\
    \quad md vs.\ sm & 0.5675 & 0.023 & 24.926 & 0.00 & [0.523, 0.612] \\
    \quad lg vs.\ sm & 0.9641 & 0.080 & 12.005 & 0.00 & [0.807, 1.122] \\
    \midrule
    \multicolumn{5}{l}{Ans Tok Rep} \\
    \quad high vs.\ low & 0.4998 & 0.036 & 14.046 & 0.00 & [0.430, 0.569] \\
    \midrule
    \multicolumn{5}{l}{Gold-to-Dis Ratio} \\
    \quad high vs.\ low & -0.5138 & 0.083 & -6.220 & 0.00 & [-0.676, -0.352] \\
    \midrule
    \multicolumn{5}{l}{Position} \\
    \quad 0.2 vs.\ 0.0 & -0.8926 & 0.034 & -26.408 & 0.00 & [-0.959, -0.826] \\
    \quad 0.4 vs.\ 0.0 & -0.7459 & 0.034 & -22.000 & 0.00 & [-0.812, -0.679] \\
    \quad 0.6 vs.\ 0.0 & -0.6666 & 0.034 & -19.611 & 0.00 & [-0.733, -0.600] \\
    \quad 0.8 vs.\ 0.0 & -0.6333 & 0.034 & -18.610 & 0.00 & [-0.700, -0.567] \\
    \quad 1.0 vs.\ 0.0 & -0.4755 & 0.034 & -13.867 & 0.00 & [-0.543, -0.408] \\
    \bottomrule
\end{tabular}
\caption{Logistic Regression Results (NM)}
\label{tab:logreg:nm}
\end{table}

\changed{NM shows the strongest size effect: medium and especially large gold contexts yield substantial gains over small ones, even after accounting for confounders. High answer-token repetition improves performance, but gold-to-distractor ratio has a negative coefficient, indicating that ratio alone does not explain the size advantage. Position is highly impactful, with penalties for placement beyond the beginning of the window. \textbf{NM is the most position-sensitive and gold-size-dependent benchmark, underscoring that larger gold contexts are crucial for stabilizing NIAH performance.}}

\end{document}